\documentclass[10pt,twocolumn,letterpaper]{article}

\usepackage{iccv}

\usepackage[utf8]{inputenc}

\usepackage{times}
\usepackage{epsfig}
\usepackage{graphicx}
\usepackage{subcaption}
\usepackage{amsmath}
\usepackage{amssymb}

\usepackage[usestackEOL]{stackengine}

\usepackage[dvipsnames,table]{xcolor}

\usepackage{stfloats}

\usepackage{enumerate}
\usepackage{dirtytalk}
\usepackage{longtable}
\usepackage{multirow}
\usepackage{booktabs}
\usepackage{pifont}
\usepackage{float}
\usepackage{psfrag}
\usepackage[percent]{overpic}

\newcommand{\xmark}{\textcolor{red}{\ding{55}}}%
\newcommand{\cmark}{\textcolor{green}{\ding{51}}}%
\newcommand{\fns}[1]{\footnotesize{#1}}

\usepackage[obeyFinal]{easy-todo}

\usepackage[pagebackref=true,breaklinks=true,letterpaper=true,colorlinks,bookmarks=false]{hyperref}

\definecolor{medium-gray}{gray}{0.7}
\definecolor{light-gray}{gray}{0.87}
\definecolor{lighter-gray}{gray}{0.92}

\renewcommand\midrule{\specialrule{0.4pt}{0pt}{1pt}}

\iccvfinalcopy 

\def\iccvPaperID{56} 

\ificcvfinal\fi

\begin{document}

\title{On Moving Object Segmentation from Monocular Video with Transformers}

\author{Christian Homeyer\\
Robert Bosch GmbH, Corporate Research, Computer Vision Lab Hildesheim, Germany\\
Image and Pattern Analysis Group, Heidelberg University, Germany\\
{\tt\small homeyer@math.uni-heidelberg.de}
\and
Christoph Schnörr\\
Image and Pattern Analysis Group, Heidelberg University, Germany\\
{\tt\small schnoerr@math.uni-heidelberg.de}
}

\maketitle
\ificcvfinal\thispagestyle{empty}\fi

 \begin{abstract}
 	Moving object detection and segmentation from a single moving camera is a challenging task, requiring an understanding of recognition, motion and 3D geometry. 
 	Combining both recognition and reconstruction boils down to a fusion problem, where appearance and motion features need to be combined for classification and segmentation.\\
 	In this paper, we present a novel fusion architecture for monocular motion segmentation - $ \mathbf{M^{3}} $Former, which leverages the strong performance of transformers for segmentation and multi-modal fusion. As reconstructing motion from monocular video is ill-posed, we systematically analyze different 2D and 3D motion representations for this problem and their importance for segmentation performance.
 	Finally, we analyze the effect of training data and show that diverse datasets are required to achieve SotA performance on Kitti and Davis.
 \end{abstract}
\section{Introduction}
Interaction in a dynamic world requires reasoning about your surroundings and other dynamic agents. Motion segmentation plays a crucial part in autonomous perception systems, as we need this information for higher-level planning and navigation. 
It has exciting applications in down-stream tasks such as e.g. Neural Scene Synthesis \cite{liu2023robust} or Simultaneous Localization and Mapping (SLAM) \cite{zhao2022particlesfm}. Humans and animals can effortlessly perceive even completely unknown objects when observing them moving. This is in stark contrast to common image detectors \cite{cheng2021mask2former}, which are trained on large-scale datasets and are dependent on their respective finite label spaces. 
Combining motion and appearance data can resolve this issue and create generic object detectors, that generalize better across datasets \cite{dave2019towards, neoral2021monocular}. 
\begin{figure}[H]
	\centering
	\begin{overpic}[width=1.0\linewidth, height=7cm, tics=0, clip]
		{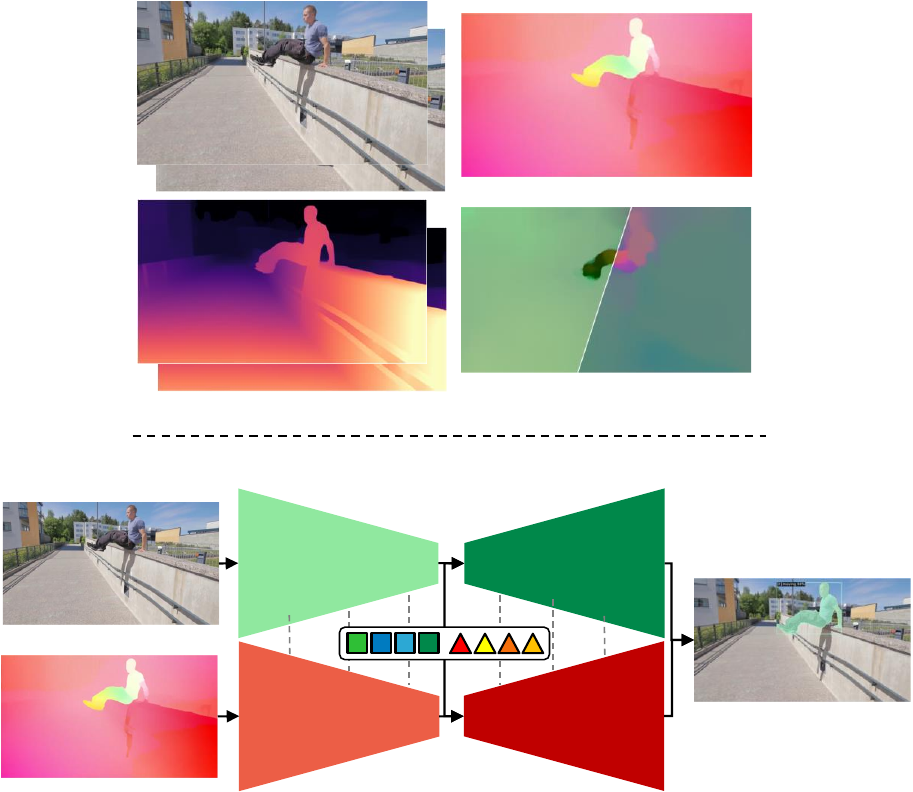}
		\put(27, 85){\footnotesize{Images}}
		\put(54, 85){\footnotesize{Optical Flow \cite{teed2020raft}}}
		\put(26, 39.5){\footnotesize{Depth \cite{ranftl2021vision}}}
		\put(56, 39.5){\footnotesize{Scene Flow \cite{teed2021raft}}}
		
		\put(53, 46){\footnotesize{$ R $}}
		\put(78, 46){\footnotesize{$ T $}}
		
		\put(4, 32){\footnotesize{Appearance}}
		\put(7, -2){\footnotesize{Motion}}
		
		\put(43.5, 18.5){\footnotesize{Queries}}
		\put(36, -2){\footnotesize{2-stream Transformer}}
		
		\put(82.5, 27){\footnotesize{Motion}}
		\put(78, 24){\footnotesize{Segmentation}}
	\end{overpic}
	\caption{Our \textbf{M}ulti-\textbf{M}odal \textbf{M}ask2Former ($ \textbf{M}^{3} $Former) framework for motion segmentation. Based on a monocular video, we compute a reconstruction based on frozen expert models \cite{teed2020raft, teed2021raft, ranftl2021vision}. This allows us to create (pseudo-) multi-modal data. We perform motion segmentation as a top-down fusion task with a segmentation transformer. We experiment both with 2D and 3D motion as input to our model.}
	\label{fig:eye}
\end{figure}
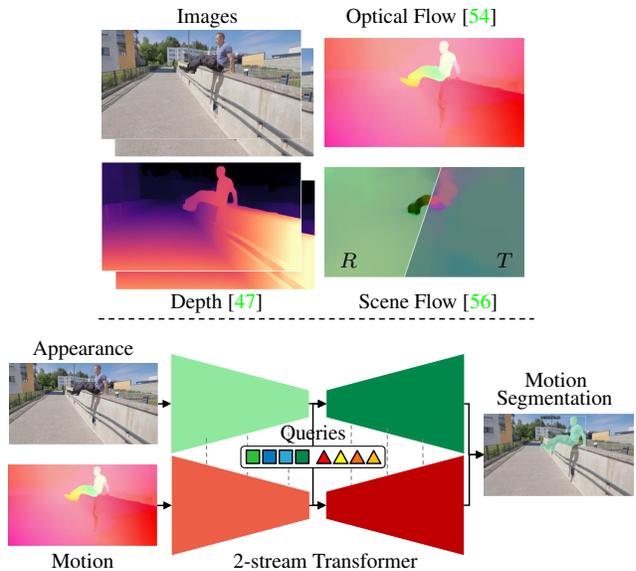
These findings align with the two-stream hypothesis in Neuroscience \cite{goodale1992separate}, which states that both appearance and motion are vital to biological visual systems. Motion segmentation can therefore be considered a multi-modal fusion problem. 
In this paper, we present a novel two-stream fusion architecture for motion segmentation. We combine both appearance and motion features in a transformer architecture \cite{cheng2021mask2former}. 

We call our framework \textbf{M}ulti-\textbf{M}odal \textbf{M}ask2Former ($\textbf{M}^{3}$Former), since we combine information from multiple modalities with masked attention. Since monocular video provides only a single modality stream, we make use of frozen expert models \cite{ranftl2021vision, teed2020raft, teed2021raft} for computing different motion representations, see Figure \ref{fig:eye}. Our contributions are fourfold:
\begin{itemize}
	\item We design a novel two-stream architecture with Encoder and Decoder. We analyze the performance of different fusion strategies within this framework.
	\item We systematically analyze the effect of different motion representations (Optical Flow, Scene Flow, higher-dimensional embeddings) from previous work within our framework.
	\item We empirically showcase the effect of diverse training data. Balancing different sources of motion patterns and semantic classes is crucial for strong performance on real-life video.
	\item We introduce a very simple augmentation technique for better multi-modal alignment. By introducing neg. examples with no motion information, we force the network to not over-rely on appearance data alone.
\end{itemize}

\subsection{Problem Statement}
Given a video $ \{ I_{1},\; I_{2},\; \dots,\; I_{N} \} $ from a single camera, we want to detect and segment \textit{generic independently moving objects}.
An \textit{object} is defined as a spatially connected group of pixels, belonging to the same semantic class. All labels are merged into a single \say{object}, since only the motion state matters.   
Detectors only see a finite number of classes during training. \textit{Generic} object detection assumes an inbalance between the set of training and test class labels. We want to identify any moving object, even if we have never seen the class during training.
An object is defined as \textit{independently moving} when its apparent motion is not due to camera ego-motion. The object is still considered moving when only a part is in motion, e.g. when a person moves an arm, then the whole person should be segmented. 

\section{Related Work}
\label{sec:related-work}
Segmenting objects based on their motion is a long standing problem in Computer Vision with a rich history \cite{darrell1991robust, irani1992detecting, torr1998geometric, torr1999problem, torr1998robust, shi1998motion, tron2007benchmark, vidal2003optimal, xu20193d, yuan2007detecting, brox2010object, ochs2013segmentation, bideau2016s, wulff2017optical} dating back to the early 90's. 

\textbf{Spatio-temporal Grouping and Geometric Modeling.}
Traditional approaches treat the problem as a spatio-temporal grouping problem, where similar 3D motions are clustered together \cite{torr1998geometric, shi1998motion, vidal2003optimal, vidal2004motion, yan2006general, brox2010object, ochs2013segmentation, fragkiadaki2011detection, bideau2016s, tsai2016video, xu20193d}. 
However, they focus on theoretical analysis with perfect input data, work on simplistic scenes and/or use sparse point trajectories.
 
A dominant line of work focuses on segmentation from two-frame optical flow, either by devising handcrafted geometric constraints \cite{bideau2016s, tokmakov2017learning}, e.g. motion angle and plane plus parallax (P+P) \cite{sawhney19943d}, or by learning directly from motion data \cite{bideau2018moa, yang2019unsupervised, liu2021emergence, yang2021self, xie2022segmenting}. Such approaches are affected by noisy inputs and cannot deal with degenerate cases like coplanar-colinear motion \cite{yuan2007detecting} and camera motion degeneracy \cite{torr1999problem}. 
Similar to us, \cite{lv2018learning} uses two RGB-D frames as input data and use a CNN to separate static background and dynamic foreground. However, they focus on high-quality depth maps and model motion with 2D optical flow and camera poses.
In order to deal with all motion cases and have a generic approach, \cite{yang2021learning} formulates extensive criteria beyond 2D motion. This requires a depth prior \cite{ranftl2021vision} and additional specialized neural network modules \cite{yang2020upgrading, brachmann2019neural}.
Our approach is indifferent towards geometric modeling: We analyze the importance of motion models in Section \ref{seq:experiments} by ablating different representations common in the literature. 
We will later see, that the effect on the downstream segmentation task is highly dependent on the datasets involved and the underlying quality of the geometric model inputs. Interestingly, weaknesses in geometric modelling can be compensated with local and global image information very effectively.

\textbf{Learning Video Object Segmentation.}
Object detection and segmentation from videos is closely related to salient object detection. Existing methods rely either on appearance features \cite{jain2017pixel} or motion features from optical flow \cite{bideau2018moa, liu2021emergence}. 
One line of research specializes on unsupervised motion segmentation \cite{lu2019see}, mostly from optical flow \cite{bideau2018moa, yang2019unsupervised, liu2021emergence, yang2021self, xie2022segmenting}. While this opens the avenue to train on large unlabeled datasets, training from 2D optical flow alone does not resolve degenerate motion cases. Other recent work focuses on leveraging vision transformers for generic object discovery \cite{xie2019object, bao2022discovering, bao2023object, singh2022simple, choudhury2022guess, elsayed2022savi++, karazija2022unsupervised, singh2022simple}. They focus either on unsupervised motion segmentation, video segmentation or generic object feature learning, where motion segmentation potentially acts as input \cite{bao2023object}. Their training objectives are not aligned with the presented task definition of \cite{dave2019towards, neoral2021monocular}, where incomplete motion patterns should result in complete semantic object instances. Therefore, we focus on supervised motion segmentation in this work.
\begin{figure*}[h!]
	\centering
	\begin{overpic}[width=0.9\linewidth, height=5.75cm, tics=0, clip]
		{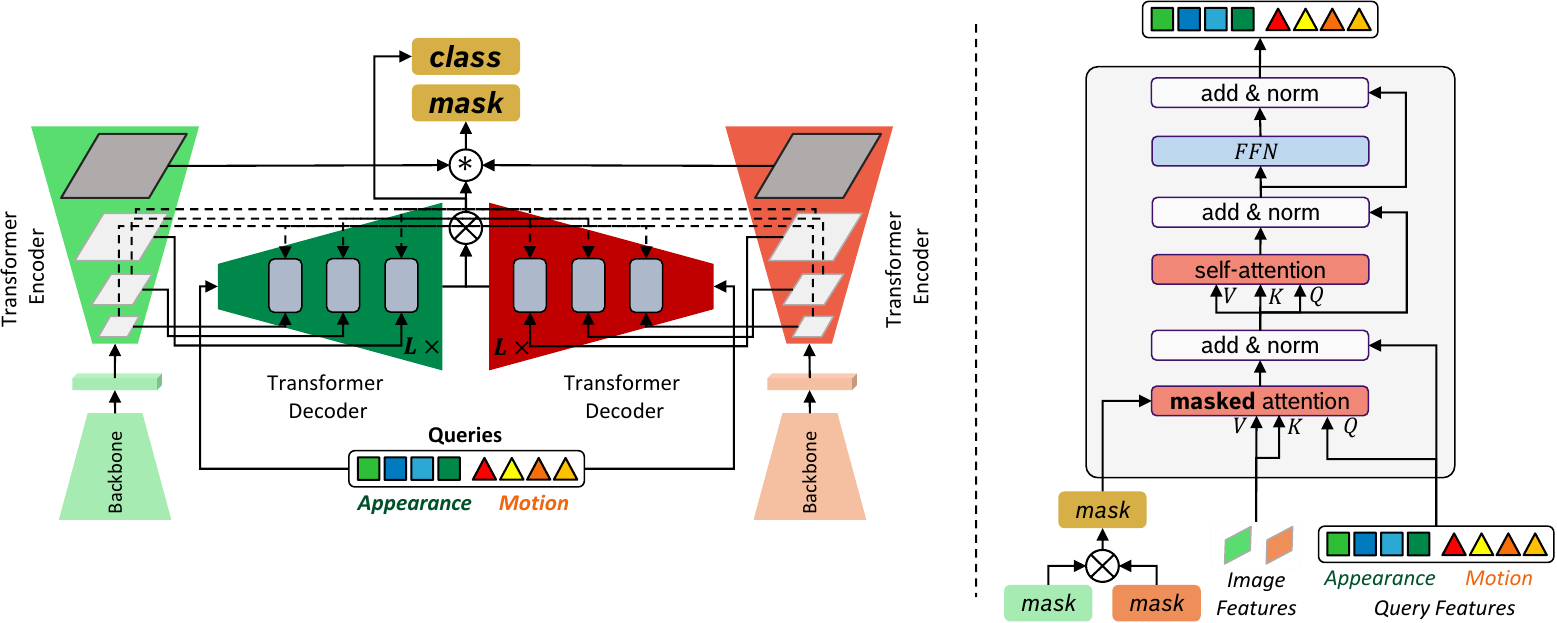}		
	\end{overpic}
	\caption{The $ \textbf{M}^{3} $Former architecture. Using two multi-modal streams, we fuse separate image and motion features across the streams. For each stream, we apply a backbone to learn multi-scale features $z$. We have two separate sets of query embeddings, i.e motion and appearance. Both multi-scale features $z$ and query embeddings $ q $ interact with each other through the attention mechanism.}
	\label{fig:architecture}
\end{figure*}
Many older approaches have focused on a binary foreground/background separation \cite{tokmakov2017learning, dutt2017fusionseg, lv2018learning}, which would require additional post-processing in order to detect individual objects. 
Another line of work utilizes binary motion segmentation as an auxiliary task for monocular scene reconstruction \cite{vijayanarasimhan2017sfm, wulff2017optical, yang2018every, ranjan2019competitive, zhao2022particlesfm, liu2023robust}. While this achieves promising results, it showcases the chicken-and-egg nature of the problem: In order to reconstruct video, we would like to separate the scene into dynamic foreground and static background beforehand. On the flipside, we need accurate 3D motion fields to infer this assignment in retrospective. In this work, we use a generic top-down approach to learn instance segmentation in an end-to-end manner similar to \cite{dave2019towards, yang2021learning, neoral2021monocular}. Video Instances Segmentation (VIS) \cite{yang2019video} is a highly active research topic for video data \cite{heo2022vita}. While \cite{dave2019towards, xie2019object} extend their model to an online-tracker, we focus on instance segmentation in this work. Extending motion segmentation to VIS would be an exciting avenue for future research.

\textbf{Multi-modal Fusion.}
Advances in motion segmentation are closely related to instance segmentation. Recent detectors \cite{cheng2021mask2former, zou2023segment, kirillov2023segment} achieve strong performance due to training on large, standard datasets such as COCO \cite{lin2014microsoft} and leveraging newer transformer architectures \cite{carion2020end, zhu2020deformable, cheng2021mask2former}. Pure image based detectors are limited to a fixed number of object labels in the training set. In the same manner, large amounts of data would be needed in order to train a robust motion detector based on image data alone. This can be alleviated by leveraging inductive biases from explicit motion estimates. Similar to \cite{dave2019towards, xie2019object, neoral2021monocular} we aim to generalize to arbitrary object categories by fusing both appearance and motion information. Motion segmentation is therefore closely related to multi-modal fusion. 
Since we only have monocular video as input, we create non-rgb \textit{pseudo-modalities} with off-the-shelf expert networks for optical flow \cite{teed2020raft}, depth \cite{ranftl2021vision} or scene flow \cite{teed2021raft}. This approach shares similarities to multi-modal vision expert models \cite{liu2023prismer} or recent multi-modal segmentation transformer \cite{zou2023segment, kirillov2023segment}.

Prior work focused on fusion with CNN-architectures \cite{tokmakov2017learning, dutt2017fusionseg, xie2019object, dave2019towards, neoral2021monocular}. Fusing with convolutional layers has the downside that both modalities/features will influence each other in a fixed manner. 
This inflexibility can worsen performance when the information from one modality is corrupted. Furthermore, switching between modalities or extending the architecture from image data to video data cannot be done in a CNN without retraining. We adress these issues by using a transformer with a two-stream architecture consisting of an appearance and motion branch.
Similar to \cite{mohamed2021modetr, zhou2020motion}, we fuse features flexibly based on attention \cite{vaswani2017attention}. However, instead of using a shared decoder we fuse features at multiple locations in the network.  
Compared to prior work, we further fuse multi-scale features in order to achieve higher-resolution masks instead of single-scale features. Finally, our work is also closely related to \cite{nagrani2021attention}, in the sense that we analyze the effect of different fusion mechanisms on downstream task performance. However, instead of focusing on audio-visual classification, we perform motion segmentation.

\section{Our Approach}
We introduce the $\textbf{M}^{3}$Former architecture for this task as is illustrated in Figure \ref{fig:architecture}. The main idea of our approach is to flexibly fuse multi-scale features from both appearance and motion data with attention.
\label{sec:approach}

\subsection{Motion Representations}
\label{sec:motion}
While previous work has explored the use of optical flow \cite{dave2019towards, yang2021self, xie2022segmenting} and higher level rigid motion costs \cite{yang2021learning, neoral2021monocular}, a detailed comparison of different motion representations for a single architecture has not been conducted. 
We progressively explore segmentation performance depending on the motion representation as input data. We analyze both the performance of single-modality inference and fusion with appearance features. 
Given two images $ I_{1},\; I_{2} \in \mathbb{R}^{H\times W \times 3} $, we are interested in the motion $ F_{1\mapsto 2} $ between both frames.

\textbf{Optical Flow.} Optical flow is a 2D translation field $ F \in \mathbb{R}^{H\times W\times 2} $. We use RAFT \cite{teed2020raft} in our work and take a robust version provided by \cite{neoral2021monocular}.

\textbf{Higher-dimensional Motion Costs.}
Optical flow is a 2D projection of the actual 3D motion. Multiple motions can map to the same projection, therefore the reconstruction is ambiguous. Reconstructing object and camera motion from optical flow has multiple degenerate cases \cite{yang2021learning}. Degenerate cases appear commonly in applications, e.g. all vehicles on a road drive colinear.
In order to detect moving objects robustly, we need some form of 3D prior indepent from Structure-from-Motion. The authors of \cite{yang2021learning} formulate four handcrafted criteria for computing a higher dimensinal cost function $ C_{12} \in \mathbb{R}^{H\times W \times 14} $ between two frames. This cost function has a higher cost in regions, that violate the static scene assumption. 
Computation involves estimating optical flow \cite{teed2020raft}, optical expansion \cite{yang2020upgrading}, camera motion \cite{hartley1997defense} and monocular depth \cite{ranftl2021vision}. The authors of \cite{neoral2021monocular} extend this cost function to a three-frame formulation $ C_{13} \in \mathbb{R}^{H \times W \times 28} $ by using backward $ F_{2\mapsto 1} $ and forward motion $ F_{2\mapsto 3} $. The computation of this cost embedding involves up to four neural networks, each trained on their own specific datasets.  

\textbf{Scene Flow.}
There exists a simpler minimal formulation - 3D scene flow. Given two RGBD frames $ \{ I_{1},\; Z_{1} \} $ and $ \{ I_{2},\; Z_{2} \} $, we compute motion as a field of rigid body transformations $ F \in \mathbb{R}^{H\times W\times 6} \in SE_{3} $. RAFT-3D \cite{teed2021raft} is the direct 3D equivalent of the 2D optical flow network \cite{teed2020raft} and naturally includes a geometric optimization. The main idea of this work is to compute a motion $ g \in SE_{3} $ for each pixel without making any assumption about semantics. Pixels naturally group together into semantically meaningful objects due to moving with the same rigid body motion. We spin this idea around - given multiple rigid body motions in a scene we want to infer an instance segmentation. 
While there are many diverse datasets for optical flow training \cite{baker2011database, ranjan2017optical, butler2012naturalistic, geiger2013vision}, there are fewer datasets for scene flow training \cite{mayer2016large}. We found, that existing model weights do not transfer well to all of our training datasets. We therefore finetune RAFT-3D for our training data, but use published checkpoints \cite{teed2021raft} during the evaluation. Performance of 3D motion estimation is largely dependent on the depth map quality. Training is done mostly with high-quality or ground truth depth. During inference on in-the-wild data, we do not have access to accurate absolute scale monocular depth for both $ Z_{1},\; Z_{2} $. We ablate the performance of motion estimation and segmentation depending on the depth quality. 
 
\subsection{Fusion}
\label{sec:fusion}
Image based detectors can solve the segmentation and detection task well, but perform poorly on motion classification. Simply using monocular video data for motion segmentation is a challenging task to learn with limited training data. 
The task gets solvable when using motion as an intermediate data representation, which acts as inductive biases. However, in order to robustly segment semantically meaningful moving objects, combining both image and motion data together is crucial. The motion segmentation task therefore can be considered a \textit{multi-modal fusion} problem.

Transformers are very flexible - Adapting a transformer for example to Video Instance Segmentation only requires a change in Positional Encoding and little finetuning \cite{cheng2021mask2former}. This flexibility is a key advantage, since it leaves the possibility open to use longer temporal windows in the future. In a similar manner, we add a modality specific positional encoding and combine data from multiple modalities instead of temporal frames. When using multiple modalities, we combine features within a two-stream architecture with dedicated parameters $ \Theta_{rgb},\; \Theta_{motion} $. Each branch is trained on it's own modality individually first and then fusion is learned by finetuning both branches together. We experiment with multiple methods for fusing information at different locations. We base our different streams on the SotA segmentation architecture Mask2Former \cite{cheng2022masked}. 

\textbf{Multi-headed Attention.}
A transformer layer consists of Multi-Headed Self-Attention (MSA) \cite{vaswani2017attention}, Layer Normalisation (LN) and Multilayer Perceptron (MLP) blocks, applied using residual connections. Given input 
tokens $ z^{l} $ at layer $ l $, we have
\begin{align}
\label{eq:mha}
	y^{l} &= MSA \left( LN\left( z^{l} \right)\right) + z^{l} \\
	z^{l+1} &= MLP\left( LN\left( y^{l} \right)\right) + y^{l} \qquad .
\end{align}
The MSA operation computes dot-product attention \cite{vaswani2017attention}, where query, key and values are linear projections of the same input tensor: $ MSA \left( \mathbf{X} \right) = Attention\left( \mathbf{W}^{Q}\mathbf{X},\; \mathbf{W}^{K}\mathbf{X},\; \mathbf{W}^{V}\mathbf{X}\right)$. Multi-Headed Cross Attention (MCA) computes attention between two input tensors $ \mathbf{X} $ and $ \mathbf{Y} $, where $ \mathbf{X} $ acts as the query and $ \mathbf{Y} $ as keys and values: 
$ MCA\left( \mathbf{X},\; \mathbf{Y}\right) = Attention \left( \mathbf{W}^{Q}\mathbf{X},\; \mathbf{W}^{K}\mathbf{Y},\; \mathbf{W}^{V}\mathbf{Y}\right)$. 
Fusion in a vision transformer architecture is simple: Given two separate token sequences $ z_{rgb} $ and $ z_{motion} $, we can generate a longer sequence $ z = \left[ z_{rgb} || z_{motion} \right] $ by concatenation. Running this longer sequence through the transformer layer lets both modalities exchange information.
We have both \textit{self-attention} and \textit{cross-attention} layers with a learned attention mask $ \mathbf{M}^{l-1} $ \cite{cheng2021per} in the decoder. Since it is a query based detector, we not only have high-resolution spatial input feature tokens $ z $ (see Figure \ref{fig:architecture}), but also 256-dimensional object query embeddings $ q $. Masked cross-attention is computed between $ z $ and $ q $, while self-attention is performed only on $ q $ to learn global context. 
We have two sets of object embeddings: \textit{appearance} $ q_{rgb} $ and \textit{motion} $ q_{motion} $. We concatenate spatial features $ \left[ z_{rgb}|| z_{motion} \right] $, object query embeddings $ \left[ q_{rgb} || q_{motion} \right] $ and the respective attention masks $ \left[ \mathbf{M}_{rgb} || \mathbf{M}_{motion} \right] $ as can be seen in Figure \ref{fig:architecture} on the right. Attention can flow freely through the network with the learned masks, 
\begin{figure}[h!]
	\centering
	\begin{overpic}[width=0.7\linewidth, height=4cm, tics=0, clip]
		{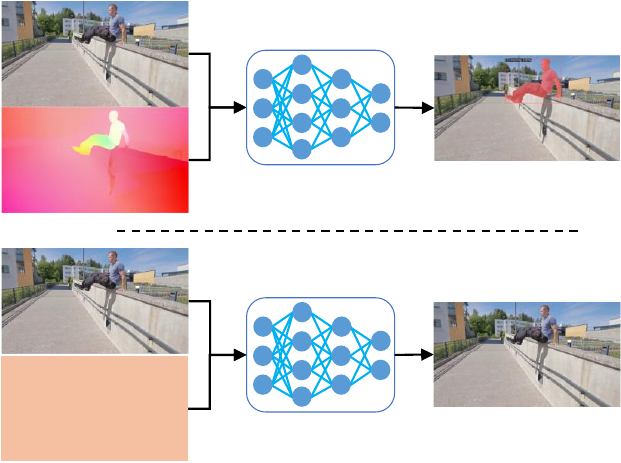}
		\put(-22, 43){\footnotesize{Flow Field}}
		\put(-17, 9){\footnotesize{\textit{Const.}}}
		\put(-22, 5){\footnotesize{Flow Field}}
		
		\put(105, 54){\footnotesize{\textit{Positive}}}
		\put(105, 50){\footnotesize{Example}}
		
		\put(105, 16){\footnotesize{\textit{Negative}}}
		\put(105, 12){\footnotesize{Example}}
		
	\end{overpic}
	\caption{Alignment from regularization: We force the model to not overrely on appearance data by introducing neg. examples.}
	\label{fig:augmentation}
\end{figure}
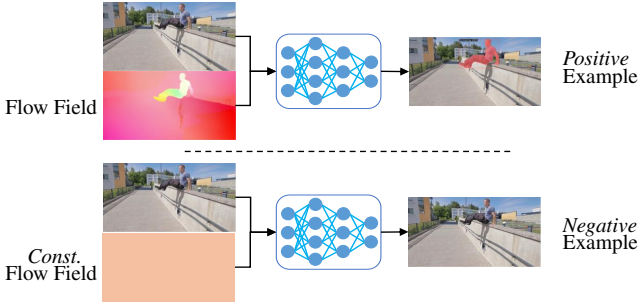
i.e. all \textit{important} image tokens can interact with motion tokens and all queries. A final prediction is made by combining individual outputs with a single convolutional layer. 

\textbf{Attention Bottlenecks.}
Pairwise attention has quadratic complexity, which can be critical. In order to tame this complexity, the authors of \cite{nagrani2021attention} proposed fusion bottleneck tokens. We found that in practice memory only becomes a problem when using many queries. In the same manner, we experimented with bottleneck object embeddings $ q_{mbt} $ and let both branches interact only through these bottlenecks. 

\textbf{Deformable Attention.} The encoder of the Mask2-Former architecture uses multi-scale deformable attention \cite{zhu2020deformable}. This is a mechanism for sampling only few interesting spatial locations from input features maps. The pairwise attention is thus limited to a reduced set and has linear complexity w.r.t the spatial input size. For fusion, we can simply concatenate both feature maps along the x-axis, such that $ f = \left[ f_{rgb},\; f_{motion} \right] $ since both input modalities share the same spatial dimensions. When fusing at the encoder level, we perform this operation on each scale of the feature pyramid and add a modality specific positional encoding similar to \cite{cheng2021mask2former}. 

\begin{table*}[b]
	\centering
	\resizebox{\textwidth}{!}{%
		\begin{tabular}{lccccccc}
			\toprule
			\textbf{Dataset} & \textbf{Groundtruth data} & \textbf{Diversity Motion} & \textbf{Diversity Classes} & \textbf{non-rigid motion} & \textbf{degenerate cases} & \textbf{\#Train} & \textbf{\#Test} \\
			\midrule
			FlyingTings3D & Depth, 2D/3D Motion, Odometry & High & High & \xmark & \xmark & 40 100 & 7800 \\
			Monkaa & Depth, 2D/3D Motion, Odometry & Medium & Medium & (\cmark) & \xmark & 23 356 & 2588 \\
			Driving & Depth, 2D/3D Motion, Odometry & Low & Low & \xmark & \cmark & 9954 & 1106 \\
			Davis & - & Medium & High & \cmark & \xmark & 2232 & 1620 \\
			Kitti & Lidar, 2D/3D Motion, Odometry & Low & Low & \xmark & \cmark & 180 & 20 \\
			Virtual Kitti & Depth, 2D/3D Motion, Odometry & Low & Low & \xmark & \cmark & 29 811 & 3314 \\
			\bottomrule
	\end{tabular}}
	\caption{Motion segmentation datasets. Available datasets have different motion patterns and moving semantic classes. 
	}
	\label{tab:datasets}
\end{table*}

\textbf{Multi-modal Alignment.}
Alignment between modalities is a vast and important topic in multi-modal models, we refer the reader to \cite{baltruvsaitis2018multimodal} for a more extensive overview. 
Both modalities might not contain the same amount of information and models need to be able to flexibly decide which data source to trust. 
In our case motion maps might be noisy and the model needs to figure out to rely on the 
\begin{table}[H]
	\resizebox{0.5\textwidth}{!}{%
		\begin{tabular}{lccccccc}
			& \textbf{FT3D} & \textbf{Monkaa} & \textbf{Driving} & \textbf{Vkitti} & \cellcolor{lighter-gray} \textbf{Kitti} & \cellcolor{lighter-gray} \textbf{Davis} & \textbf{YTVOS} \\
			Mix 0 & \cmark & \xmark & \xmark & \xmark & \cellcolor{lighter-gray} \xmark & \cellcolor{lighter-gray} \xmark & \xmark\\
			Mix 1 \cite{yang2021learning, neoral2021monocular} & \cmark & \cmark & \cmark & \xmark & \cellcolor{lighter-gray} \xmark & \cellcolor{lighter-gray} \xmark & \xmark \\
			Mix 2 & \cmark & \xmark & \xmark & \cmark & \cellcolor{lighter-gray} \xmark & \cellcolor{lighter-gray} \cmark & \xmark \\
			Mix 3 & \cmark & \cmark & \cmark & \cmark & \cellcolor{lighter-gray} \cmark & \cellcolor{lighter-gray} \cmark & \xmark \\
			\midrule
			Mix 4 \cite{dave2019towards} & \cmark & \xmark & \xmark & \xmark & \cellcolor{lighter-gray} \xmark & \cellcolor{lighter-gray} \cmark & \cmark \\
			\bottomrule
	\end{tabular}}
	\caption{We experiment with different dataset mixes. Colored datasets are used for evaluation. Single-modality models are trained on Mix 0, Fusion models on Mix 1 -3.
		Mix 1 is a common setting proposed by \cite{yang2021learning, neoral2021monocular} to test generalization. Because this setting lacks diversity in semantic classes and motion patterns, we propose more appropriate mixes that resolve common failure cases.}
	\label{tab:mixes}
\end{table}
appearance information for high segmentation quality. At the same time datasets \cite{lamdouar2020betrayed} exist, where motion can act as a stronger cue for discovering moving objects.

We notice in our experiments in Section \ref{seq:experiments}, that models usually overrely on appearance data for motion segmentation and thus introduce many false positives. This issue is especially present for ill-posed 2D motion representations. 
We thus propose a very simple augmentation strategy as can be seen in Figure \ref{fig:augmentation}: With a given probability $ p_{neg} $, we introduce negative examples, where motion data is augmented to a random constant flow field within value range. Without any variation in the motion data, models should place semantic objects from the appearance stream into the background. We experiment with multiple values for $ p_{neg} $.

\section{Experiments}
\label{seq:experiments}
In our experiments we want to answer the following research questions: 
\begin{itemize}
	\item[] {\color{green!55!blue} What motion representation is most useful for motion segmentation? How important is fusion with appearance data?}
\end{itemize}
We use a vanilla Mask2Former \cite{cheng2022masked} model for single-modality training. All experiments are done with a ResNet50 \cite{he2016deep} backbone, so that we are comparable to related approaches. Scaling the network is not focus of this paper, but would be a promising direction for future work.

\subsection{Datasets}
\label{sec:data}

The authors of \cite{dave2019towards, yang2021learning, neoral2021monocular} have made the effort to create motion labels on multiple datasets. Table \ref{tab:datasets} shows used motion segmentation datasets and their characteristics.
We use common datasets: Sceneflow \cite{mayer2016large}, KITTI \cite{geiger2013vision}, Virtual Kitti \cite{cabon2020virtual} and Davis \cite{pont20172017}. Scenes range from autonomous driving, random synthetic scenes to real world casual videos with humans and animals. Table \ref{tab:mixes} shows different training data mixes from the literature and our experiments. Related work \cite{yang2021learning, neoral2021monocular} train their fusion models solely on the SceneFlow datasets and evaluate generalization on Davis, Kitti and YTVOS \cite{xu2018youtube}. We drop YTVOS, because performance heavily correlates to Davis. We keep this training setting for our fusion experiments. Single-modality motion segmentors are trained on FlyingThings3D. We note how common failure cases result due to a lack of diverse training data. Mix 1 does not contain many degenerate motion patterns and non-rigid moving objects. We therefore progressively diverge from this setting and analyze the effect of data on performance in Section \ref{sec:exp_fusion}. We balance individual datasets, such that samples are drawn with approx. equal likelihood during training, i.e. we use a naive sampling strategy. We believe this to be a step in the right direction, as large scale training is necessary for true real-world generalization abilities. 

\textbf{Metrics.}
We report standard instance segmentation COCO metrics such as $ mAP $, $ AP_{50} $. 
We further include other segmentation metrics, such as Precision (Pu), Recall (Ru) and F-score (Fu) \cite{dave2019towards}, 
foreground precision \cite{yang2021learning} and the number of false positives and false negatives over the whole split \cite{neoral2021monocular}. Since datasets come in different sizes, we normalize the number of false positives/negatives. In our ablations, we mainly focus on $ mAP $, $ FP $ and $ FN $, because they act as a good proxy. More details can be found in Suppl. Sec. \ref{sup:eval}.

\begin{table}[bp]
	\centering
	\resizebox{0.45\textwidth}{!}{%
		\begin{tabular}{lccc}
			\cellcolor{lighter-gray} \fns{Modality} & \cellcolor{lighter-gray} \fns{$ AP $} & \cellcolor{lighter-gray} \fns{$ AP_{50} $} & \cellcolor{lighter-gray} \fns{$ AP_{75} $}  \\
			\midrule
			\fns{RGB} & \fns{56.53} & \fns{76.71} & \fns{57.5} \\
			\midrule
			\fns{Scene Flow$^{\dagger} $} & \fns{\textbf{75.19}} & \fns{\textbf{89.52}} & \fns{\textbf{77.03}} \\
			\fns{Optical Flow$^{\dagger} $} & \fns{\underline{72.24}} & \fns{\underline{87.43}} & \fns{\underline{74.52}} \\
			\fns{Scene Flow \cite{teed2021raft}} & \fns{55.39} & \fns{75.31} & \fns{56.26} \\
			\fns{Motion embedding \cite{neoral2021monocular}} & \fns{53.30} & \fns{75.20} & \fns{54.9} \\
			\fns{Optical Flow \cite{teed2020raft}} & \fns{52.45} & \fns{72.75} & \fns{52.73} \\
			\bottomrule
	\end{tabular}}
	\caption{Comparison of different input data for motion segmentation on FlyingThings3D. $ ^{\dagger} $ denotes ground truth data.}
	\label{tab:ft3d}
\end{table}

\subsection{Modalities for Motion Segmentation}
In our first experiments, we focus on single modalities. We train for 30 epochs, for more details see Suppl. Sec. \ref{sup:implementation}. Table \ref{tab:ft3d} shows the results on the test split of FlyingThings3D. 
We achieve best results with 3D input data, which suggests that 3D motion 
makes the task easier for the network to learn and generally outperforms 2D motion.
The gap between predicted and groundtruth motion leaves room for improvement for off-the-shelf estimators. Interestingly, we include a pure image baseline model. We can train a strong image detector on this dataset, because foreground objects are consistently in motion and distinct from the background. Note how this would not be the case if the data contained object classes, which can move but don't. We will later see, how pure image baselines only perform favorably on metrics which do not punish false positives.

\subsection{Why One Modality Is Not Enough}
When generalizing to real-world data with a very different distribution of objects and motion patterns, single-modality models will perform much worse as can be seen in Table \ref{tab:single_mod}. For our pure image baseline, we use the COCO \cite{lin2014microsoft} pretrained model from \cite{cheng2022masked}. In order to create a stronger baseline, we only use classes, which can move on their own or are likely to be in motion, e.g. cars or persons (see more information in Suppl. Sec. \ref{sup:eval}). 
3D motion requires 3D geometry. Monocular depth prediction in dynamic environments is an open problem \cite{kopf2021robust, zhang2021consistent} and is challenging on in-the-wild data. During training we used perfect ground truth depths for computing the scene flow. On in-the-wild data this will not be the case. We ablate multiple scenarios for depth prediction quality. For autonomous driving data we compare the performance for rel. monocular depth, abs. monocular depth and stereo depth. For monocular depth prediction we take DPT \cite{ranftl2021vision} and UniMatch \cite{xu2022unifying} for stereo as two SotA single-timeframe models. We compute the abs. depth of each frame by aligning it with the groundtruth as \cite{yang2021learning}. 
\begin{table}[b!]
	\centering
	\setstackgap{L}{9pt}
	\resizebox{0.48\textwidth}{!}{%
		\begin{tabular}{lccccccc}
			\toprule
			& \multicolumn{3}{c}{\textbf{Kitti}} & \multicolumn{3}{c}{\textbf{Davis}} \\
			Modality & \cellcolor{Dandelion} $ AP_{50} \uparrow $ & \cellcolor{Dandelion} FP $ \downarrow $ & \cellcolor{Dandelion} FN $ \downarrow $ & \cellcolor{BlueGreen} $ AP_{50} \uparrow $ & \cellcolor{BlueGreen} FP $ \downarrow $ & \cellcolor{BlueGreen} FN $ \downarrow $ \\
			\midrule
			RGB (Coco) & \textbf{58.2} & 1.34 & \textbf{0.17} & \textbf{50.51} & 0.92 & \textbf{0.07} \\
			Optical Flow \cite{teed2020raft} & 25.1 & 0.99 & 0.43 & 30.2 & 0.63 & 0.13 \\
			\Centerstack{Scene Flow \cite{teed2021raft} \\ rel. scale} & 29.6 & 0.54 & 0.42 & 11.0 & \textbf{0.24} & 0.22 \\
			\Centerstack{Scene Flow \cite{teed2021raft} \\ abs. scale} & 36.8 & \underline{0.50} & 0.40  & \textcolor{gray}{39.84} & \textcolor{gray}{0.41} & \textcolor{gray}{0.14} \\
			\Centerstack{Scene Flow \cite{teed2021raft} \\ stereo} & \underline{44.4} & \textbf{0.10} & \underline{0.40} &  - & - & - \\
			\Centerstack{Motion \\ embedding \cite{neoral2021monocular}} & 28.9 & 0.59 & 0.43 & \underline{33.9} & \underline{0.57} & \underline{0.13} \\
			\bottomrule
		\end{tabular}
	}
	\caption{Zero-shot performance of single-modality models on KITTI and Davis. Results in grey are only for few selected videos, where a reconstruction with SfM is possible.}
	\label{tab:single_mod}
\end{table} 
Alignment is not possible on casual video clips like DAVIS without a reference. The reconstruction of casual videos is still an open research problem in itself \cite{liu2023robust}. However, we propose a simple strategy for depth alignment based on an end-to-end SLAM
\begin{table}[h!]
	\centering
	\resizebox{0.45\textwidth}{!}{%
		\begin{tabular}{ccccccc}
			\toprule
			& \multicolumn{3}{c}{\textbf{Kitti}} & \multicolumn{3}{c}{\textbf{Davis}} \\
			$ p_{neg} $ & \cellcolor{Dandelion} $ AP_{50} \uparrow $ & \cellcolor{Dandelion} FP $ \downarrow $ & \cellcolor{Dandelion} FN $ \downarrow $ & \cellcolor{BlueGreen} $ AP_{50} \uparrow $ & \cellcolor{BlueGreen} FP $ \downarrow $ & \cellcolor{BlueGreen} FN $ \downarrow $ \\
			\midrule
			0 & 15.55 & \textbf{14.42} & 107.8 & 19.846 & 897.90 & 263.85 \\
			30 & \textbf{37.16} & 23.38 & \textbf{83.19} & \textbf{23.98} & \textbf{616.53} & \textbf{258.81} \\
			\bottomrule
	\end{tabular}}
	\caption{Ablation of neg. examples augmentation (FP and FN are not normalized). Experiments were run on Mix 1 with image and optical flow data.}
	\label{tab:neg}
\end{table}
system \cite{teed2021droid}. This reconstruction is only possible on few selected video clips, but acts as a proof-of-concept. More information can be found in Suppl. Sec. \ref{sup:davis}.

\textbf{Motion Data Is Not Equally Useful.}
It can be seen in Table \ref{tab:single_mod}, that motion representations can provide different value depending on the dataset. While optical flow is a generic motion representation, which can be inferred reliably on most datasets, 3D scene flow is heavily dependent on the depth quality. The motion embeddings from \cite{neoral2021monocular} offer a great trade-off since they do not require multiple scale-correct depth maps, but still contain 3D costs. 
Once depth is reliably provided, high quality scene flow gives the best results as can be seen on Kitti. However, there remains a large gap in mAP to the image baseline. Reasons for this gap are multiplefold: 
i) Davis contains many non-rigid motion patterns. Since this has not been in the training data, the model did not learn to group motions and oversegments the scene. 
ii) 3D geometry cannot be reliably reconstructed, therefore 3D motion is very noisy. 
iii) Kitti has fast camera motion and many objects move colinear to it. At the same time most scenes contain both static and moving objects of the same class, which is in contrast to training data. Thus, Optical Flow based detectors have many false positives. 3D motion based detectors are dependent on the depth quality. 
iv) Often multiple objects share the same forward motion, therefore they are grouped together and the scene is undersegmented. These cases are not present in dataset mix 0 and 1.
On the other hand, a pure image detector will detect any semantic object and introduce many false positives.

\subsection{Fusion Between Appearance and Motion Data}
\label{sec:exp_fusion}
In order to create robust motion segmentation, we resolve the before mentioned problems by fusing appearance and motion information. Since we want to \textit{retain} semantic object knowledge of an image detector, we freeze the image branch that is pretrained on COCO similar to previous work \cite{dave2019towards, neoral2021monocular}. We take the pretrained motion branches on FT3D and finetune a fusion model on the respective data mix. Training and implementation details can be found in Supp. Sec \ref{sup:implementation}.
Alignment between appearance and motion features is very important. The model should not rely too much on appearance to overrule the classification from motion. In Table \ref{tab:neg} we show the effect of introducing neg. examples. 
\begin{table}[h!]
	\setstackgap{L}{8pt}
	\centering
	\resizebox{0.47\textwidth}{!}{%
		\begin{tabular}{lcccccccc}
			\toprule
			& & \multicolumn{3}{c}{\textbf{Kitti}} & \multicolumn{3}{c}{\textbf{Davis}} \\
			Data & Modality & \Centerstack[c]{Fusion\\mechanism} & \cellcolor{Dandelion} $ AP_{50} \uparrow $ & \cellcolor{Dandelion} FP $ \downarrow $ & \cellcolor{Dandelion} FN $ \downarrow $ & \cellcolor{BlueGreen} $ AP_{50} \uparrow $ & \cellcolor{BlueGreen} FP $ \downarrow $ & \cellcolor{BlueGreen} FN $ \downarrow $ \\
			\midrule
			& & \cellcolor{light-gray} D & \cellcolor{light-gray} 37.16 & \cellcolor{light-gray} 0.12 & \cellcolor{light-gray} 0.41 & \cellcolor{light-gray} 23.98 & \cellcolor{light-gray} 0.39 & \cellcolor{light-gray} 0.16  \\
			& \multirow{-2}{*}{RGB + OF} & \cellcolor{lighter-gray} E+D & \cellcolor{lighter-gray} 39.65 & \cellcolor{lighter-gray} 0.15 & \cellcolor{lighter-gray} 0.40 & \cellcolor{lighter-gray} 35.88 & \cellcolor{lighter-gray} 0.29 & \cellcolor{lighter-gray} 0.15 \\
			& & D & 27.5 & 0.05 & 0.50 & 19.25 & 0.30 & 0.19 \\
			& \multirow{-2}{*}{RGB + SF} & E+D & 26.5 & 0.10 & 0.49 & 15.4 & 0.11 & 0.23 \\
			& & \cellcolor{light-gray} D & \cellcolor{light-gray} 27.6 & \cellcolor{light-gray} \textbf{0.05} & \cellcolor{light-gray} 0.50 & \cellcolor{light-gray} \textcolor{gray}{21.8} & \cellcolor{light-gray} \textcolor{gray}{0.38} & \cellcolor{light-gray} \textcolor{gray}{0.14} \\
			& \multirow{-2}{*}{RGB + SF*} & \cellcolor{lighter-gray} E+D & \cellcolor{lighter-gray} 37.8 & \cellcolor{lighter-gray} 0.33 & \cellcolor{lighter-gray} 0.38 & \cellcolor{lighter-gray} \textcolor{gray}{38.4} & \cellcolor{lighter-gray} \textcolor{gray}{0.13} & \cellcolor{lighter-gray} \textcolor{gray}{0.17} \\
			& & D & 27.6 & \textbf{0.05} & 0.50 & - & - & - \\
			& \multirow{-2}{*}{RGB + SF**} & E+D & 51.0 & 0.10 & 0.35 & - & - & - \\
			& & \cellcolor{light-gray} D & \cellcolor{light-gray} 29.51 & \cellcolor{light-gray} \underline{0.06} & \cellcolor{light-gray} 0.47 & \cellcolor{light-gray} 27.47 & \cellcolor{light-gray} 0.30 & \cellcolor{light-gray} 0.17 \\
			\multirow{-10}{*}{Mix 1} & \multirow{-2}{*}{RGB + Cost} & \cellcolor{lighter-gray} E+D & \cellcolor{lighter-gray} 47.9 & \cellcolor{lighter-gray} 0.22 & \cellcolor{lighter-gray} 0.35 & \cellcolor{lighter-gray} 33.27 & \cellcolor{lighter-gray} 0.49 & \cellcolor{lighter-gray} 0.14 \\
			\midrule
			& \multirow{2}{*}{RGB + OF} & D & \underline{70.82} & 0.32 & \underline{0.16} & 54.01 & 0.13 & \textbf{0.01} \\
			& & E+D & 60.88 & 0.29 & 0.24 & \underline{61.12} & 0.11 & 0.12 \\
			& & \cellcolor{light-gray} D & \cellcolor{light-gray} \textbf{72.07} & \cellcolor{light-gray} 0.31 & \cellcolor{light-gray} \textbf{0.16} & \cellcolor{light-gray} 58.76 & \cellcolor{light-gray} 0.11 & \cellcolor{light-gray} 0.13 \\
			& \multirow{-2}{*}{RGB + SF}  & \cellcolor{lighter-gray} E+D & \cellcolor{lighter-gray} 56.3 & \cellcolor{lighter-gray} 0.38 & \cellcolor{lighter-gray} 0.27 & \cellcolor{lighter-gray} 53.9 & \cellcolor{lighter-gray} \textbf{0.08} & \cellcolor{lighter-gray} 0.15 \\
			& & D & 68.46 & 0.26 & 0.21 & 58.40 & \underline{0.09} & 0.13 \\
			\multirow{-6}{*}{Mix 3} & \multirow{-2}{*}{RGB + Cost} & E+D & 65.12 & 0.22 & 0.23 & \textbf{64.10} & 0.11 & \underline{0.12} \\
			\bottomrule
	\end{tabular}}
	\caption{Fusion of appearance with different modalities. * denotes abs. scale depth ** denotes stereo depth}	
	\label{tab:ablation_modality}
\end{table}

As can be seen, this simple augmentation can stop the model to rely too much on appearance data and reduces false positives (we show the total number of false positives/negatives over the whole split). 
On Kitti the number of false positives is harder to reduce, because both positive and negative examples of moving objects are hidden inside a flow field with large variance due to the fast driving motion.
We keep 30\% neg. examples as augmentation in future experiments. When scaling up to larger dataset mix 3, we set $ p_{neg} = 5\% $ in order to reduce training time as a trade-off.

We can choose multiple fusion strategies in our two-stream architecture:
i) deformable Attention in Encoder (E). 
ii) Vanilla attention in Decoder (D).
iii) Multi-modal Bottleneck Tokens (MBT) \cite{nagrani2021attention} in Decoder.
iv) Fusion in both Encoder and Decoder (E+D).
We ablate these strategies in Table \ref{tab:ablation_fusion} in Suppl. Sec. \ref{sup:fusion_ablation}. We found, that there is no optimal strategy for all motion representations and training data. We observed, that the training dynamics are affected by the fusion mechanism and hypothesize, that the strategies can potentially converge to similar results when given enough training time. Finally, there is no optimal strategy for both Kitti and Davis. We therefore opted for the simple late fusion in the decoder or fusion in both encoder and decoder for later experiments. Our results in Table \ref{tab:ablation_modality} show, that motion cues generally reduce false positives and the fusion with appearance data closes the gap in precision. 3D motion representations can give stronger performance when they are available in high quality.

\textbf{Beyond Small-scale Datasets.}
Our previous experiments have shown that a simple detector baseline is hard to beat for segmentation precision. While image data is very valuable for precision, motion data helps in reducing false positive detections as can be seen in Table \ref{tab:single_mod}. Motion and Fusion models over- or undersegment the scene due to a lack of diverse training data. As can be seen in Table \ref{tab:sota} we could not replicate the performance of \cite{neoral2021monocular} with the training setting of Mix 1 \cite{neoral2021monocular, yang2021learning}. 
\begin{table*}[h!]
	\centering
	\resizebox{1.0\textwidth}{!}{%
		\begin{tabular}{lcccccccccccccccccc}
			\toprule
			& & \multicolumn{8}{c}{\textbf{Kitti}} & \multicolumn{8}{c}{\textbf{Davis}} \\
			Method & \Centerstack[c]{Training\\data} & \cellcolor{Dandelion} $ AP \uparrow $ & \cellcolor{Dandelion} bg & \cellcolor{Dandelion} obj & \cellcolor{Dandelion} Pu & \cellcolor{Dandelion} Ru & \cellcolor{Dandelion} Fu & \cellcolor{Dandelion} FP $ \downarrow $ & \cellcolor{Dandelion} FN $ \downarrow $ & \cellcolor{BlueGreen} $ AP \uparrow $ & \cellcolor{BlueGreen} bg & \cellcolor{BlueGreen} obj & \cellcolor{BlueGreen} Pu & \cellcolor{BlueGreen} Ru & \cellcolor{BlueGreen} Fu & \cellcolor{BlueGreen} FP $ \downarrow $ & \cellcolor{BlueGreen} FN $ \downarrow $ \\
			\midrule
			RGB sem. baseline \cite{cheng2021mask2former} & COCO & 42.2 & 96.6 & 69.25 & 60.70 & \underline{93.96} & 69.25 & 1.34 & \underline{0.17} & 35.05 & 0.92 & 0.68 & 0.61 & \textbf{0.88} & 0.68 & 0.92 & \underline{0.07} \\
			Learning rigid motions \cite{yang2021learning} * & Mix 1 & 20.0 & - & - & - & - & - & - & - & 4.2 & - & - & - & - & - & - & - \\
			Generic MoSeg\cite{dave2019towards} & Mix 4 & 20.0 & - & - & - & - & - & - & - & 20.8 & - & - & - & - & - & - & - \\
			Raptor \cite{neoral2021monocular} & Mix 1 & 40.07 & \underline{98.97} & \underline{86.3} & \textbf{89.37} & 86.3 & \underline{86.3} & \textbf{0.11} & 0.35 & \underline{40.9} & 94.20 & 73.3 & 71.57 & 80.20 & 73.3 & 0.25 & 0.10 \\
			\midrule
			Ours RGB + OF &  & 25.67 & 98.46 & 76.63 & 81.12 & 76.70 & 76.63 & 0.15 & 0.40 & 19.28 & \textbf{95.34} & 64.85 & 67.32 & 67.82 & 64.85 & 0.29 & 0.15 \\
			Ours RGB + SF* & Mix 1 & 26.70 & 96.65 & 64.13 & 62.71 & 77.08 & 64.13 & 0.33 & 0.38 & \textcolor{gray}{15.20} & \textcolor{gray}{96.50} & \textcolor{gray}{69.45} & \textcolor{gray}{71.07} & \textcolor{gray}{69.70} & \textcolor{gray}{69.45} & \textcolor{gray}{0.13} & \textcolor{gray}{0.17} \\
			Ours RGB + Cost \cite{neoral2021monocular} & & 32.40 & 98.65 & 79.39 & 78.47 & 84.87 & 79.39 & \underline{0.22} & 0.35 & 16.85 & 93.70 & 59.78 & 62.39 & 64.67 & 59.78 & 0.50 & 0.14 \\
			\midrule
			Ours RGB + OF  & Mix 2 & 40.08 & 97.83 & 66.84 & 74.59 & 68.10 & 66.84 & 0.13 & 0.35 & \textbf{43.52} & 92.97 & \textbf{76.72} & \underline{76.62} & \underline{83.48} & \textbf{76.72} & 0.25 & 0.09 \\ 
			Ours RGB + OF & Mix 3 & \underline{50.91} & \textbf{99.19} & 85.99 & 83.26 & 91.30 & 85.99 & 0.32 & 0.16 & 32.25 & 94.53 & 73.73 & 73.12 & 77.23 & 73.73 & \underline{0.12} & \textbf{0.01} \\
			Ours RGB + SF & Mix 3 & \textbf{52.27} & 98.89 & \textbf{87.05} & \underline{84.18} & \textbf{93.99} & \textbf{87.05} & 0.31 & \textbf{0.16} & 37.07 & \underline{95.17} & \underline{76.21} & \textbf{77.24} & 77.70 & \underline{76.21} & 0.11 & 0.13 \\
			Ours RGB + Cost \cite{neoral2021monocular} & Mix 3 & 48.44 & 98.76 & 82.33 & 80.84 & 88.13 & 82.33 & 0.26 & 0.21 & 35.11 & 95.13 & 75.87 & 75.58 & 78.51 & 75.87 & \textbf{0.09} & 0.13 \\
			\bottomrule
	\end{tabular}}
	\caption{SotA Motion Segmentation on Kitti and Davis. We report our best results for the respective modality and data. Results in grey are on scenes, where a reconstruction with SfM is possible. 
		*use of abs. scale information}
	\label{tab:sota}
\end{table*}
\begin{figure*}[h!]
	\centering
	\begin{overpic}[width=0.15\linewidth, height=1.0cm, tics=0, clip]
		{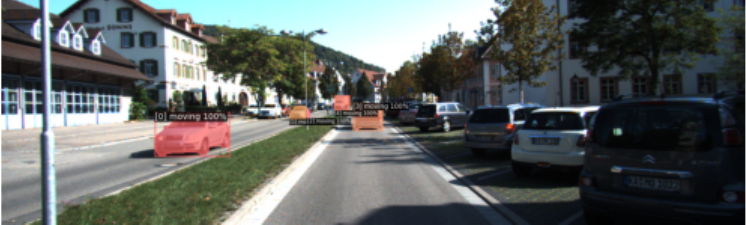}
		\put(-17, 18){\rotatebox{90}{\footnotesize{GT}}}
	\end{overpic}\hspace*{-0.3em}
	\begin{overpic}[width=0.15\linewidth, height=1.0cm, tics=0, clip]
		{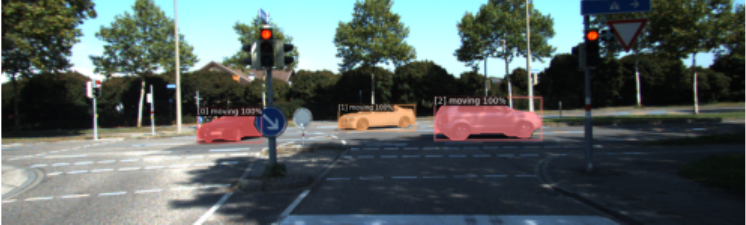}
	\end{overpic}\hspace*{-0.3em}
	\begin{overpic}[width=0.15\linewidth, height=1.0cm, tics=0, clip]
		{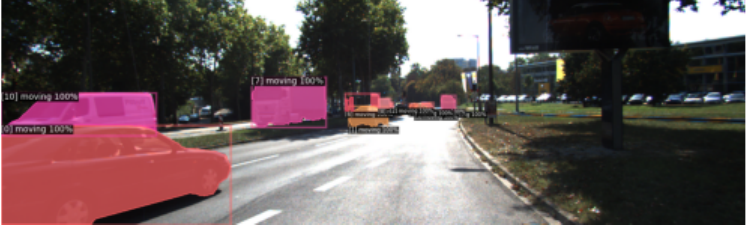}
	\end{overpic}
	\begin{overpic}[width=0.15\linewidth, height=1.0cm, tics=0, clip]
		{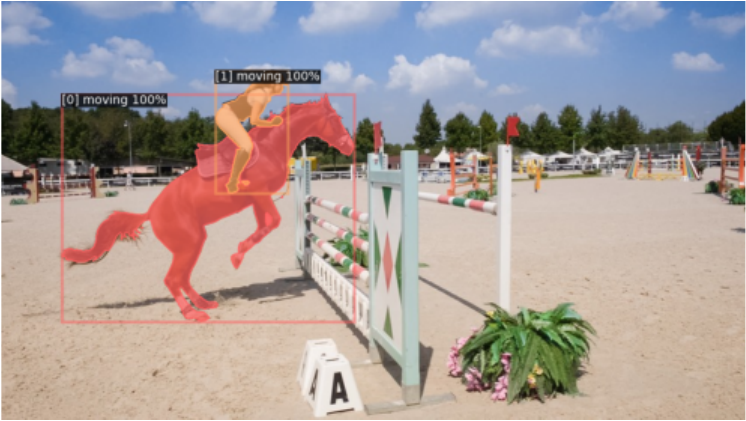}
	\end{overpic}\hspace*{-0.3em}
	\begin{overpic}[width=0.15\linewidth, height=1.0cm, tics=0, clip]
		{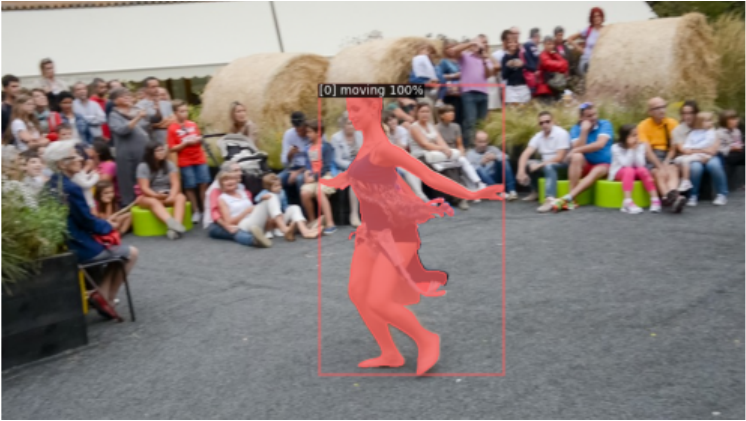}
	\end{overpic}\hspace*{-0.3em}
	\begin{overpic}[width=0.15\linewidth, height=1.0cm, tics=0, clip]
		{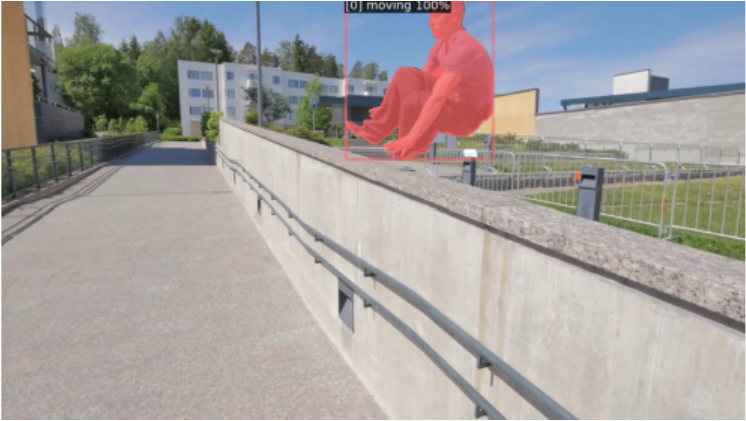}
	\end{overpic}

	\begin{overpic}[width=0.15\linewidth, height=1.0cm, tics=0, clip, trim={0.0cm 0.0cm 0.0cm 0.5cm}]
		{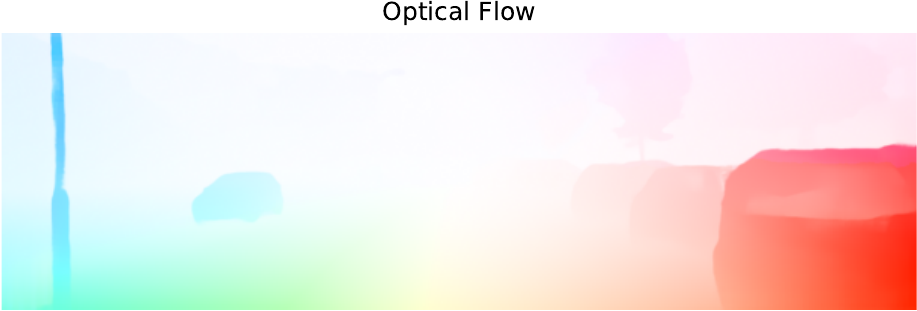}
		\put(-30, 2){\rotatebox{90}{\footnotesize{Optical}}}
		\put(-15, 9){\rotatebox{90}{\footnotesize{Flow}}}
	\end{overpic}\hspace*{-0.3em}
	\begin{overpic}[width=0.15\linewidth, height=1.0cm, tics=0, clip, trim={0.0cm 0.0cm 0.0cm 0.5cm}]
		{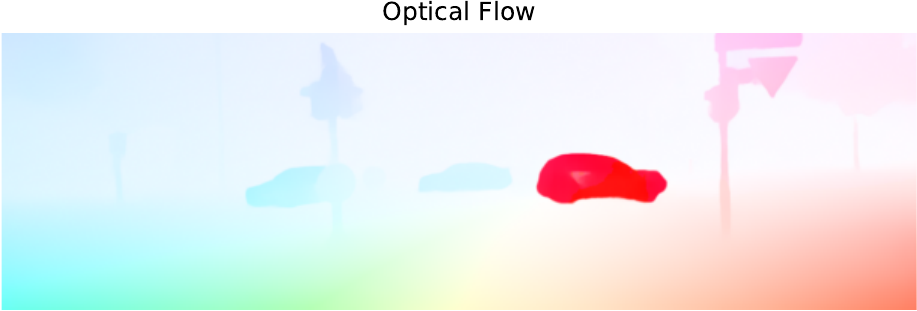}
	\end{overpic}\hspace*{-0.3em}
	\begin{overpic}[width=0.15\linewidth, height=1.0cm, tics=0, clip, trim={0.0cm 0.0cm 0.0cm 0.5cm}]
		{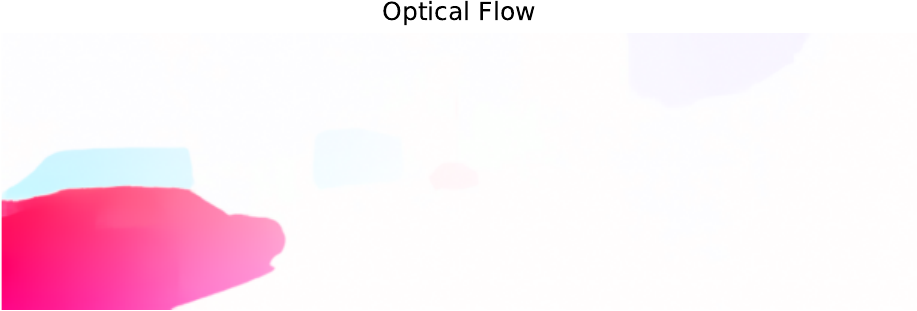}
	\end{overpic}
	\begin{overpic}[width=0.15\linewidth, height=1.0cm, tics=0, clip, trim={0.0cm 0.0cm 0.0cm 0.5cm}]
		{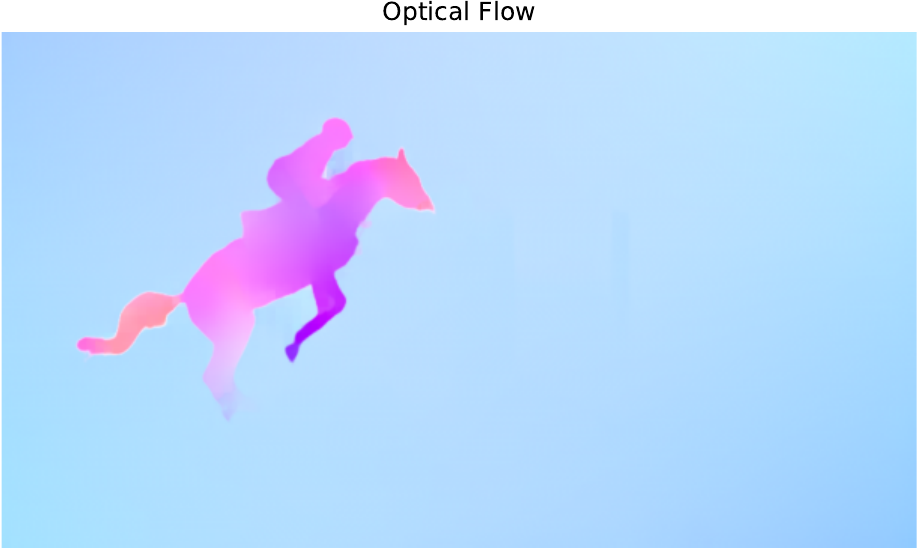}
	\end{overpic}\hspace*{-0.3em}
	\begin{overpic}[width=0.15\linewidth, height=1.0cm, tics=0, clip, trim={0.0cm 0.0cm 0.0cm 0.5cm}]
		{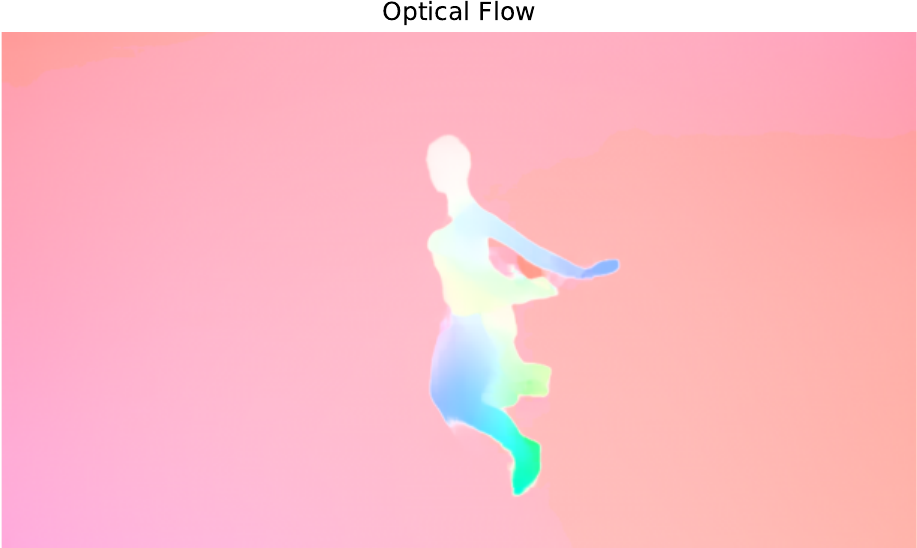}
	\end{overpic}\hspace*{-0.3em}
	\begin{overpic}[width=0.15\linewidth, height=1.0cm, tics=0, clip, trim={0.0cm 0.0cm 0.0cm 0.5cm}]
		{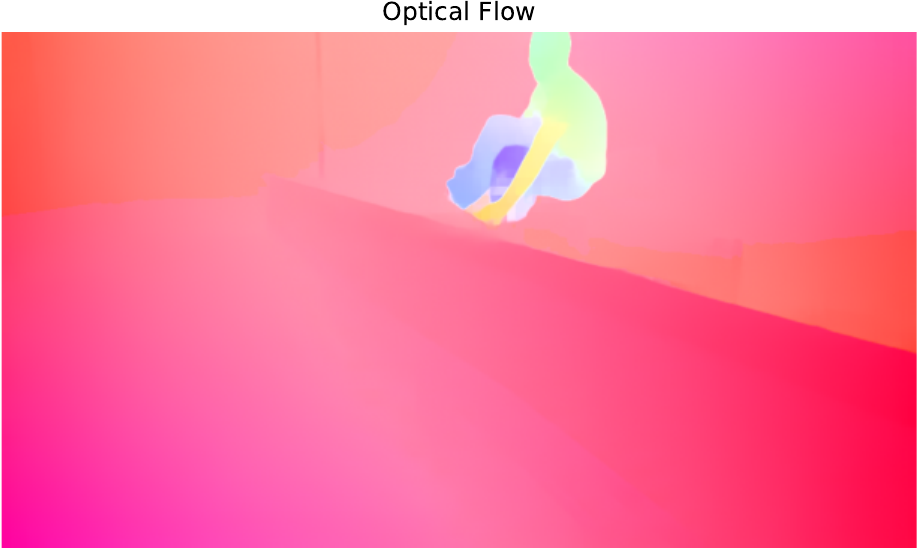}
	\end{overpic}
	
	\begin{overpic}[width=0.15\linewidth, height=1.0cm, tics=0, clip, trim={0.5cm 0.0cm 0.5cm 0.0cm}]
		{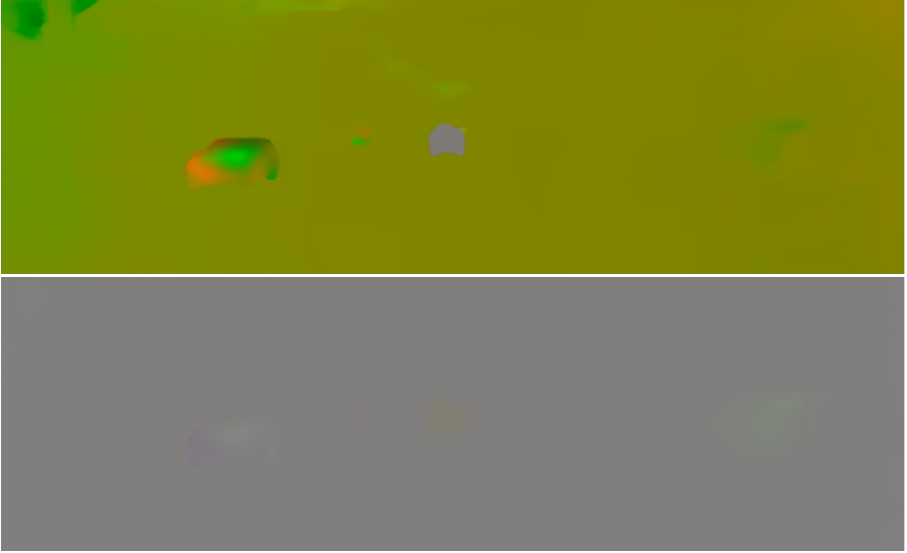}
		\put(-30, 6){\rotatebox{90}{\footnotesize{Scene}}}
		\put(-15, 9){\rotatebox{90}{\footnotesize{Flow}}}
	\end{overpic}\hspace*{-0.3em}
	\begin{overpic}[width=0.15\linewidth, height=1.0cm, tics=0, clip, trim={0.5cm 0.0cm 0.5cm 0.0cm}]
		{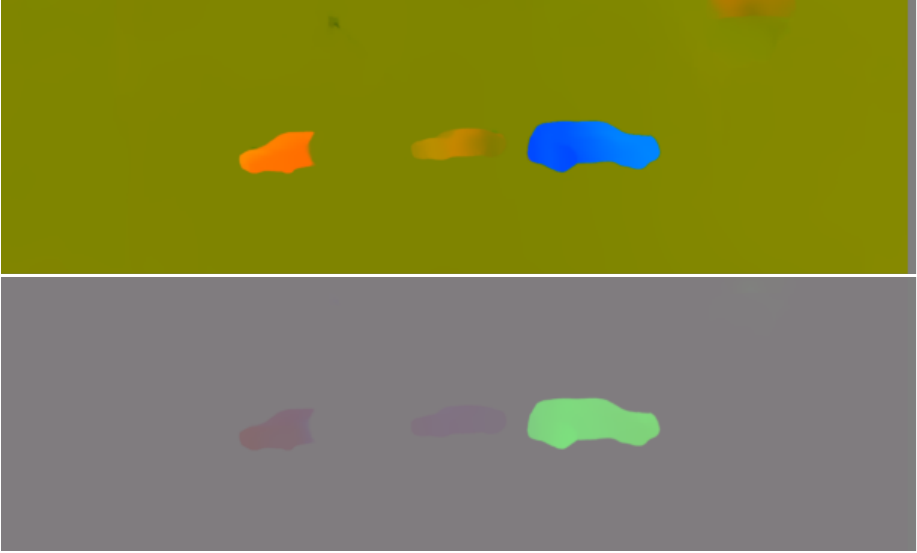}
	\end{overpic}\hspace*{-0.3em}
	\begin{overpic}[width=0.15\linewidth, height=1.0cm, tics=0, clip, trim={0.5cm 0.0cm 0.5cm 0.0cm}]
		{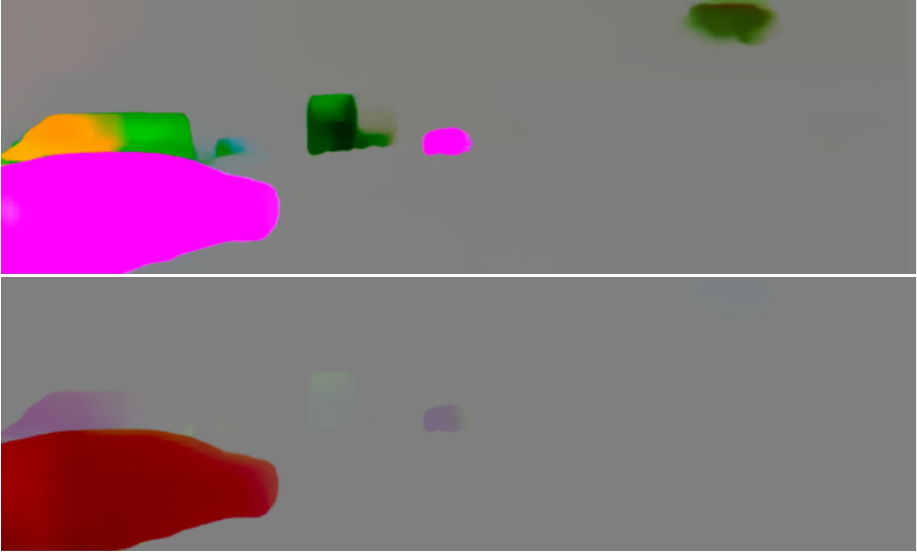}
	\end{overpic}\hspace*{-0.1em}
	\begin{overpic}[width=0.15\linewidth, height=1.0cm, tics=0, clip, trim={0.5cm 0.0cm 0.5cm 0.5cm}]
		{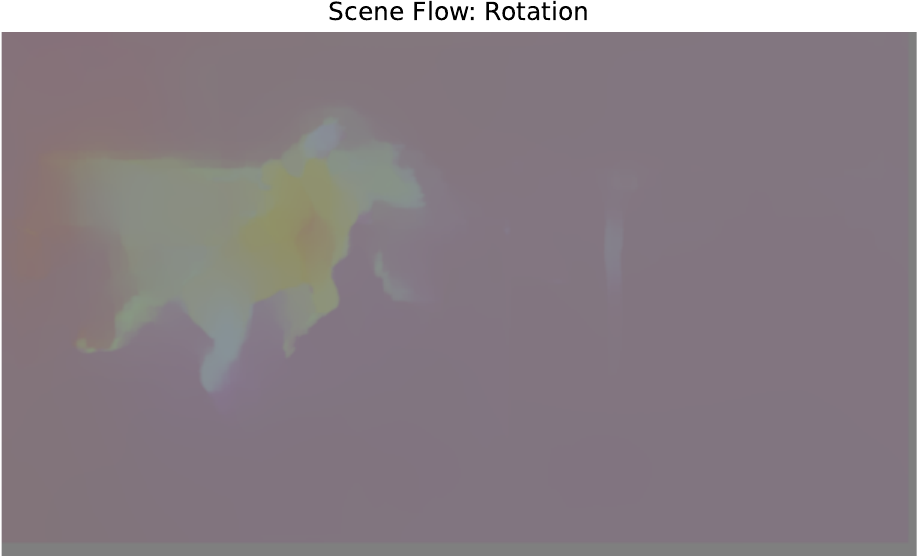}
	\end{overpic}\hspace*{-0.3em}
	\begin{overpic}[width=0.15\linewidth, height=1.0cm, tics=0, clip, trim={0.5cm 0.0cm 0.5cm 0.5cm}]
		{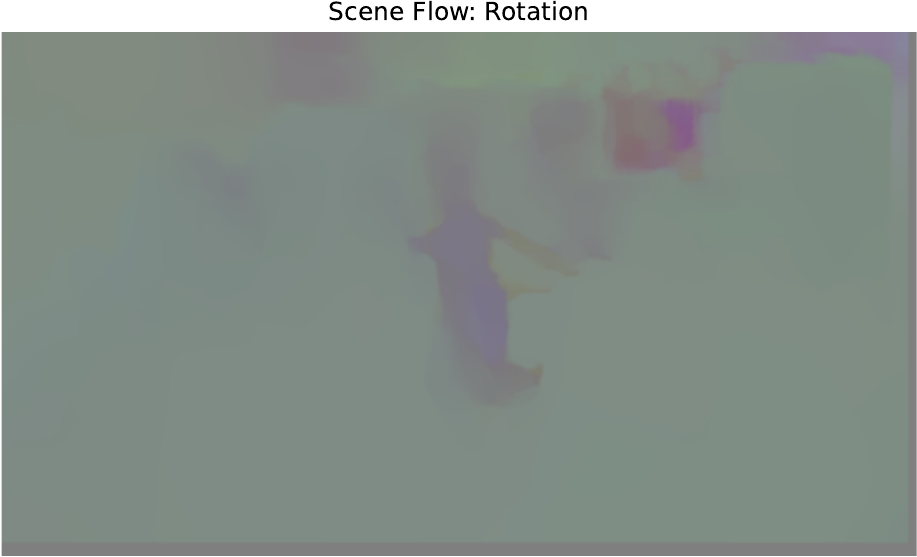}
	\end{overpic}\hspace*{-0.3em}
	\begin{overpic}[width=0.15\linewidth, height=1.0cm, tics=0, clip, trim={0.5cm 0.0cm 0.5cm 0.5cm}]
		{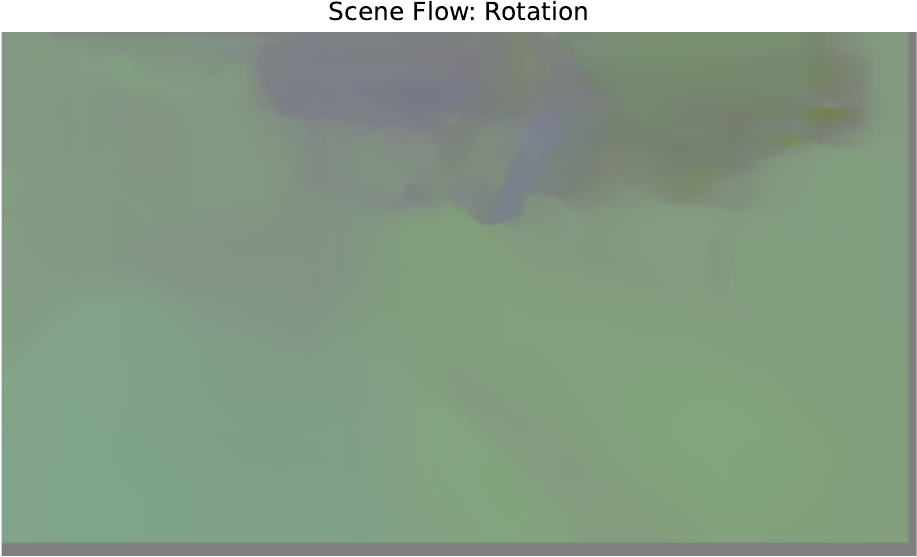}

	\end{overpic}
	\\[0.1em]
	
	\begin{overpic}[width=0.15\linewidth, height=1.0cm, tics=0, clip]
		{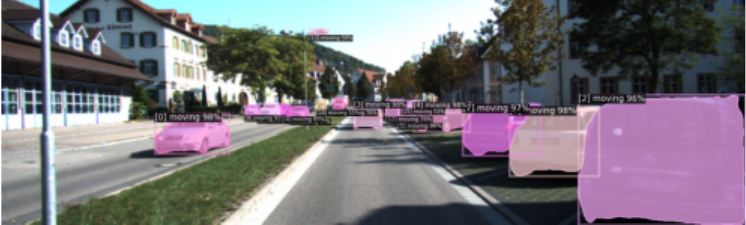}
		\put(-30, 12){\rotatebox{90}{\footnotesize{RGB}}}
		\put(-15, 15){\rotatebox{90}{\footnotesize{\cite{cheng2021mask2former}}}}
	\end{overpic}\hspace*{-0.3em}
	\begin{overpic}[width=0.15\linewidth, height=1.0cm, tics=0, clip]
		{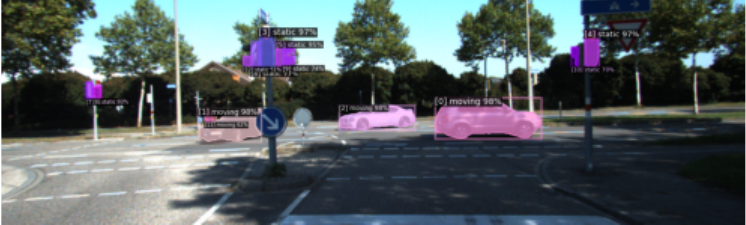}
	\end{overpic}\hspace*{-0.3em}
	\begin{overpic}[width=0.15\linewidth, height=1.0cm, tics=0, clip]
		{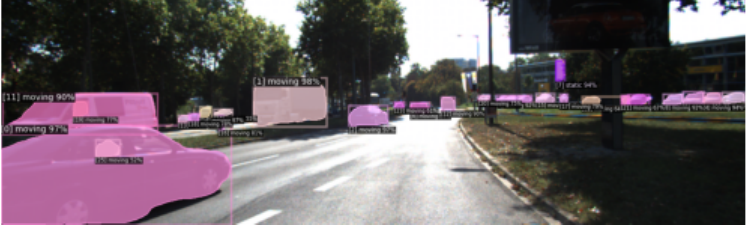}
	\end{overpic}\hspace*{-0.1em}
	\begin{overpic}[width=0.15\linewidth, height=1.0cm, tics=0, clip]
		{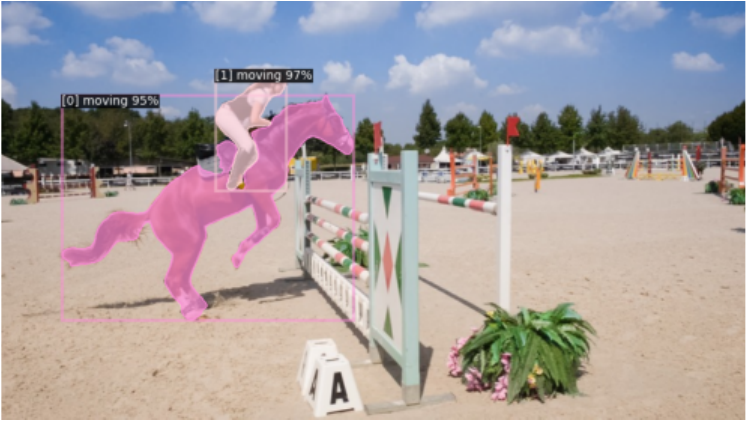}
	\end{overpic}\hspace*{-0.3em}
	\begin{overpic}[width=0.15\linewidth, height=1.0cm, tics=0, clip]
		{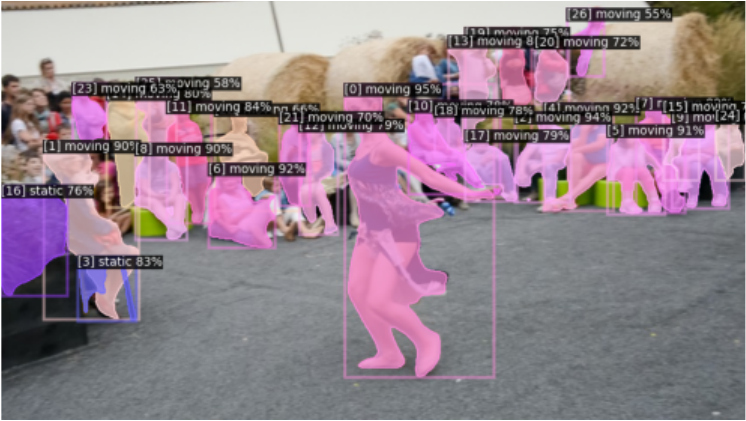}
	\end{overpic}\hspace*{-0.3em}
	\begin{overpic}[width=0.15\linewidth, height=1.0cm, tics=0, clip]
		{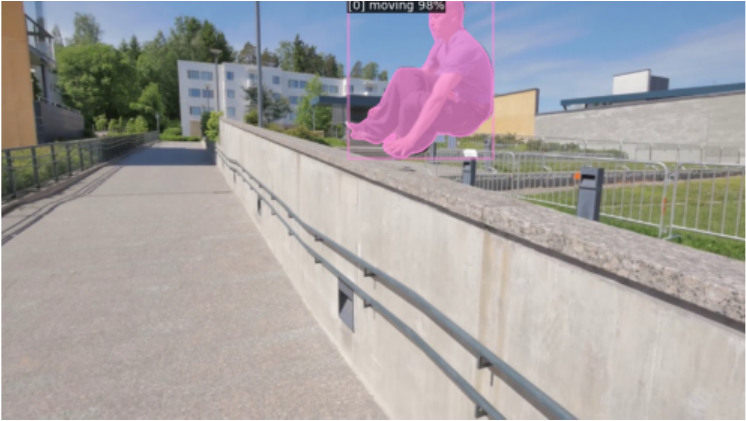}
	\end{overpic}
	
	\begin{overpic}[width=0.15\linewidth, height=1.0cm, tics=0, clip]
		{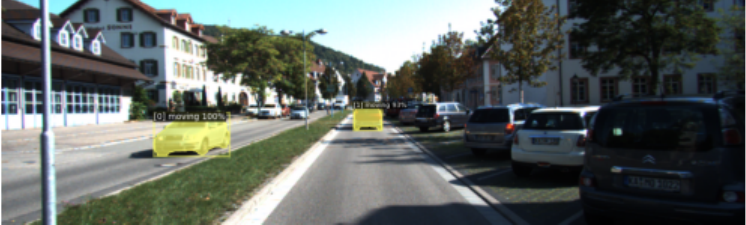}
		\put(-30, 7){\rotatebox{90}{\footnotesize{Raptor}}}
		\put(-15, 15){\rotatebox{90}{\footnotesize{\cite{neoral2021monocular}}}}
	\end{overpic}\hspace*{-0.3em}
	\begin{overpic}[width=0.15\linewidth, height=1.0cm, tics=0, clip]
		{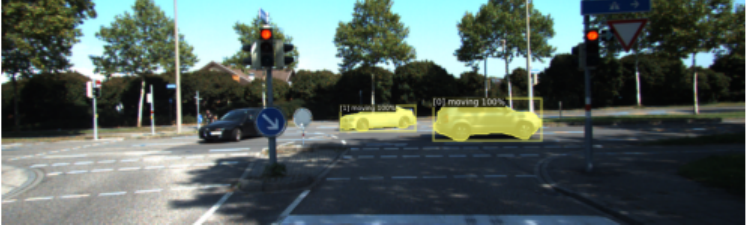}
	\end{overpic}\hspace*{-0.3em}
	\begin{overpic}[width=0.15\linewidth, height=1.0cm, tics=0, clip]
		{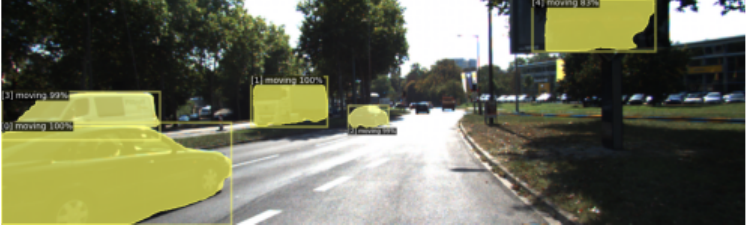}
	\end{overpic}\hspace*{-0.1em}
	\begin{overpic}[width=0.15\linewidth, height=1.0cm, tics=0, clip]
		{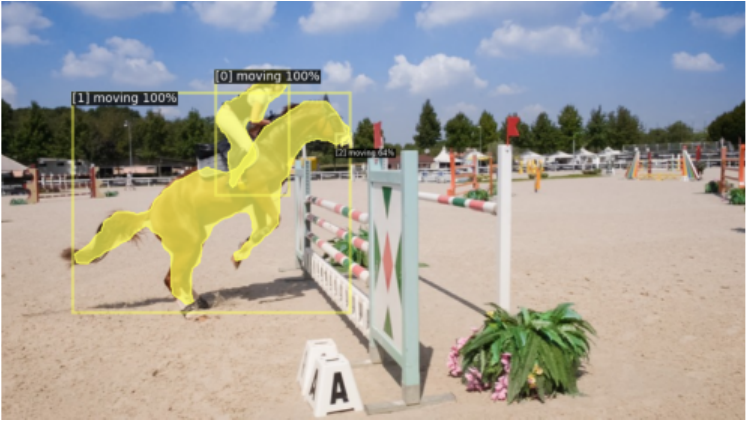}
	\end{overpic}\hspace*{-0.3em}
	\begin{overpic}[width=0.15\linewidth, height=1.0cm, tics=0, clip]
		{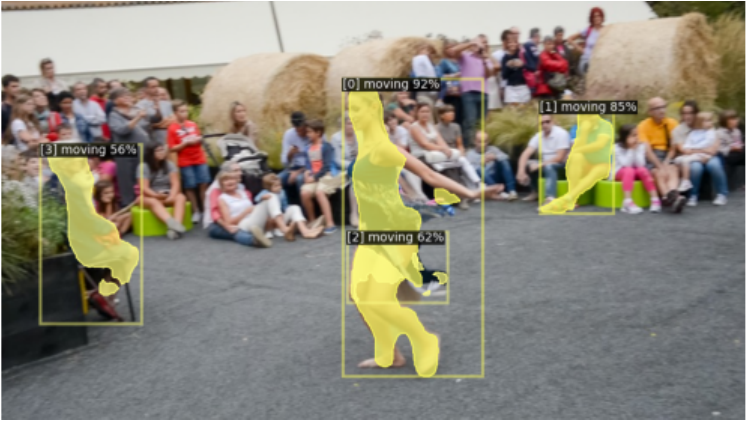}
	\end{overpic}\hspace*{-0.3em}
	\begin{overpic}[width=0.15\linewidth, height=1.0cm, tics=0, clip]
		{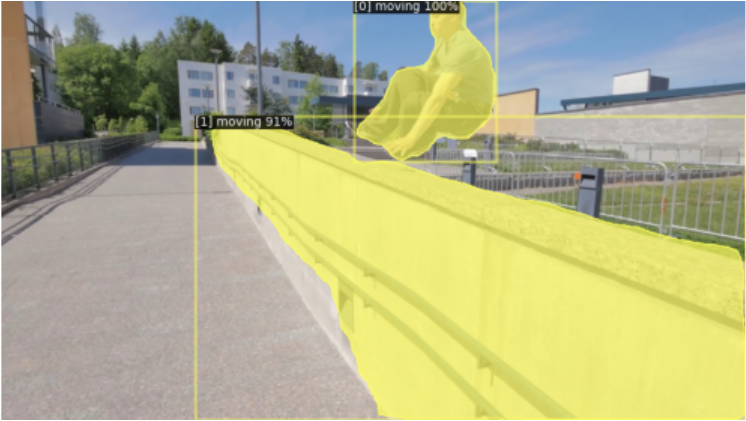}
	\end{overpic}

	\begin{overpic}[width=0.15\linewidth, height=1.0cm, tics=0, clip]
		{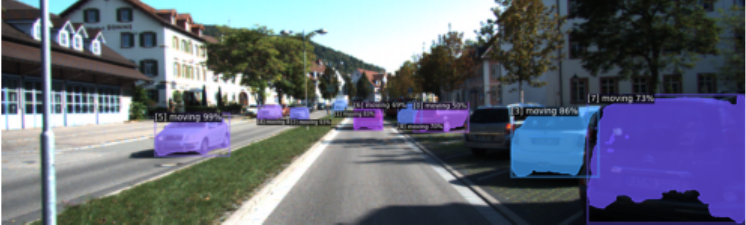}
		\put(-30, 11){\rotatebox{90}{\footnotesize{Ours}}}
		\put(-15, 2){\rotatebox{90}{\footnotesize{RGB+OF}}}
	\end{overpic}\hspace*{-0.3em}	
	\begin{overpic}[width=0.15\linewidth, height=1.0cm, tics=0, clip]
		{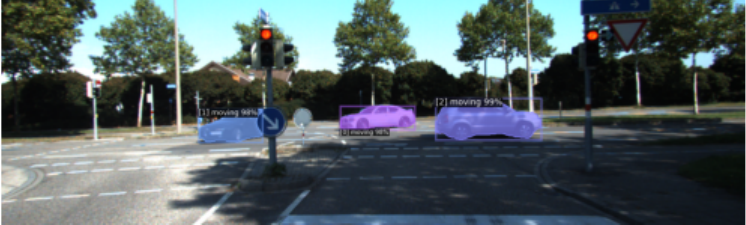}
	\end{overpic}\hspace*{-0.3em}
	\begin{overpic}[width=0.15\linewidth, height=1.0cm, tics=0, clip]
		{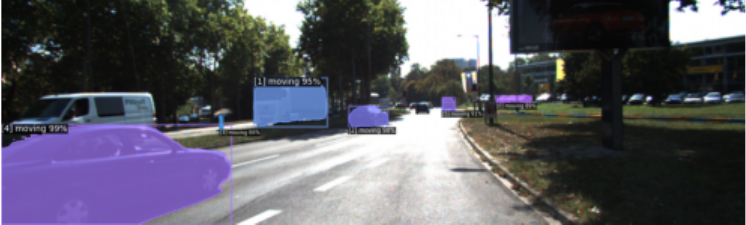}
	\end{overpic}\hspace*{-0.1em}
	\begin{overpic}[width=0.15\linewidth, height=1.0cm, tics=0, clip]
		{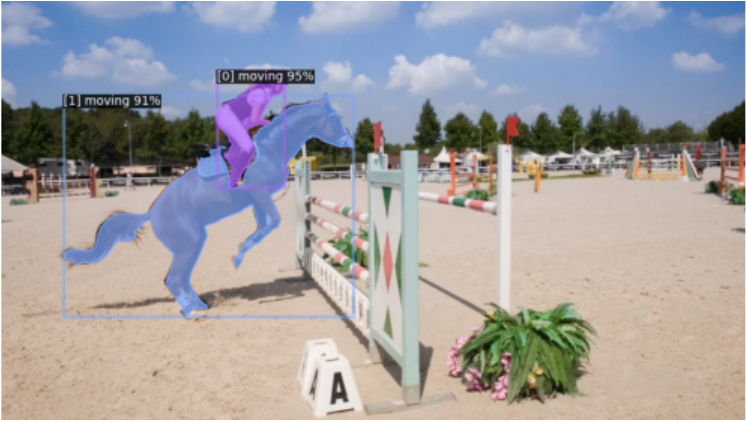}
	\end{overpic}\hspace*{-0.3em}
	\begin{overpic}[width=0.15\linewidth, height=1.0cm, tics=0, clip]
		{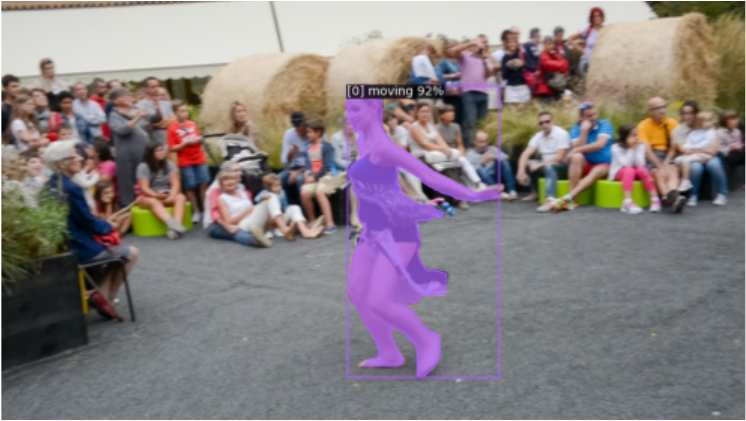}
	\end{overpic}\hspace*{-0.3em}
	\begin{overpic}[width=0.15\linewidth, height=1.0cm, tics=0, clip]
		{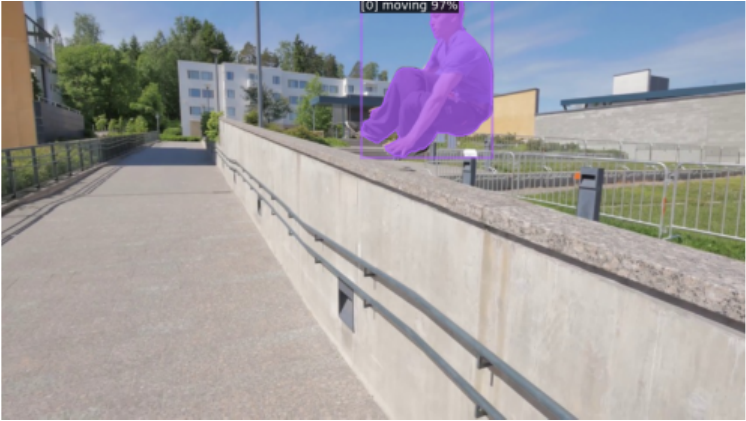}
	\end{overpic}
	
	\begin{overpic}[width=0.15\linewidth, height=1.0cm, tics=0, clip]
		{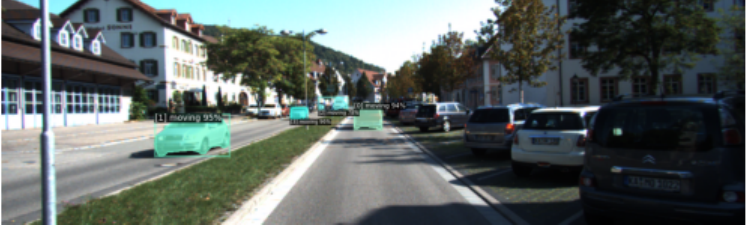}
		\put(-30, 8){\rotatebox{90}{\footnotesize{Ours}}}
		\put(-15, -5){\rotatebox{90}{\footnotesize{RGB+SF}}}
	\end{overpic}\hspace*{-0.3em}	
	\begin{overpic}[width=0.15\linewidth, height=1.0cm, tics=0, clip]
		{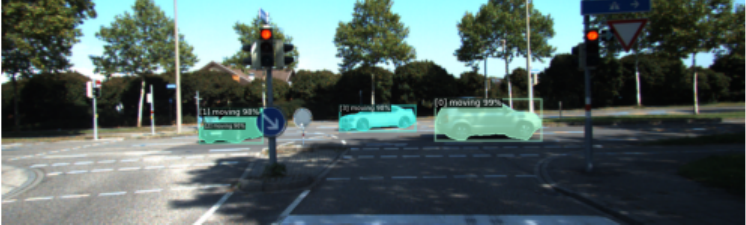}
	\end{overpic}\hspace*{-0.3em}
	\begin{overpic}[width=0.15\linewidth, height=1.0cm, tics=0, clip]
		{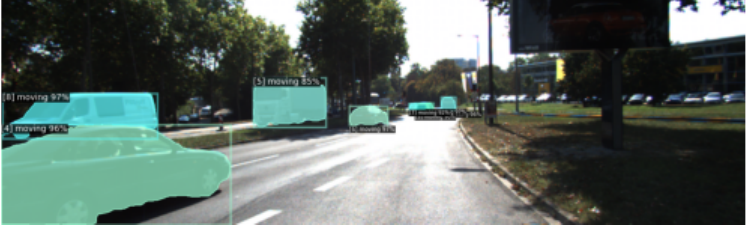}
	\end{overpic}\hspace*{-0.1em}
	\begin{overpic}[width=0.15\linewidth, height=1.0cm, tics=0, clip]
		{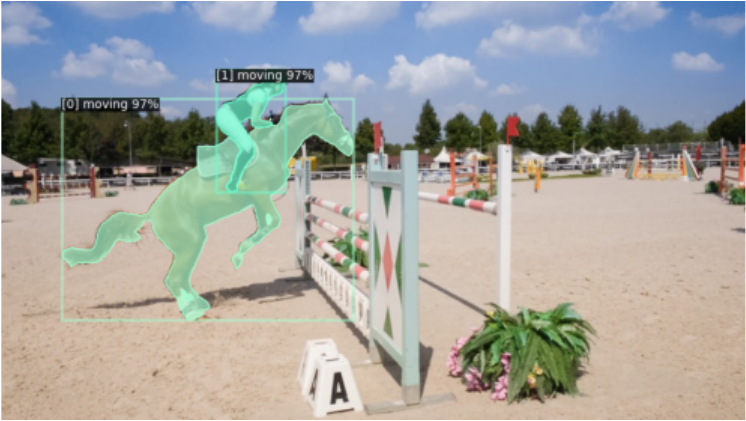}
	\end{overpic}\hspace*{-0.3em}
	\begin{overpic}[width=0.15\linewidth, height=1.0cm, tics=0, clip]
		{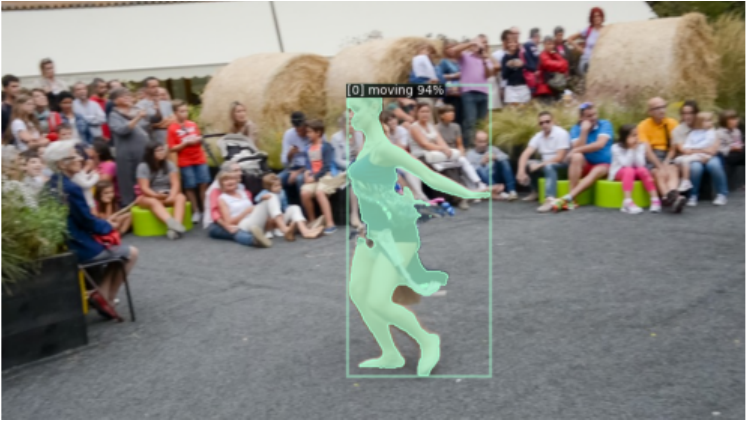}
	\end{overpic}\hspace*{-0.3em}
	\begin{overpic}[width=0.15\linewidth, height=1.0cm, tics=0, clip]
		{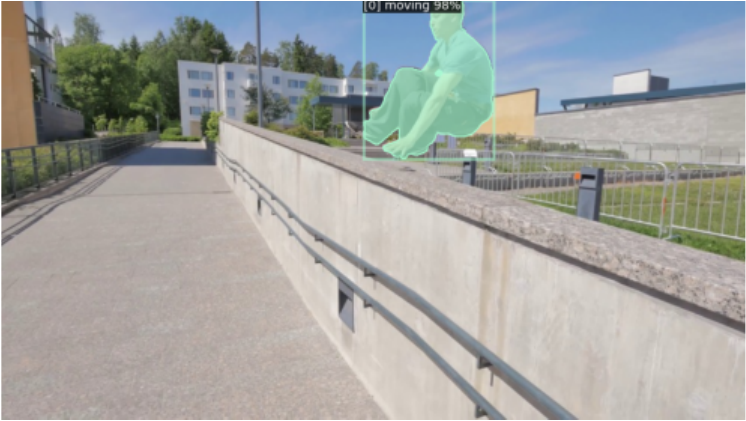}
	\end{overpic}

	\caption{Qualitative comparison on Kitti and Davis.}
\label{fig:qual}
\vspace{-13pt}
\end{figure*} 
Mix 1 does neither contain real data with non-rigid motions nor realistic driving scenes with many traffic participants. Since a variety of datasets exists, we can fix this problem by training on more diverse data. We combine sources from up to six datasets in our training. Our new mixes 2 and 3 offer multiple cases of non-rigid motions, multiple objects moving in union and hard degenerate motion scenarios. See Figure \ref{fig:mix_data} in Appendix for examples of how different training data can improve previous failures. It can be seen in Table \ref{tab:ablation_fusion} and \ref{tab:ablation_modality}, how both training data and different modalities affect performance. We observe that depending on the training data, performance of $ \textbf{M}^3$Former improves drastically (see Figure \ref{fig:training_data} in Suppl. Sec. \ref{sup:dataset_mixes}).

Table \ref{tab:sota} shows the SotA in supervised motion segmentation on Kitti and Davis. We visualize examples of the test splits in Figure \ref{fig:qual}. Note how for Mix 3 results, the models still have never seen the evaluation data. 
Our results are not necessarily surprising, as we partially trained on the target domain. However, we find that our incremental improvements behave quite causal: \textit{Most failure modes of the model disappear when supervised properly}. Previous shortcomings can be resolved solely with better datasets instead of architectural changes. Our proposed model architecture is simple and flexible. 
Surprisingly, our results show that even by using 2D optical flow, we can reach SotA performance on Kitti without using any real driving data. It can be seen that a minimal 3D motion representation like scene flow can be effective even with noisy data. 
Models can pick up strong cues for moving objects even from context alone. For example, a car that is placed on a driving lane is likely in motion compared to one parked to the side. Interestingly, creating a balanced dataset is a new optimization problem in itself \cite{ranftl2021vision}. We observe, that adding just more data sources can detoriate performance on Davis. Depending on the downstream-application, the data needs to be correctly balanced. We leave this for future work.

\section{Conclusion}
\label{sec:conclusion}
We systematically analyzed the motion segmentation problem from monocular video. In our experiments we identified the importance of different 2D and 3D motion representations on multiple datasets. 
We proposed a novel transformer fusion architecture $\textbf{M}^{3}$Former which fuses appearance and motion information on multiple scales. We analyzed multiple fusion schemes within this framework.
Our approach achieves SotA performance by leveraging the flexible attention mechanism and diverse training data. Our findings showed that both 2D and 3D motion can give strong performance when trained on appropriate data. Since appearance data mostly drives segmentation, the importance of high-quality motion estimates gets weaker when scaling the data size. 
\paragraph{ACKNOWLEDGMENT.}
Research presented here has been supported by the Robert Bosch GmbH. 
We thank our colleagues Annika Hagemann, Fabian Schmidt, Tamas Kapelner and EashwaraErahan for fruitful discussions.

{\small
\bibliographystyle{ieee_fullname}
\bibliography{draft}
}

\clearpage
\section*{Supplementary Material}

In this supplementary material, we provide further details on our approach (Section \ref{sup:ours}), experiment settings (Section \ref{sup:exp}) and further experimental results (Section \ref{sup:ablat}) as well as visualizations (Section \ref{sup:visu}). 

\section{Our Approach}
\label{sup:ours}

\subsection{Placement in the Literature}
\label{sup:literature}
There is a vast amount of related literature on segmentation, motion segmentation, moving object discovery and unsupervised feature learning. 
Table \ref{tab:related} summarizes the development in this field and where our approach fits in. We compare with previous work on supervised motion segmentation.
\begin{table*}[b!]
	\centering
	\resizebox{1.0\textwidth}{!}{%
		\begin{tabular}{lccccccc}
			\toprule
			\textbf{Paper} & \textbf{Publication} & \textbf{End-to-End} & \textbf{\# Input Frames} & \textbf{Supervision} & \textbf{Modality} & \textbf{Task} & \textbf{Fusion} \\
			\midrule
			\cite{bideau2016s} & ECCV 2016 & \xmark & 2 & \xmark & Optical Flow & BS & \xmark \\
			\cite{tokmakov2017learning} & ICCV 2017 & \cmark & $ \geq 3 $ & \cmark & RGB + Optica	l Flow & BS & GRU \\
			\cite{bideau2018moa} & ECCV 2018 & \cmark & 2 & \xmark & Optical Flow & BS & \xmark \\
			\cite{lv2018learning} & ECCV 2018 & \cmark & 2 & \cmark & RGB-D & BS, Odometry & \xmark \\
			\cite{bideau2018best} & CVPR 2018 & (\cmark) & 2 & \cmark & RGB + Optical Flow & BS & \xmark \\
			\cite{lu2019see}& CVPR 2019 & \cmark & 2 & \xmark & RGB & BS & Attention \\
			\cite{xie2019object} & CVPR 2019 & \cmark & $ > 2 $ & \xmark & RGB + Optical Flow & IS / (VIS) & Convolution \\
			\cite{dave2019towards} & ICCV 2019 & \cmark & 2 & \cmark & RGB + Optical Flow & IS / (VIS) & Convolution \\
			\cite{mohamed2021modetr} & NeurIPS 2020 & \cmark & 2 & \cmark & RGB + Optical Flow & Detection & Attention \\
			\cite{zhou2020motion} & AAAI 2020 & \cmark & 2 & \cmark & RGB + Optical Flow & BS & Attention \\
			\cite{liu2021emergence} & NeurIPS 2021 & \cmark & 2 & \xmark & Optical Flow & BS & \xmark \\
			\cite{yang2021self} & ICCV 2021 & \cmark & 2 & \xmark & Optical Flow & BS & \xmark \\
			\cite{yang2021learning} & CVPR 2021 & \cmark & 2 & \cmark & Geom. Costs & BS / IS & \xmark \\
			\cite{neoral2021monocular} & BMVC 2021 & \cmark & 3 & \cmark & RGB + Geom. Costs & IS & Convolution \\
			\cite{bao2022discovering} & CVPR 2022 & \cmark & 3 & \xmark & RGB & Object Discovery & GRU \\ 
			\cite{karazija2022unsupervised} & NeurIPS 2022 & \cmark & 1 & \xmark & RGB & BS & \xmark \\ 
			\cite{singh2022simple} & NeurIPS 2022 & \cmark & $ \leq 2 $ & \xmark & RGB & VOS / Object Discovery & RNN \\ 
			\cite{elsayed2022savi++} & NeurIPS 2022 & \cmark & $ \leq 2 $ & \xmark & RGB & VOS / Object Discovery & RNN \\ 
			\cite{choudhury2022guess} & BMVC 2022 & \cmark & $ \leq 2 $ & \xmark & RGB + Optical Flow & Binary Segmentation & Spectral Clustering \\
			\cite{bao2023object} & CVPR 2023 & \cmark & 3 & \xmark & RGB & Object Discovery & Attention \\
			\midrule
			Ours & 2023 & \cmark & $ \geq 2 $ & \cmark & RGB + 3D scene flow & IS & Attention \\
			\bottomrule
	\end{tabular}}
	\caption{Taxonomy of related segmentation literature. We distinguish Binary Segmentation (BS), Instance Segmentation (IS) and Object Segmentation (OS) as task acronyms.}
	\label{tab:related}
\end{table*}

\subsection{Architecture}
\label{sup:arch}
We base our approach on the Mask2Former \cite{cheng2022masked} architecture. Our two-stream fusion model is depicted in Figure \ref{fig:architecture}. It features two identical branches with its own dedicated parameters $ \Theta_{rgb} $, $ \Theta_{motion} $, i.e. two sets of backbone, encoder and decoder. Learned attention masks $ M^{l-1}_{rgb} $, $ M^{l-1}_{motion} $ let selected features $ z^{l}_{rgb} $, $ z^{l}_{motion} $ from both streams at scale $ l $ interact with each other and two sets of queries $ q_{rgb} $ and $ q_{motion} $. Finally, we fuse information from both streams into a single prediction with $ 1 \times 1 $ convolution layers: We fuse output masks and class logits for all (100) queries. A twin-stream architecture is motivated out of convenience since we can combine pretrained branches and finetune them together. We believe that in the future a much lighter motion branch would suffice and further optimizations can be made since segmentation is mainly driven by the appearance branch. However, there may exist datasets where motion features mainly drive object detection and segmentation as can be seen in Section \ref{sup:moca}. We keep the default settings of \cite{cheng2022masked} in terms of architecture hyperparameters. We use a ResNet-50 \cite{he2016deep} backbone, pretrained on ImageNet \cite{russakovsky2015imagenet}. Every motion stream has a $ 1\times 1 $ convolution layer as projection layer before the backbone.\\
\textbf{Transformer Encoder.}
We use the  multi-scale deformable attention Transformer \cite{zhu2020deformable} for encoding the backbone features. We use 6 layers applied to feature maps at resolution $ 1/8 $, $ 1/16 $ and $ 1/32 $.\\
\textbf{Transformer Decoder.}
We use the same transformer decoder as \cite{cheng2022masked} with 9 layers in total and 100 queries. We also supervise intermediate predictions with the auxiliary loss. 

\section{Experimental Setup}
\label{sup:exp}

\subsection{Training Details}
\label{sup:implementation}
We follow a similar training setup as \cite{cheng2022masked}. Our networks are optimized using AdamW \cite{loshchilov2017decoupled} with a learning rate of $ 1.0 \times 10^{-4} $ and a weight decay of $ 0.05 $ for all backbones. A learning rate multiplier of $ 0.1 $ is applied to the backbone. We employ gradient clipping when the 2-norm exceeds $ 0.1 $ for stability. For augmentation, we use DETR-style \cite{carion2020end} random scaling, cropping and flipping. We follow the same losses $ L = \lambda_{ce}L_{ce} +  \lambda_{dice}L_{dice} + \lambda_{cls}L_{cls} $ with $ \lambda_{ce} =  5.0 $, $ \lambda_{dice} = 5.0 $. We set $ \lambda_{cls} = 2.0 $ for predictions matched with the groundtruth and $ 0.1 $ for the \say{no-object} class. Finally, we also use importance sampling like \cite{cheng2022masked} with $ K = 12544 $, i.e. $ 122 \times 122 $ points. We train our networks on two NVIDIA RTX A6000 GPU's. 

\paragraph{Single-modality training.}
We train single-modality models on the FlyingThings3D \cite{mayer2016large} dataset for $ 30 $ epochs. Out of convenience, we finetune the pretrained COCO checkpoint from \cite{cheng2022masked} and randomly initialize the classification head.  We train with a batch size of 12 and reduce the learning rate by a factor of $ 0.1 $ every 8 epochs.  

\paragraph{Multi-modal training.}
When training multi-modal models, we take the appearance branch \cite{cheng2022masked}, pretrained on COCO, and freeze it similar to the setup by \cite{dave2019towards, neoral2021monocular}. The idea is to \textit{retain} semantic object knowledge from an off-the-shelf detector and extend it for motion segmentation. We randomly initialize the classification head. We take the respective motion branch pretrained in the previous single-modality training. Both branches are then finetuned to combine knowledge from both modalities.\\
\textbf{Mix 1.}
We trained for 10 epochs with a batch size of 6 and reduce the learning rate by a factor of $ 0.1 $ after 8 epochs. We use neg. examples as augmentation with $ p_{neg} = 0.3 $.\\
\textbf{Mix 2.}
We follow the same setup as for Mix 1.\\
\textbf{Mix 3.}
Because of the larger size of the total dataset, we only train for 5 epochs and reduce the learning rate by a factor of $ 0.1 $ after 4 epochs. We use neg. examples as augmentation with $ p_{neg} = 0.05 $ as a trade-off between regularization and enough pos. examples for object discovery. We note how this trade-off needs to be carefully chosen depending on what is valued in a detector. We achieve a lower rate of false negatives in this way, but have more false positives. We believe fewer false negatives are more important from a safety perspective for applications like autonomous driving.

\subsection{Evaluation}
\label{sup:eval}
During inference, we use the standard Mask R-CNN \cite{he2017mask} setting where an image with shorter side is resized to 800 and longer side up-to 1333. We report standard COCO \cite{lin2014microsoft} metrics, foreground/background precision \cite{yang2021learning}, Precision (Pu), Recall (Ru) and F-score (Fu) \cite{dave2019towards} and the number of false positives and false negatives over the whole split \cite{neoral2021monocular}. 
Similar to \cite{neoral2021monocular}, we average over different IoU's $ \left[ 0.01,\; 0.1,\; 0.3,\; 0.5,\; 0.75,\; 0.9,\; 0.95 \right] $. Since we can compute matchings between groundtruth and prediction for these metrics for different confidence thresholds, we also average over multiple confidence values $ \left[ 0.3,\; 0.5,\; 0.7 \right] $ like \cite{neoral2021monocular}. Since datasets come in different sizes, we normalize the number of false positives/negatives. Intuitively this means, we measure the number of false positives/negatives per frame on particular data. At this point, strong models for SotA have false positive/negative detections only every $k$-th frame. 

\begin{table}[b!]
	\centering
	\begin{tabular}{lp{5cm}}
		\toprule
		\textit{moving} & person, bicycle, car,  motorcycle, airplane, bus, train, truck, boat, bird, cat, dog, horse, sheep, cow, elephant, bear, zebra, giraffe, frisbee, skis, snowboard, sports ball, kite, baseball bat, baseball glove, skateboard, surfboard, tennis racket \\
		\midrule
		\textit{static} & traffic light, hydrant, stop sign, parking meter, bench, backpack, umbrella, handbag, tie, suitcase, bottle, wine glass, cup, fork, knife, spoon, bowl, banana, apple, sandwich, orange, broccoli, carrot, hot dog, pizza, donut, cake, chair, couch, potted plant, bed, dining table, toilet, tv, laptop, mouse, remote, keyboard, cell phone, microwave, oven, toaster, sink, refrigator, book, clock, vase, scissors, teddy bear, hair drier, toothbrush \\
		\bottomrule
	\end{tabular}
	\caption{Moving object classes of COCO \cite{lin2014microsoft}.}
	\label{sup:coco_tab}
\end{table}
We came up with a strong and simple image detector baseline: Map semantically likely object classes to moving objects and others to static. Table \ref{sup:coco_tab} shows our distinction into moving and static classes of the COCO dataset. 
In theory, everything can move once a force is applied to it. However, this is hard to observe when we measure performance with current datasets (since often relevant objects move all the time). We use this mapping for the baseline model \cite{cheng2022masked}.

\subsection{Inference times and memory}
\label{sup:memory}
We measure runtime and max. memory usage on the first 300 examples from DAVIS using \textit{torchscript} on a NVIDIA GeForce RTX 2080 Ti (with Intel(R) Xeon(R) CPU E5-2630 v3 @ 2.40GHz). Measurements can be found in Table \ref{tab:timing}. We measure with a batch size of 1. Note how scaling depends on the individual complexities (of the attention mechanism). Since computation of pseudo-modalities is dependent on specific off-the-shelf expert models and input resolution, we omit a total runtime comparison. (Using an optical flow estimator like RAFT \cite{teed2020raft} runs for example at $ \approx 500 $ ms for HD-video.) It can be seen that combining multiple modalities comes with a price both for memory usage and runtime. Fusing modalities at multiple locations in the architecture adds up to this footprint. While we found bottleneck tokens \cite{nagrani2021attention} to not have a strong impact on performance, it might be very helpful when scaling up to high-resolution inputs, longer video or large batch sizes.
\begin{table}[h!]
	\begin{center}
			\resizebox{0.5\textwidth}{!}{%
			\begin{tabular}{ccc}
				Fusion mechanism & Runtime [ms] & Max. Memory [GB] \\
				\midrule
				Single-modality & 447.69 & 1.11 \\
				\midrule
				MBT \cite{nagrani2021attention} Decoder & 594.72 & 1.63 \\
				Naive Decoder & 631 & 2.20 \\
				Encoder & 664.43 & 1.77 \\
				Encoder + Decoder & 682.97 & 2.24 \\
				\bottomrule
			\end{tabular}
		}
		\caption{Runtime and memory footprint of our multi-modal architecture. Memory footprint depends on the fusion mechanism.
		}
		\label{tab:timing}
	\end{center}
\end{table}

\begin{figure*}[b!]
	\centering
	
	\vspace*{1.0em}
	\begin{overpic}[width=0.11\linewidth, height=1.5cm, tics=0, clip, trim={0.1cm 0.2cm 0.1cm 0.5cm}]
		{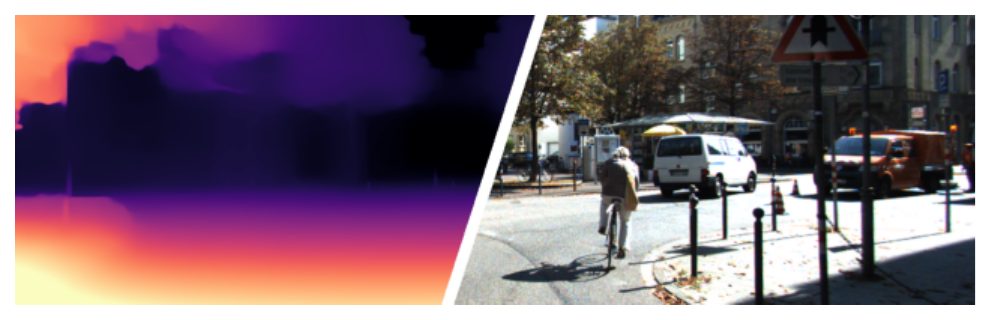}
		\put(-20, 25){\rotatebox{90}{\footnotesize{Input}}}
		\put(28, 90){\footnotesize{Mono}}
	\end{overpic}\hspace*{-0.3em}
	\begin{overpic}[width=0.11\linewidth, height=1.5cm, tics=0, clip, trim={0.1cm 0.2cm 0.1cm 0.5cm}]
		{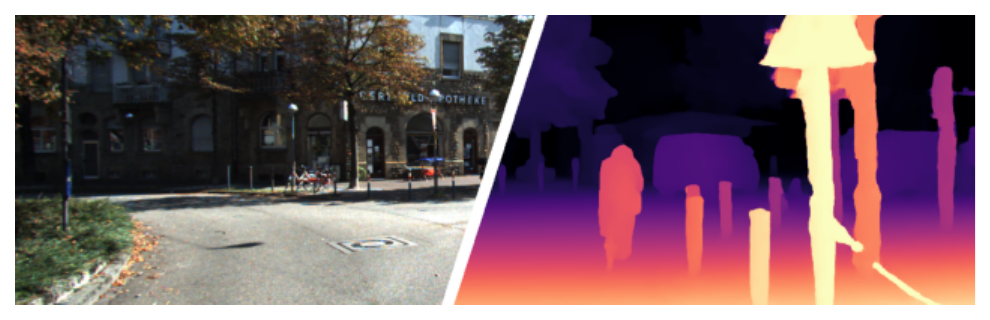}
		\put(28, 90){\footnotesize{Stereo}}
	\end{overpic}
	\begin{overpic}[width=0.11\linewidth, height=1.5cm, tics=0, clip, trim={0.1cm 0.2cm 0.1cm 0.5cm}]
		{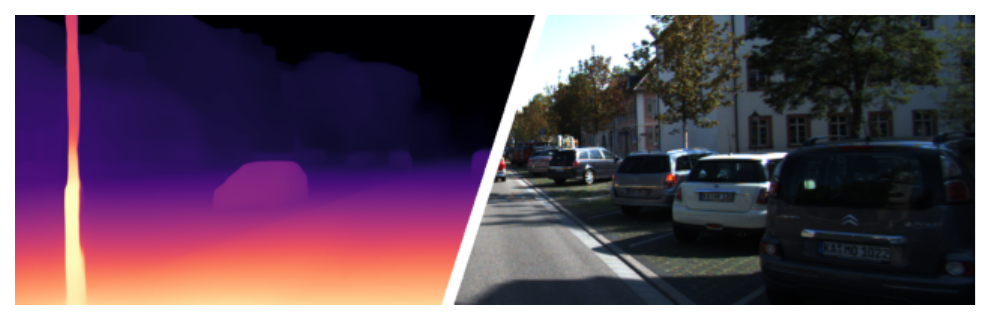}
		\put(28, 90){\footnotesize{Mono}}
	\end{overpic}\hspace*{-0.3em}
	\begin{overpic}[width=0.11\linewidth, height=1.5cm, tics=0, clip, trim={0.1cm 0.2cm 0.1cm 0.5cm}]
		{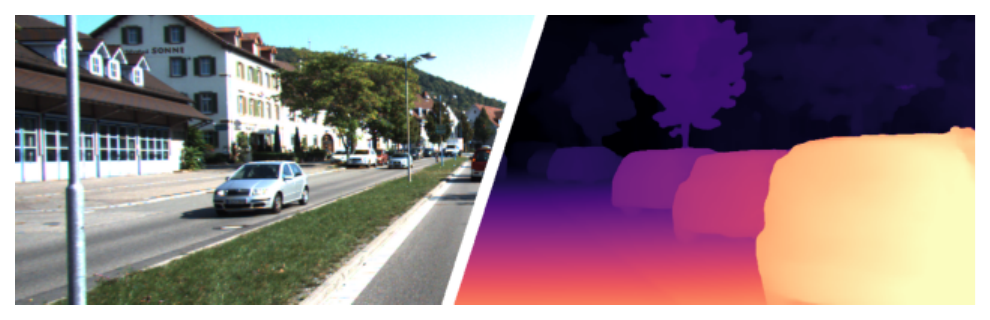}
		\put(28, 90){\footnotesize{Stereo}}
	\end{overpic} 
	\begin{overpic}[width=0.11\linewidth, height=1.5cm, tics=0, clip, trim={0.1cm 0.2cm 0.1cm 0.5cm}]
		{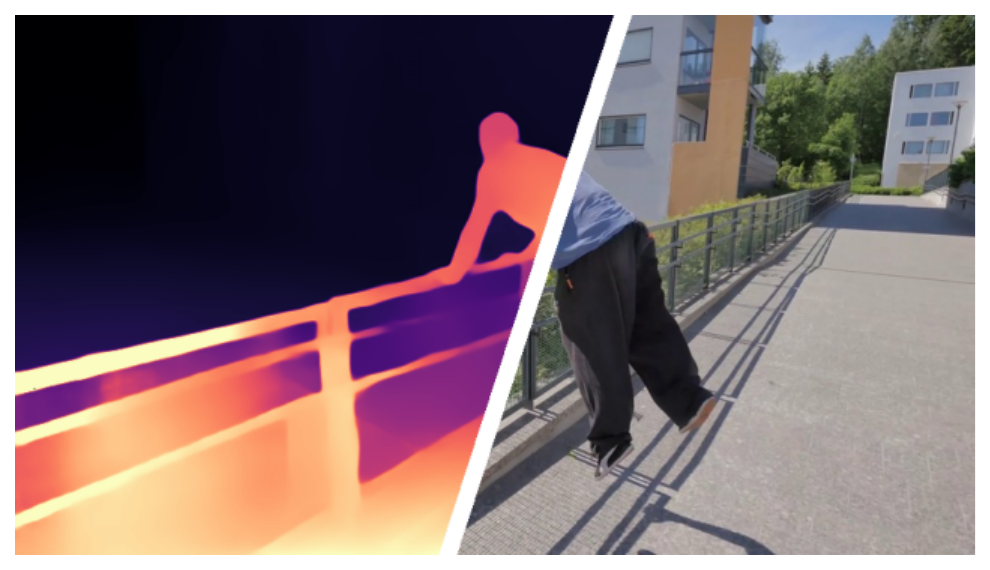}
		\put(15, 90){\footnotesize{rel. scale}}
	\end{overpic}\hspace*{-0.3em}
	\begin{overpic}[width=0.11\linewidth, height=1.5cm, tics=0, clip, trim={0.1cm 0.2cm 0.1cm 0.5cm}]
		{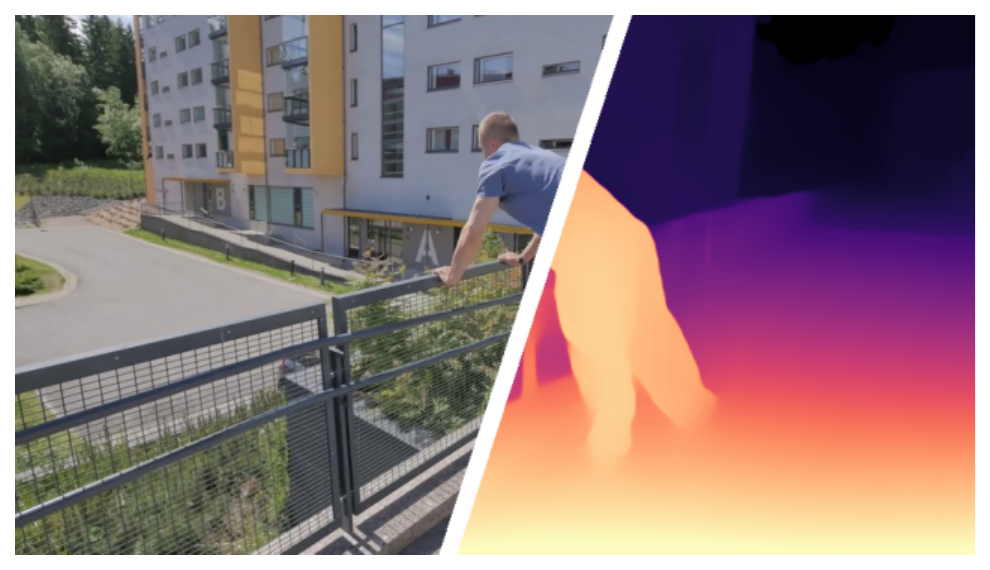}
		\put(15, 90){\footnotesize{abs. scale}}
	\end{overpic}
	\begin{overpic}[width=0.11\linewidth, height=1.5cm, tics=0, clip, trim={0.1cm 0.2cm 0.1cm 0.5cm}]
		{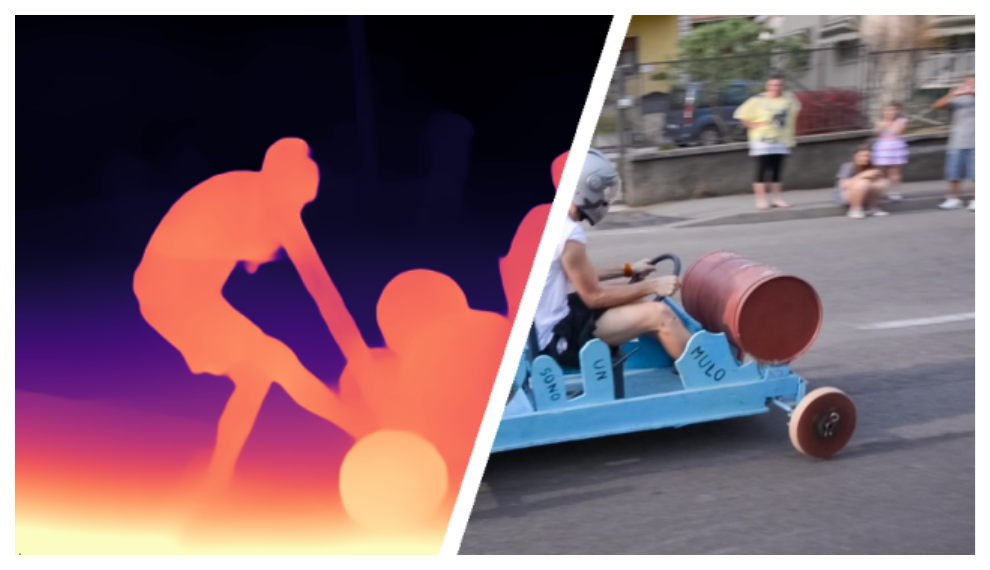}
		\put(15, 90){\footnotesize{rel. scale}}
	\end{overpic}\hspace*{-0.3em}
	\begin{overpic}[width=0.11\linewidth, height=1.5cm, tics=0, clip, trim={0.1cm 0.2cm 0.1cm 0.5cm}]
		{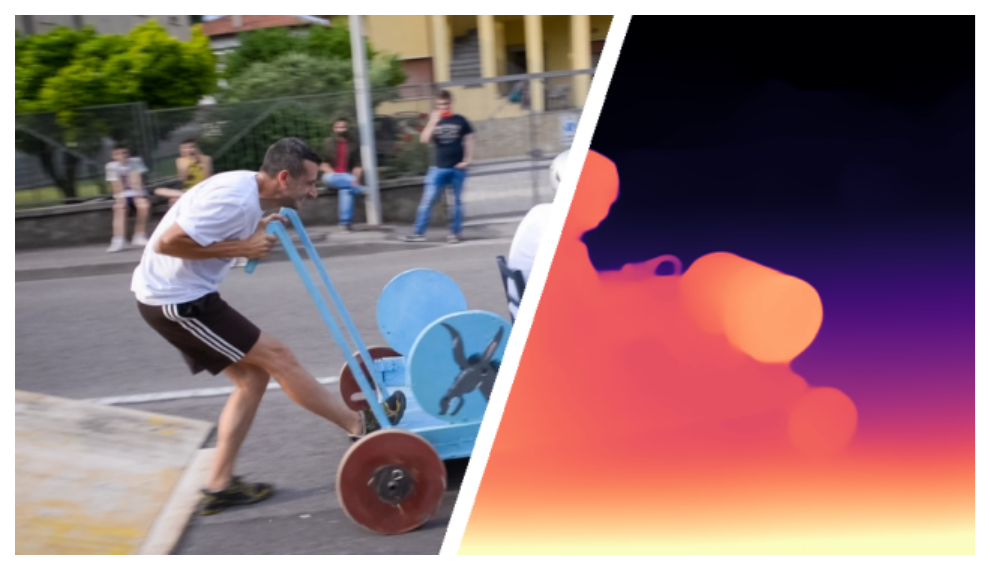}
		\put(15, 90){\footnotesize{abs. scale}}
	\end{overpic}
	
	\begin{overpic}[width=0.11\linewidth, height=1.25cm, tics=0, clip, trim={0.1cm 0.1cm 0.1cm 0.5cm}]
		{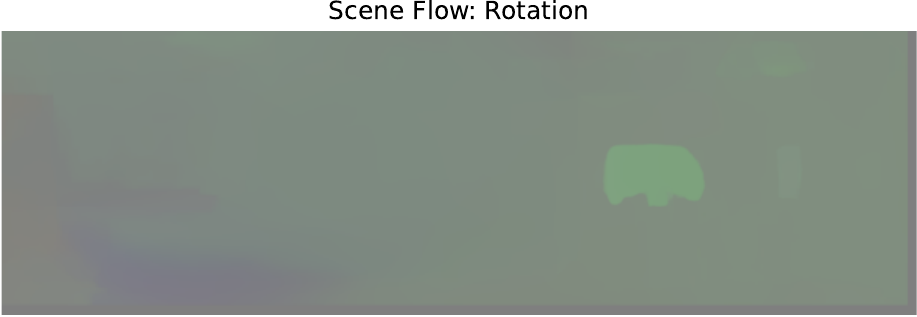}
		\put(-19, -30){\rotatebox{90}{\footnotesize{Scene Flow}}}
	\end{overpic}\hspace*{-0.3em}
	\begin{overpic}[width=0.11\linewidth, height=1.25cm, tics=0, clip, trim={0.1cm 0.1cm 0.1cm 0.5cm}]
		{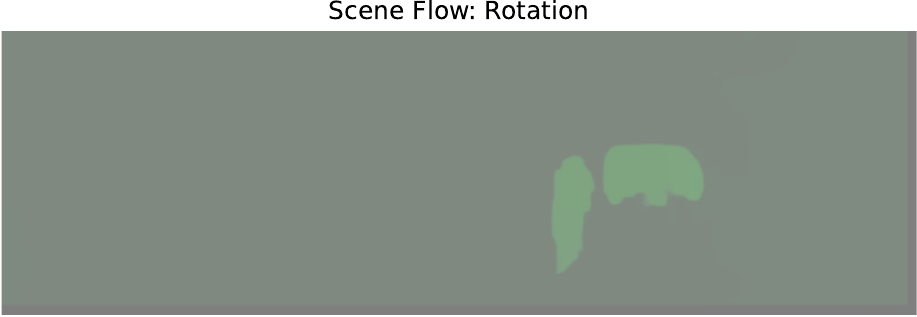}
	\end{overpic}
	\begin{overpic}[width=0.11\linewidth, height=1.25cm, tics=0, clip, trim={0.1cm 0.1cm 0.1cm 0.5cm}]
		{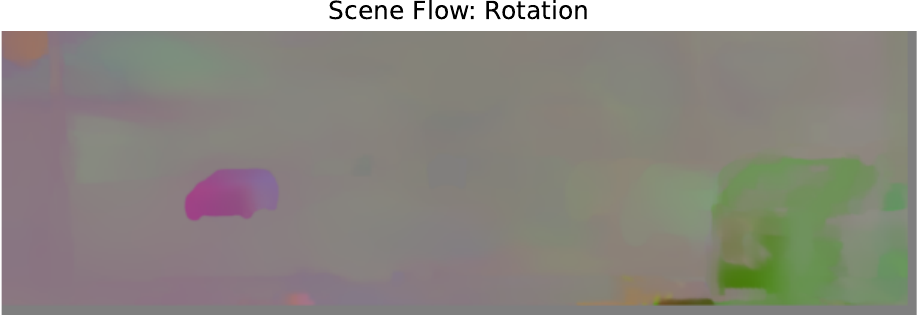}
	\end{overpic}\hspace*{-0.3em}
	\begin{overpic}[width=0.11\linewidth, height=1.25cm, tics=0, clip, trim={0.1cm 0.1cm 0.1cm 0.5cm}]
		{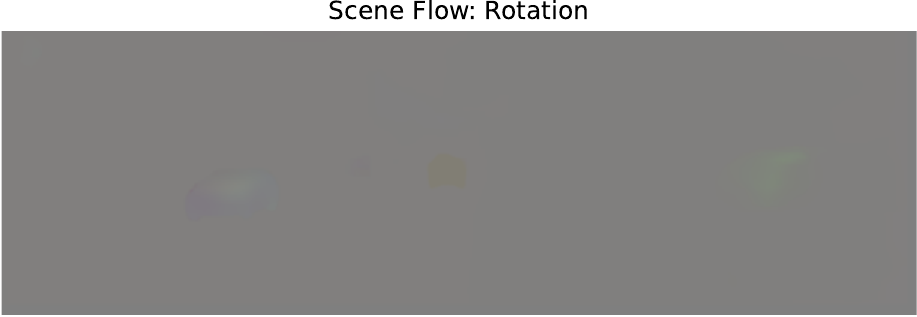}
	\end{overpic} 
	\begin{overpic}[width=0.11\linewidth, height=1.25cm, tics=0, clip, trim={0.1cm 0.1cm 0.1cm 0.5cm}]
		{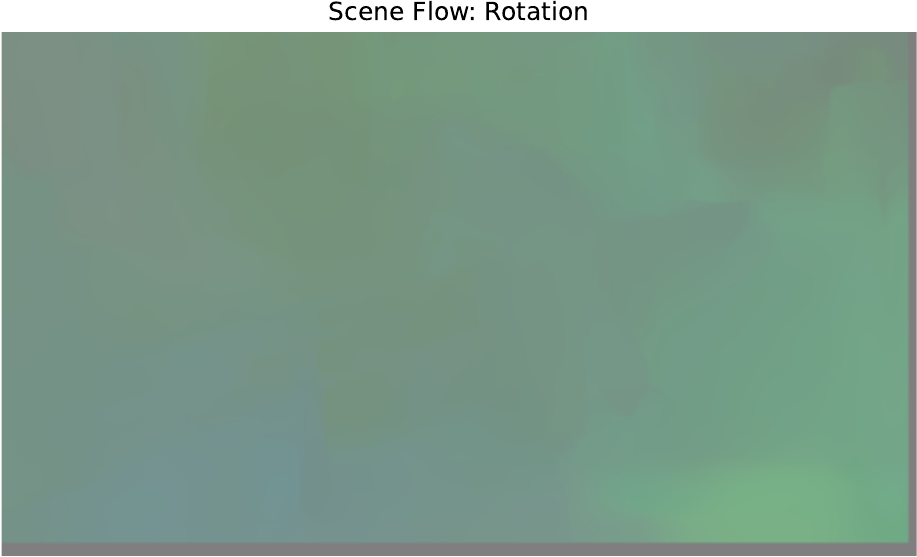}
	\end{overpic}\hspace*{-0.3em}
	\begin{overpic}[width=0.11\linewidth, height=1.25cm, tics=0, clip, trim={0.1cm 0.1cm 0.1cm 0.5cm}]
		{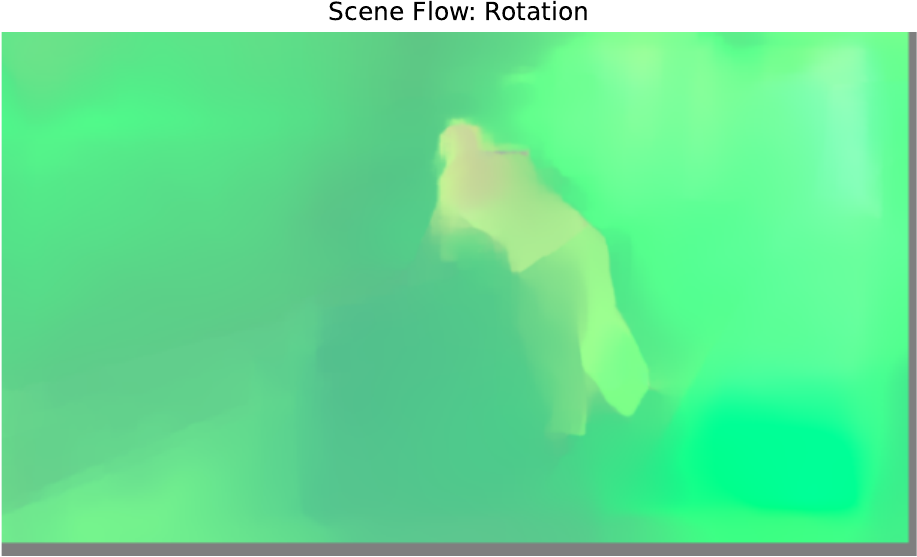}
	\end{overpic}
	\begin{overpic}[width=0.11\linewidth, height=1.25cm, tics=0, clip, trim={0.1cm 0.1cm 0.1cm 0.5cm}]
		{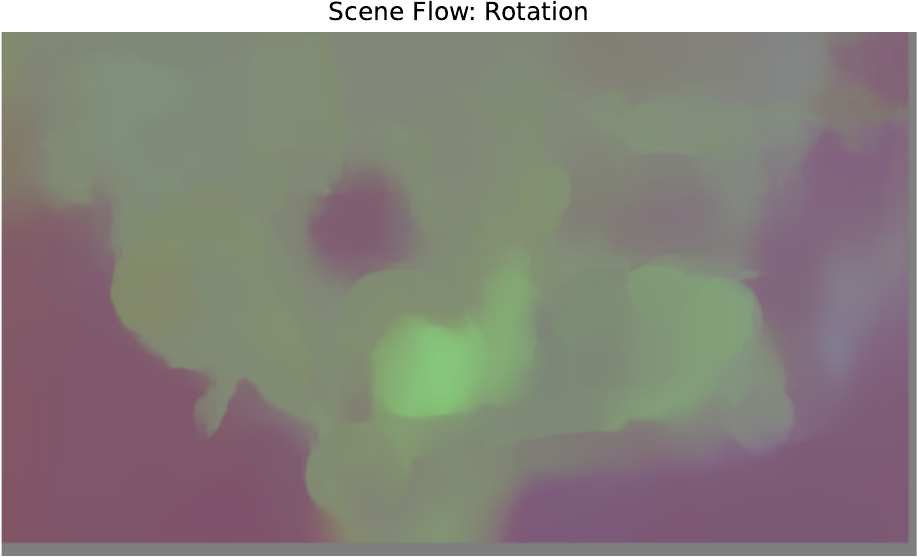} 
		
	\end{overpic}\hspace*{-0.3em}
	\begin{overpic}[width=0.11\linewidth, height=1.25cm, tics=0, clip, trim={0.1cm 0.1cm 0.1cm 0.5cm}]
		{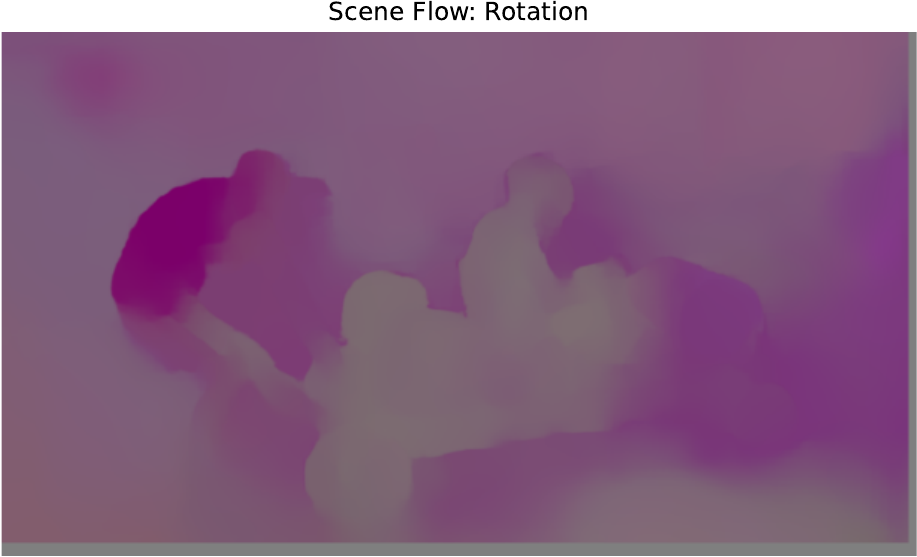}
	\end{overpic}

	\begin{overpic}[width=0.11\linewidth, height=1.25cm, tics=0, clip, trim={0.1cm 0.1cm 0.1cm 0.5cm}]
		{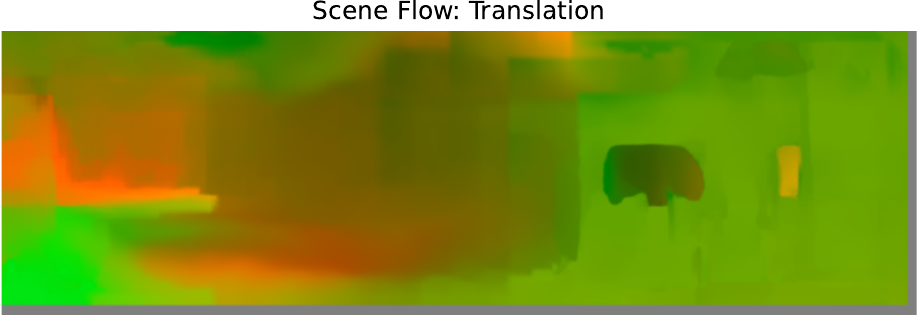}
	\end{overpic}\hspace*{-0.3em}
	\begin{overpic}[width=0.11\linewidth, height=1.25cm, tics=0, clip, trim={0.1cm 0.1cm 0.1cm 0.5cm}]
		{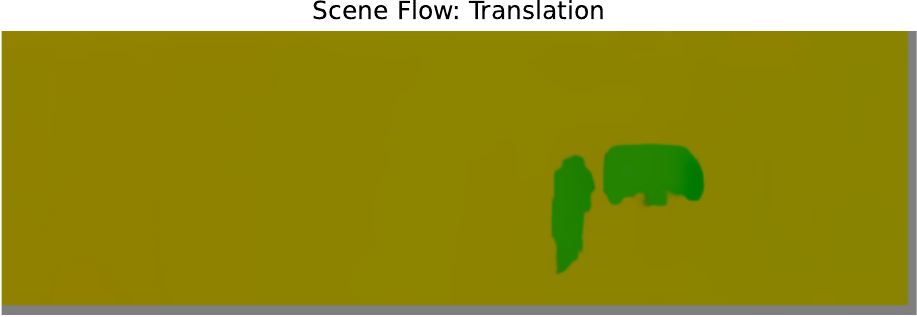}
	\end{overpic}
	\begin{overpic}[width=0.11\linewidth, height=1.25cm, tics=0, clip, trim={0.1cm 0.1cm 0.1cm 0.5cm}]
		{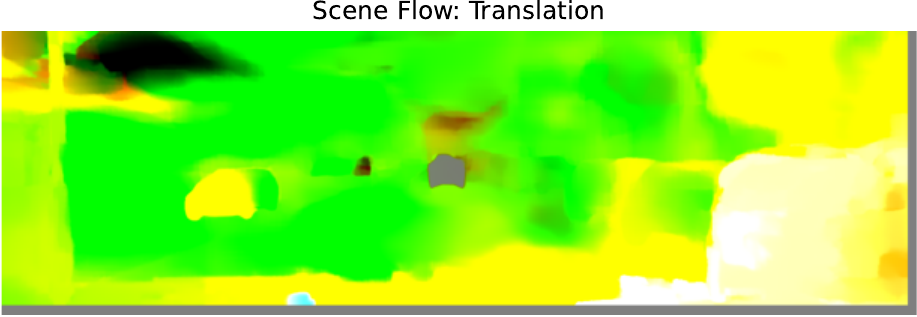}
	\end{overpic}\hspace*{-0.3em}
	\begin{overpic}[width=0.11\linewidth, height=1.25cm, tics=0, clip, trim={0.1cm 0.1cm 0.1cm 0.5cm}]
		{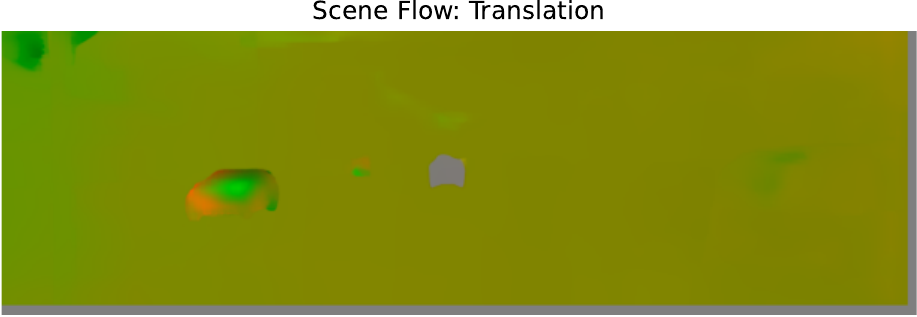}
	\end{overpic} 
	\begin{overpic}[width=0.11\linewidth, height=1.25cm, tics=0, clip, trim={0.1cm 0.1cm 0.1cm 0.5cm}]
		{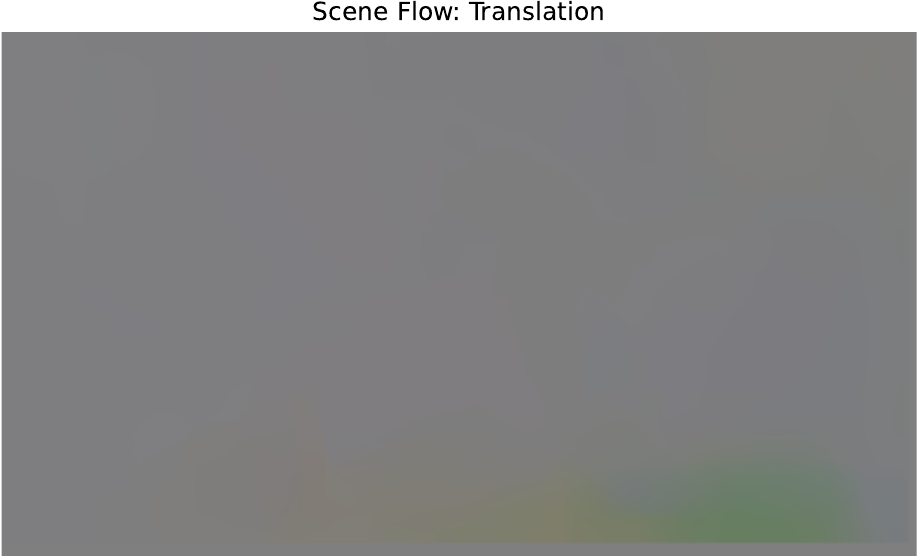}
	\end{overpic}\hspace*{-0.3em}
	\begin{overpic}[width=0.11\linewidth, height=1.25cm, tics=0, clip, trim={0.1cm 0.1cm 0.1cm 0.5cm}]
		{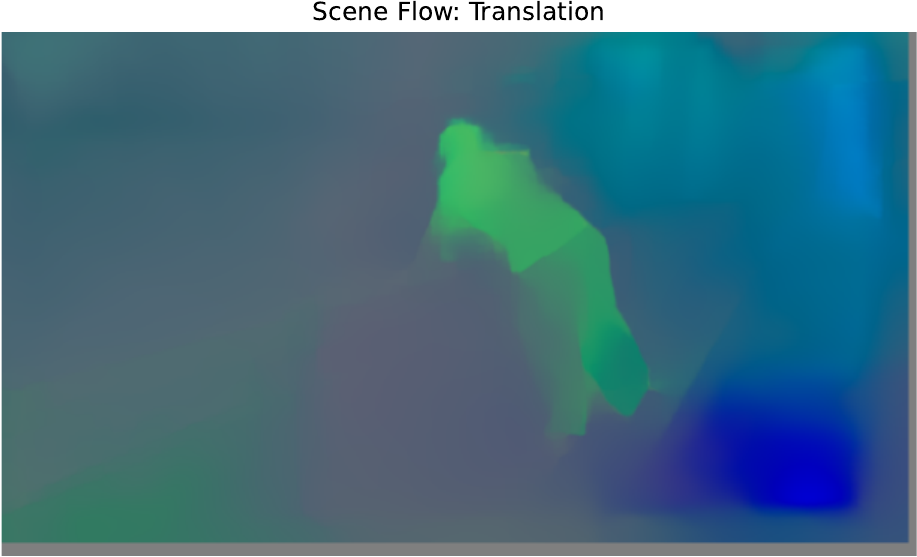}

	\end{overpic}
	\begin{overpic}[width=0.11\linewidth, height=1.25cm, tics=0, clip, trim={0.1cm 0.1cm 0.1cm 0.5cm}]
		{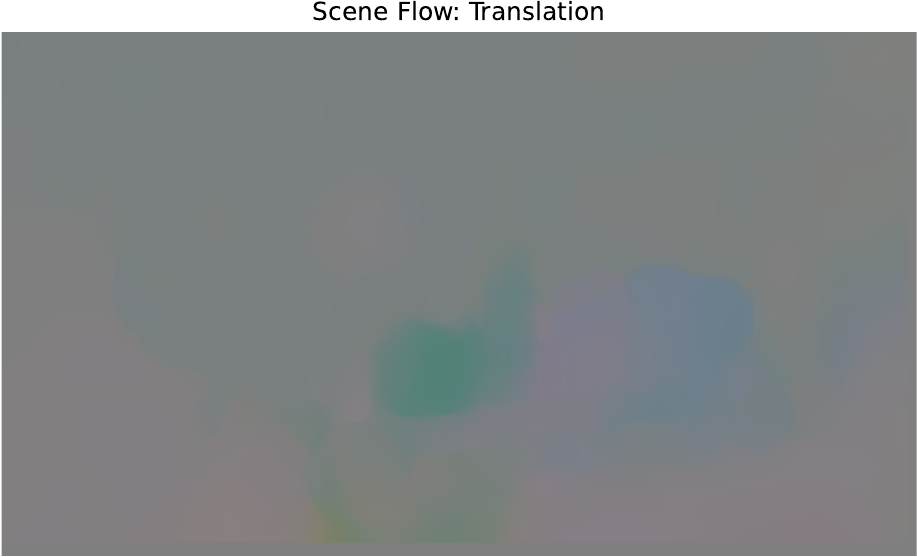} 
		
	\end{overpic}\hspace*{-0.3em}
	\begin{overpic}[width=0.11\linewidth, height=1.25cm, tics=0, clip, trim={0.1cm 0.1cm 0.1cm 0.5cm}]
		{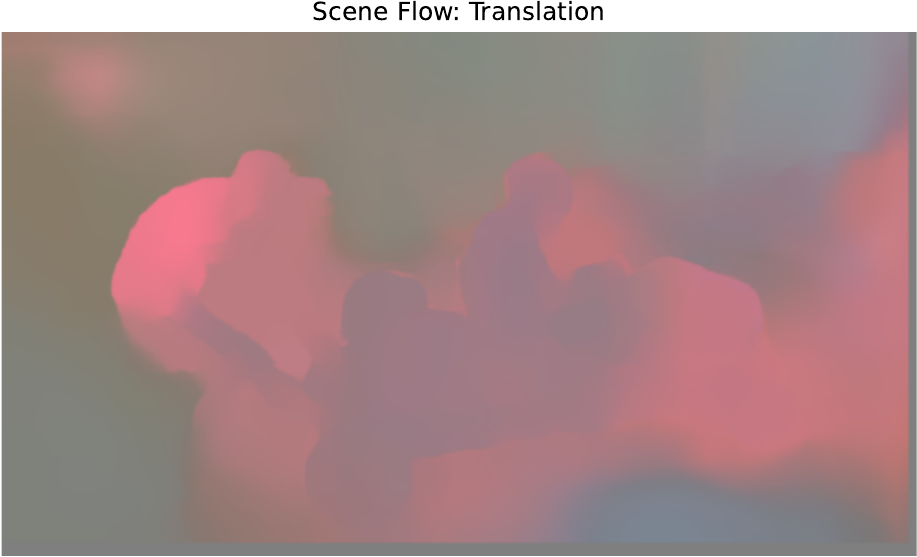}
	\end{overpic}

	\begin{overpic}[width=0.11\linewidth, height=1.5cm, tics=0, clip]
		{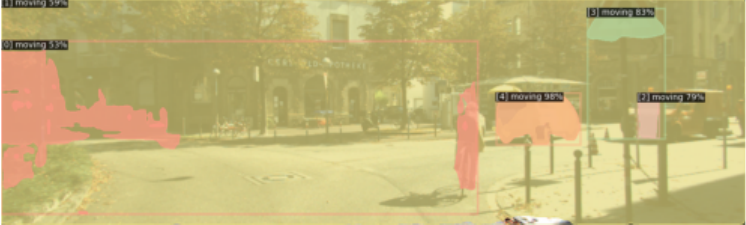}
		\put(-17, 15){\rotatebox{90}{\footnotesize{Ours SF}}}
	\end{overpic}\hspace*{-0.3em}
	\begin{overpic}[width=0.11\linewidth, height=1.5cm, tics=0, clip]
		{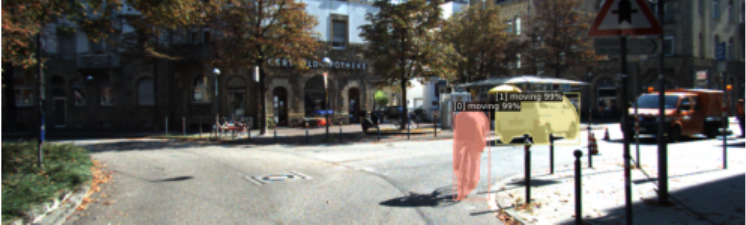}
	\end{overpic}
	\begin{overpic}[width=0.11\linewidth, height=1.5cm, tics=0, clip]
		{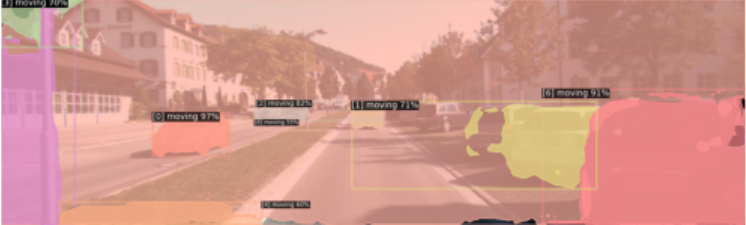}
	\end{overpic}\hspace*{-0.3em}
	\begin{overpic}[width=0.11\linewidth, height=1.5cm, tics=0, clip]
		{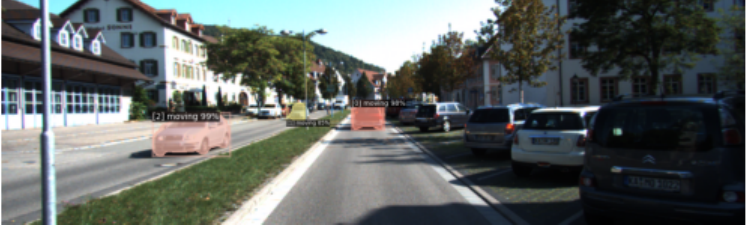}
	\end{overpic} 
	\begin{overpic}[width=0.11\linewidth, height=1.5cm, tics=0, clip]
		{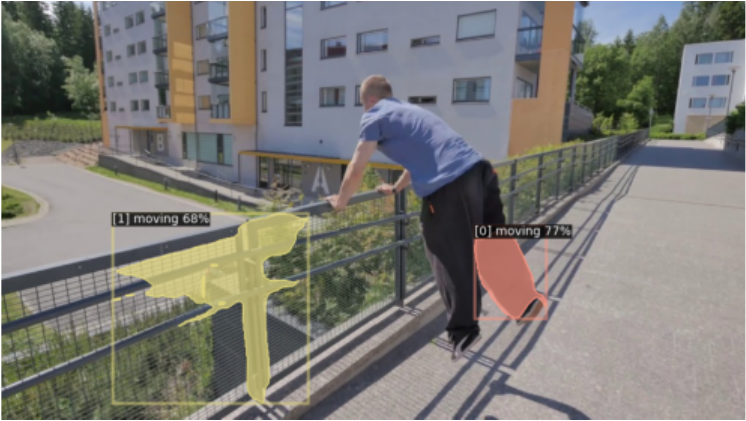} 
	\end{overpic}\hspace*{-0.3em}
	\begin{overpic}[width=0.11\linewidth, height=1.5cm, tics=0, clip]
		{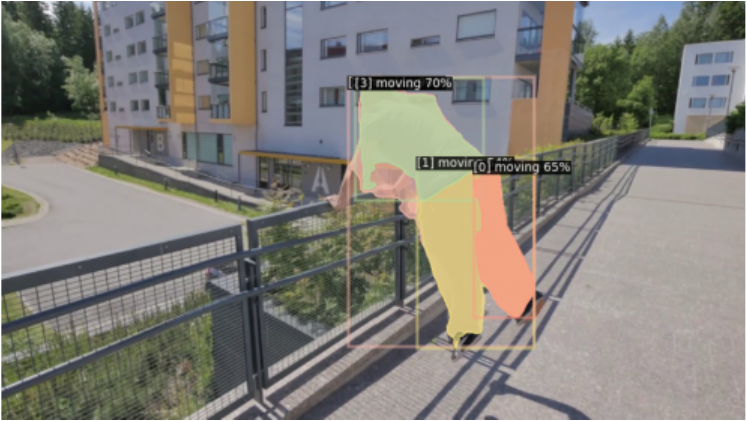}
	\end{overpic}
	\begin{overpic}[width=0.11\linewidth, height=1.5cm, tics=0, clip]
				{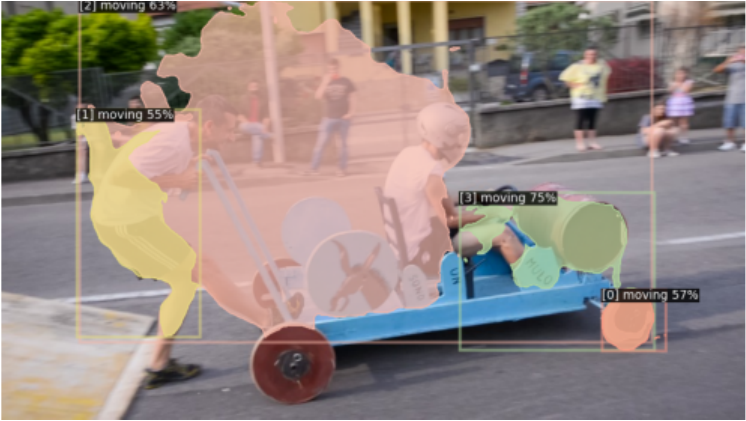} 
	\end{overpic}\hspace*{-0.3em}
	\begin{overpic}[width=0.11\linewidth, height=1.5cm, tics=0, clip]
				{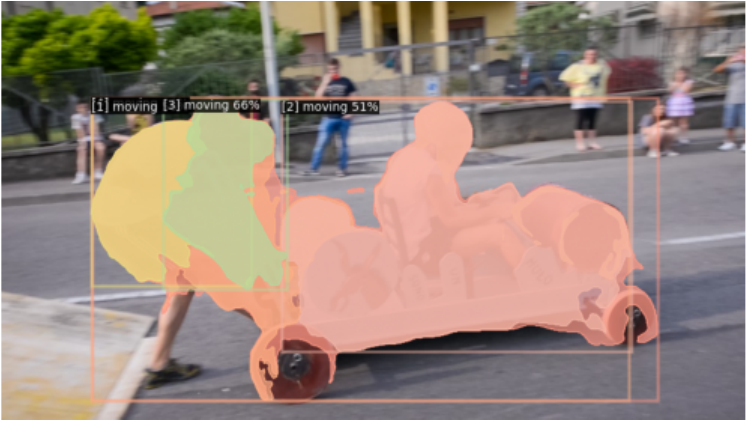}
	\end{overpic}

	\begin{overpic}[width=0.11\linewidth, height=1.5cm, tics=0, clip]
		{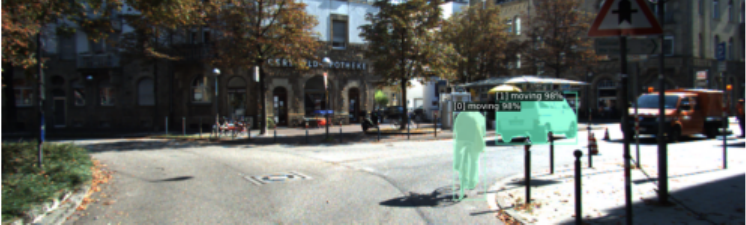}
		\put(-17, -6){\rotatebox{90}{\footnotesize{Ours RGB+SF}}}
	\end{overpic}\hspace*{-0.3em}
	\begin{overpic}[width=0.11\linewidth, height=1.5cm, tics=0, clip]
		{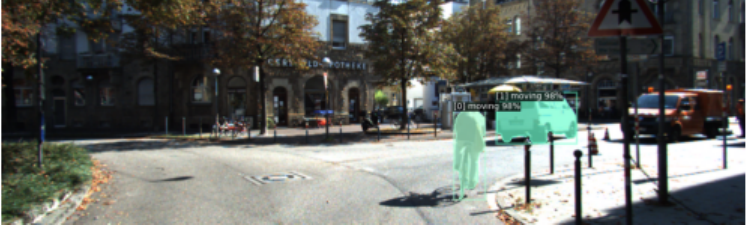}
	\end{overpic}
	\begin{overpic}[width=0.11\linewidth, height=1.5cm, tics=0, clip]
		{figures/benchmarks/mix3/rgb_sf/fuse_late_sf/kitti_5ep/abs/pred/0120.pdf}
	\end{overpic}\hspace*{-0.3em}
	\begin{overpic}[width=0.11\linewidth, height=1.5cm, tics=0, clip]
		{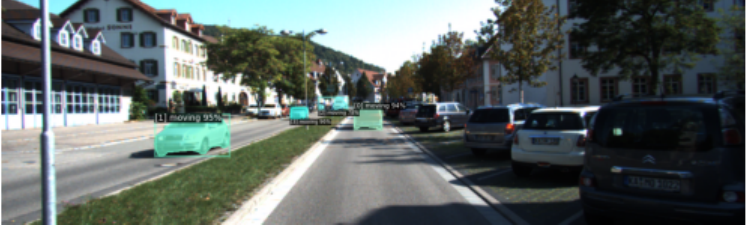}
	\end{overpic} 
	\begin{overpic}[width=0.11\linewidth, height=1.5cm, tics=0, clip]
		{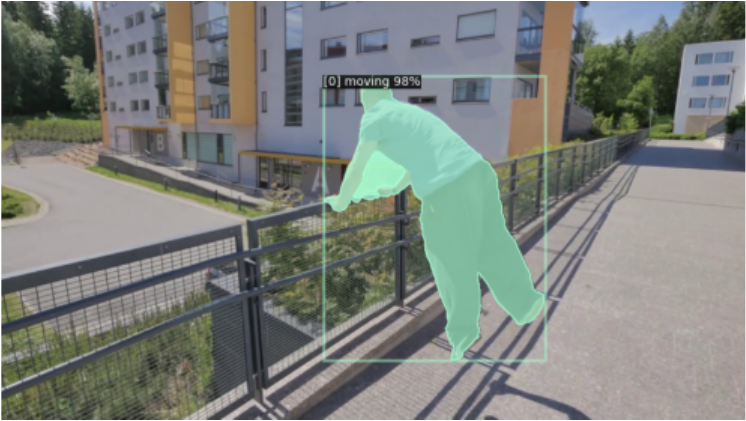}
	\end{overpic}\hspace*{-0.3em}
	\begin{overpic}[width=0.11\linewidth, height=1.5cm, tics=0, clip]
		{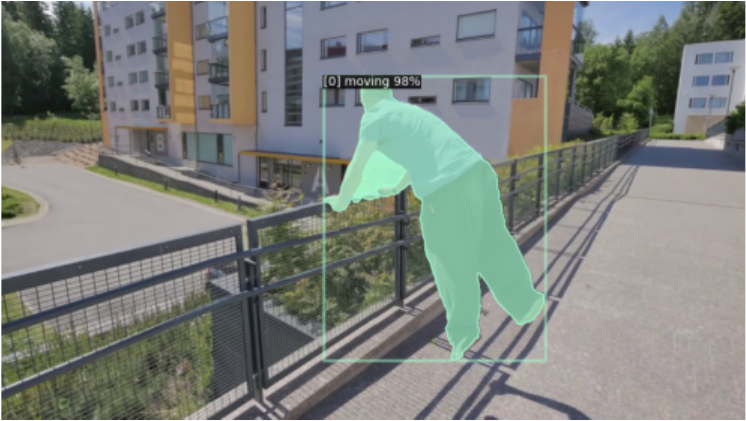}
	\end{overpic}
	\begin{overpic}[width=0.11\linewidth, height=1.5cm, tics=0, clip]		
		{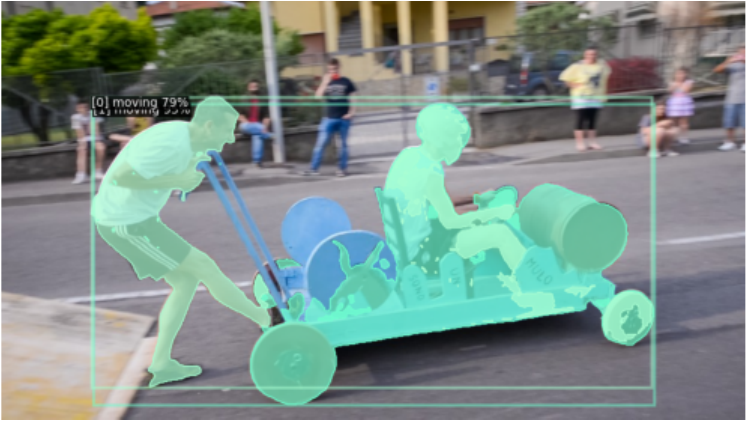}
	\end{overpic}\hspace*{-0.3em}
	\begin{overpic}[width=0.11\linewidth, height=1.5cm, tics=0, clip]		
		{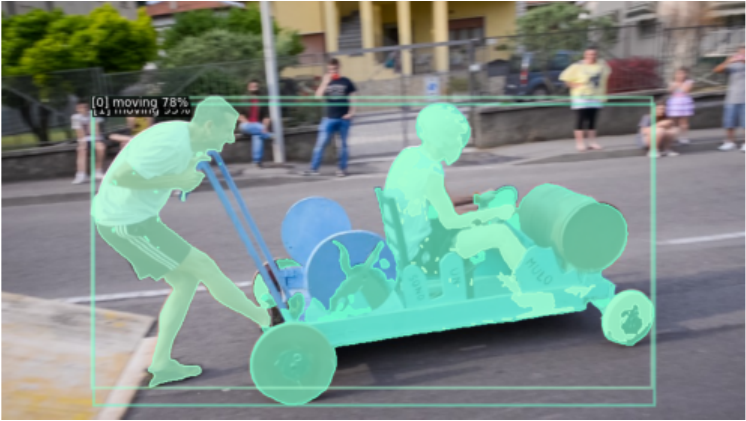} 
	\end{overpic}

	\hspace*{5em}
	\begin{overpic}[width=0.11\linewidth, height=1.5cm, tics=0, clip]
		{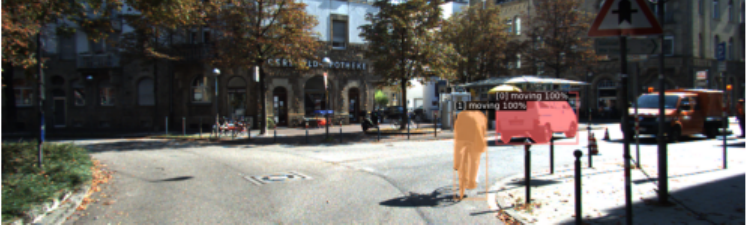}
		\put(-72, 35){\rotatebox{90}{\footnotesize{GT}}}
	\end{overpic}
	\hspace*{5em}
	\begin{overpic}[width=0.11\linewidth, height=1.5cm, tics=0, clip]
		{figures/benchmarks/gt/kitti/0120.pdf}
	\end{overpic}
	\hspace*{5em}
	\begin{overpic}[width=0.11\linewidth, height=1.5cm, tics=0, clip]
				{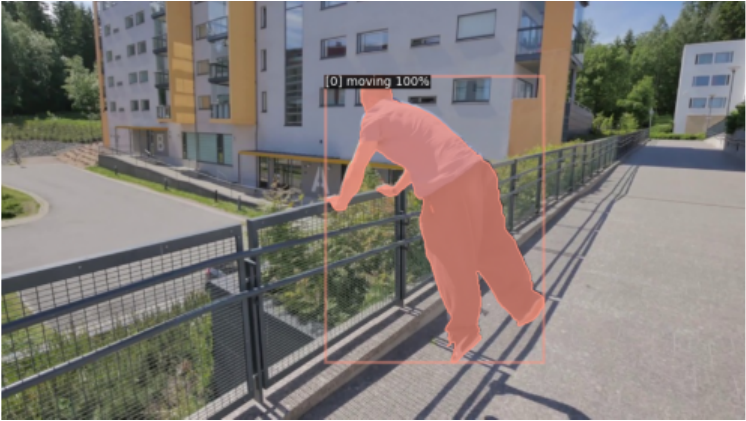}
	\end{overpic}
	\hspace*{5em}
	\begin{overpic}[width=0.11\linewidth, height=1.5cm, tics=0, clip]
				{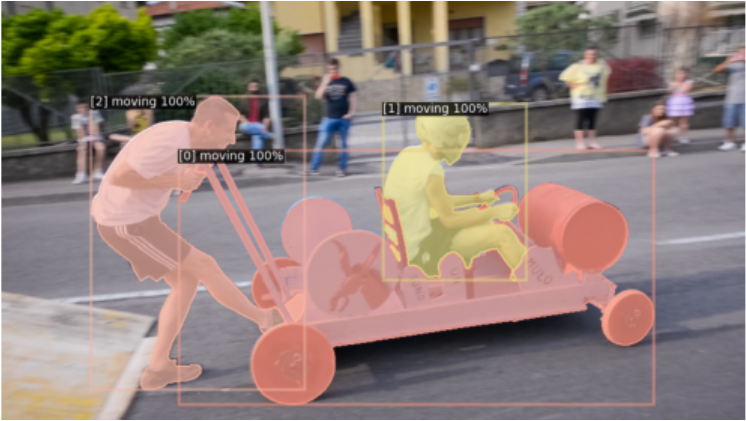}
	\end{overpic}
	\hspace*{5em}

	\caption{\textit{Quality matters} - Failures in motion estimation can ruin a segmentation. With 3D motion, this quality also depends on the additional depth prediction. On in-the-wild data, the depth map often lacks an absolute scale, which creates even more noise in the motion field. However, this noise can be compensated by combining with appearance data. We compare the quality of monocular and stereo depth on Kitti. On Davis we can correctly adjust the scales of single-image depth and improve the downstream motion segmentation when a reconstruction \cite{teed2021droid} is possible. Note how when using a two-stream model trained on Mix 3, the differences in the motion map stop to matter as the appearance stream mainly drives the segmentation.}
\label{fig:modalities}
\end{figure*}

\subsection{Reconstruction on DAVIS}
\label{sup:davis}
Accurate 3D motion estimates require scale-accurate and consistent depth, which standard single-image depth predictors \cite{ranftl2021vision} cannot provide. 
We use an end-to-end SLAM system \cite{teed2021droid} to create a map of the scene and recover the camera odometry. SLAM systems usually filter out dynamic contents and outliers which would corrupt the odometry estimates. In this manner, \cite{teed2021droid} estimates a confidence value $ c $ for the factor graph optimization both for x- and y-direction. We filter 
$ || \left[ c_{x},\; c_{y} \right] ||_{2} $ with a threshold $ \tau = 0.2 $ to reconstruct only the static scene. 
For each frame, we reproject a locally consistent window of $ 5 $ surrounding frames to create a consistent, \textit{static} reference depth map $ Z_{ref} $. Similar to \cite{ranftl2021vision}, we align each monocular depth prediction to the reference frame by estimating shift and scale parameters. It can be seen in Figure \ref{fig:modalities} on the right side that noise of the scene flow estimates (especially in the translation part of the rigid body motions) can be reduced with this strategy. However, reconstruction of casual videos is still an open problem \cite{liu2023robust} and therefore a reconstruction is not possible on all video clips of DAVIS. 
Nonetheless, this acts as a \textit{proof-of-concept} so that when such a reconstruction is possible, we can improve motion segmentation as well. Finally, our results in Figure \ref{fig:modalities} show that errors in motion estimates can be compensated very well with appearance data when the training data allows it. 

\section{Additional Results}
\label{sup:ablat}

\subsection{Ablation Fusion Strategies}
\label{sup:fusion_ablation}
We ablate different fusion mechanisms for image and optical flow input data and measure the effect of using different training data. Results can be seen in Table \ref{tab:ablation_fusion}. During our initial experiments we did not find consistent performance gains from a single fusion strategy across different i) datasets ii) input modalities. For this reason, we choose to focus on late fusion in the decoder (D) or fusion at all locations (E+D) for all dataset mixes. While bottleneck tokens reduce both time and memory complexity, the performance lacks behind a naive fusion strategy. (Early) Fusion with deformable attention in the encoder can be very effective. We hypothesize that fusion in our architecture with the attention mechanism offers high degrees of freedom. This affects training dynamics considerably. We believe that similar results could be achieved with all strategies when training for long enough. Differences in this ablation experiment could be solely observable due to the training time cutoff. 

\begin{table}[h!]
	\centering
	\setstackgap{L}{9pt}
	\resizebox{0.5\textwidth}{!}{%
		\begin{tabular}{lccccccc}
			\toprule
			& & \multicolumn{3}{c}{\textbf{Kitti}} & \multicolumn{3}{c}{\textbf{Davis}} \\
			\Centerstack[c]{Training \\data} & \Centerstack[c]{Fusion \\mechanism} & \cellcolor{Dandelion} $ AP_{50} \uparrow $ & \cellcolor{Dandelion} FP $ \downarrow $ & \cellcolor{Dandelion} FN $ \downarrow $ & \cellcolor{BlueGreen} $ AP_{50} \uparrow $ & \cellcolor{BlueGreen} FP $ \downarrow $ & \cellcolor{BlueGreen} FN $ \downarrow $ \\
			\midrule
			& \cellcolor{lighter-gray} Encoder & \cellcolor{lighter-gray} \textbf{46.18} & \cellcolor{lighter-gray} 0.17 & \cellcolor{lighter-gray} \textbf{0.37} & \cellcolor{lighter-gray} \textbf{36.1} & \cellcolor{lighter-gray} \textbf{0.24} & \cellcolor{lighter-gray}  \underline{0.15} \\
			& Decoder & 37.16 & \underline{0.12} & 0.42 & 23.98 & 0.39 & 0.16 \\
			& \cellcolor{lighter-gray} \Centerstack[c]{Decoder \\MBT \cite{nagrani2021attention}} &  \cellcolor{lighter-gray} 34.23 & \cellcolor{lighter-gray} \textcolor{red}{\textbf{0.08}} & \cellcolor{lighter-gray} 0.45 & \cellcolor{lighter-gray} 21.5 & \cellcolor{lighter-gray} 0.41 & \cellcolor{lighter-gray} 0.16 \\
			\multirow{-4}{*}{Mix 1} & Encoder + Decoder & \underline{39.65} & 0.15 & \underline{0.40} & \underline{35.88} & \underline{0.29} & \textbf{0.15} \\
			\midrule
			& \cellcolor{lighter-gray} Decoder & \cellcolor{lighter-gray} \textbf{64.27} & \cellcolor{lighter-gray} 0.32 & \cellcolor{lighter-gray} \textbf{0.26} & \cellcolor{lighter-gray} \textcolor{red}{\textbf{65.82}} & \cellcolor{lighter-gray} \textbf{0.16} & \cellcolor{lighter-gray} 0.11 \\
			\multirow{-2}{*}{Mix 2} & Encoder + Decoder & 56.41 & \textbf{0.13} & 0.35 & 65.81 & 0.25 & \textbf{0.09} \\
			\midrule
			& \cellcolor{lighter-gray} Decoder & \cellcolor{lighter-gray} \textcolor{red}{\textbf{70.82}} & \cellcolor{lighter-gray} 0.32 & \cellcolor{lighter-gray} \textcolor{red}{\textbf{0.16}} & \cellcolor{lighter-gray} 54.01 & \cellcolor{lighter-gray} 0.12 & \cellcolor{lighter-gray} \textcolor{red}{\textbf{0.01}} \\
			\multirow{-2}{*}{Mix 3} & Encoder + Decoder & 60.88 & \textbf{0.29} & 0.24 & \textbf{61.12} & \textcolor{red}{\textbf{0.11}} & 0.12 \\
			\bottomrule
	\end{tabular}}
	\caption{Use of different fusion mechanisms on image and optical flow data.}
	\label{tab:ablation_fusion}
\end{table}

\newpage
\section{More Visualizations}
\label{sup:visu}
In this section we add more visualizations to better explain our model behavior. We give further examples of the attention in both streams, failure cases, differences between training data mixes and generalization on the Moving Camouflaged Animal (MoCA) dataset \cite{lamdouar2020betrayed}.

\subsection{Multi-modal Attention}
\label{sup:attention}
\begin{figure}[h!]
	\centering
	\begin{overpic}[width=3.5cm, height=2.3cm, trim={0.5cm 2.0cm 0.5cm 2.0cm}, clip, tics=10]
		{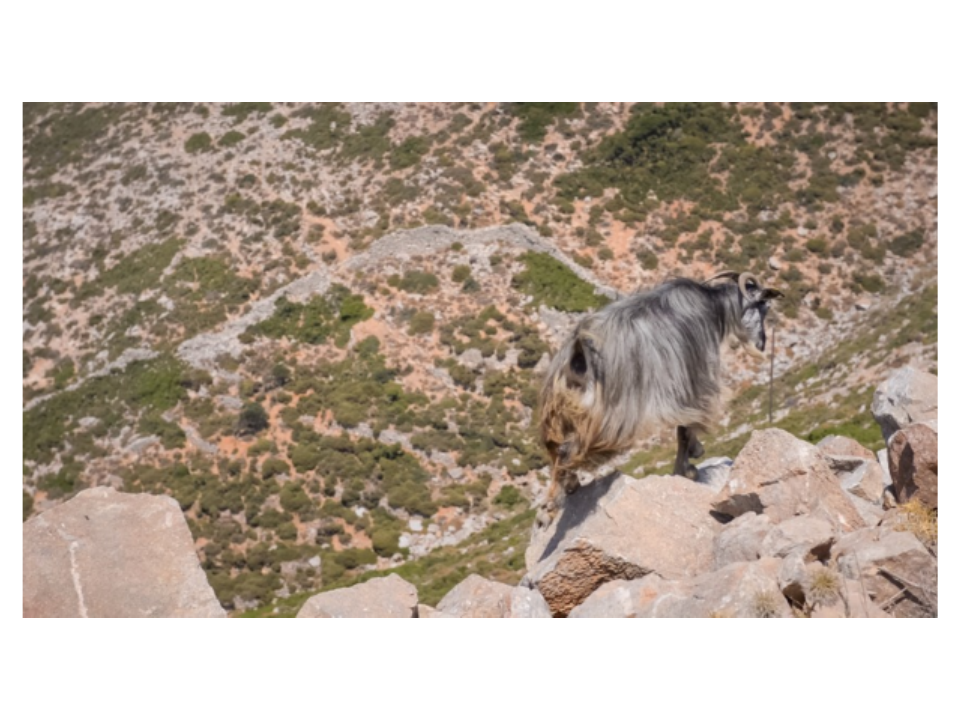}
		\put(30, 70){\footnotesize{Appearance}}
	\end{overpic}
	\begin{overpic}[width=3.5cm, height=2.3cm, trim={0.5cm 2.0cm 0.5cm 2.0cm}, clip, tics=10]
		{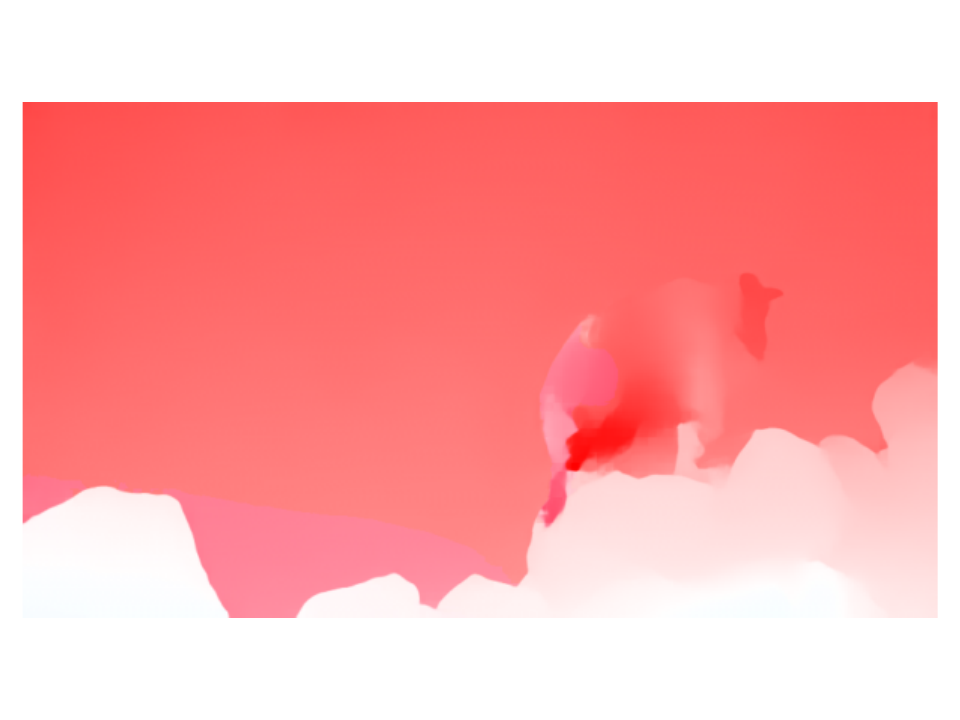}
		\put(30, 70){\footnotesize{Optical Flow}}
	\end{overpic}
	\\[\smallskipamount]

	\begin{overpic}[width=3.5cm, height=2.3cm, trim={2.0cm 3.0cm 2.0cm 3.0cm}, clip, tics=10]
		{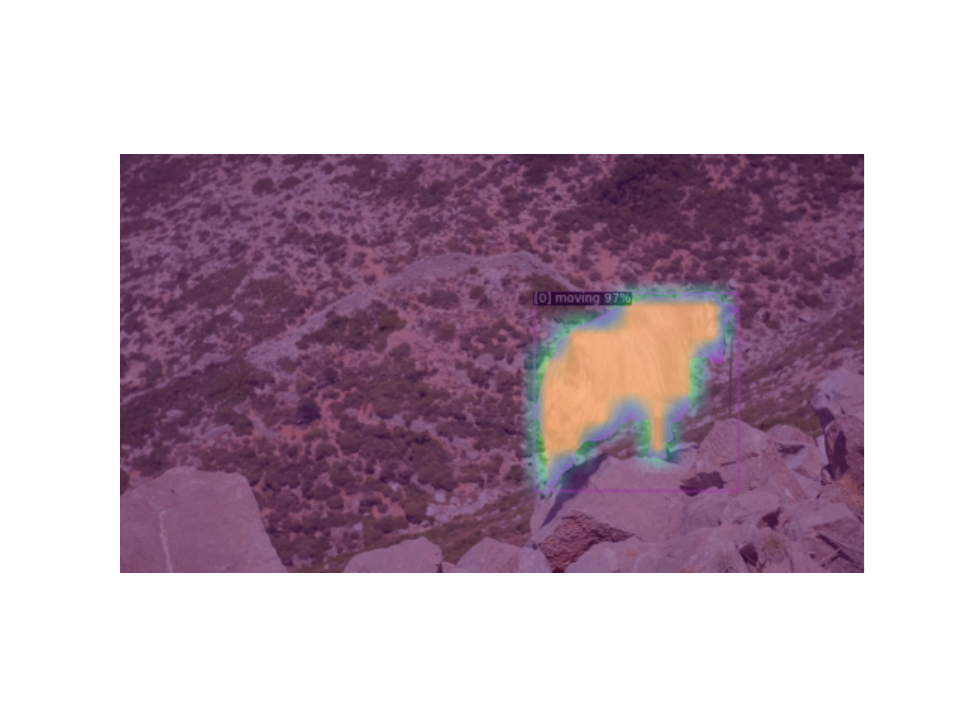}
		\put(-8, 3){\rotatebox{90}{\footnotesize{Masked Attention}}}
	\end{overpic}
	\begin{overpic}[width=3.5cm, height=2.3cm, trim={2.0cm 3.0cm 2.0cm 3.0cm}, clip, tics=10]
		{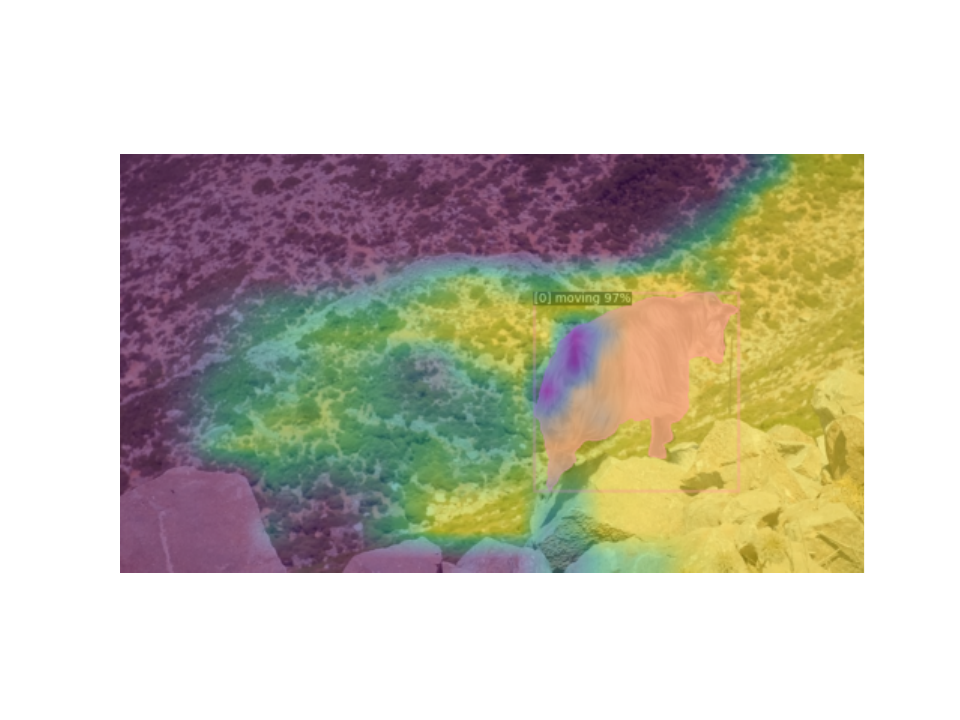}
	\end{overpic}
	\\[\smallskipamount]

	\begin{overpic}[width=3.5cm, height=2.3cm, trim={0.5cm 1.8cm 0.7cm 1.7cm}, clip, tics=10]
		{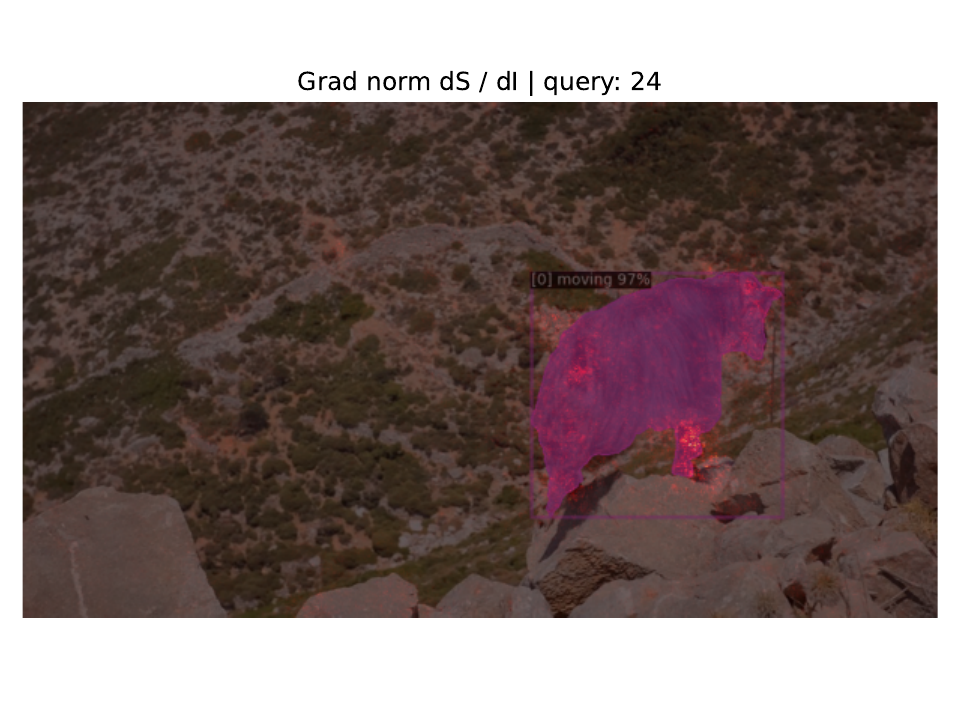}
		\put(-8, 8){\rotatebox{90}{\footnotesize{Gradient norm}}}
	\end{overpic}
	\begin{overpic}[width=3.5cm, height=2.3cm, trim={0.5cm 1.8cm 0.7cm 1.7cm}, clip, tics=10]
		{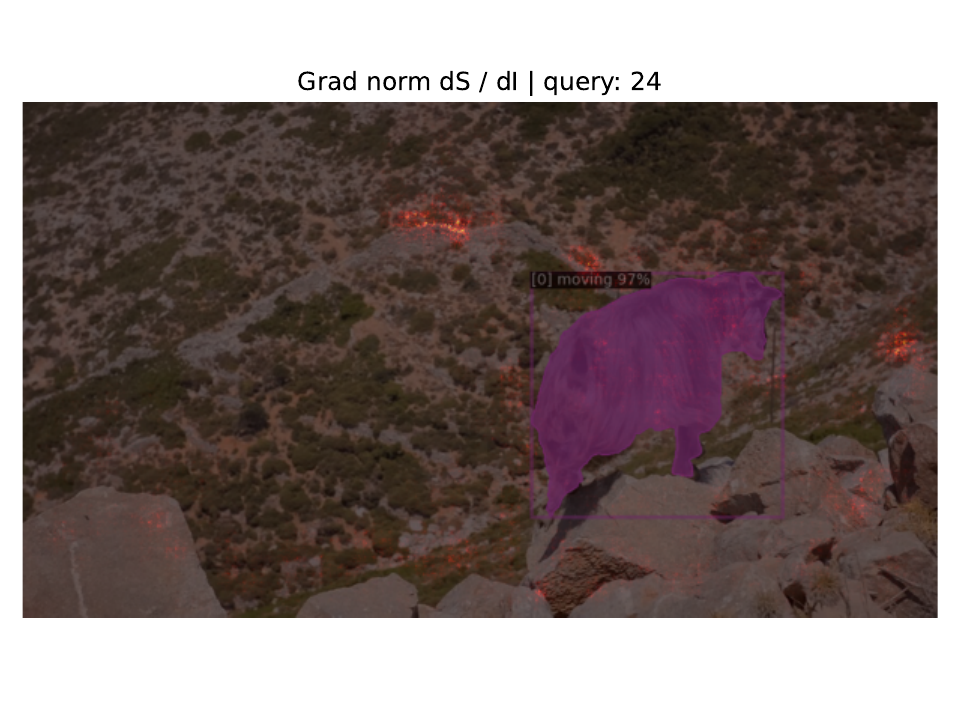}
	\end{overpic}

	\caption{\textit{What does our network see?} \textbf{a)} Multi-modal attention in our two-stream motion segmentation architecture: Masked-attention in the individual decoders. \textbf{b)} Gradient norm $ || \frac{\partial S}{\partial I} || $ of the segmentation masks $ S $ w.r.t to input data $ I $.  While the appearance stream is focused on recognizing objects, motion is more global to distinguish between moving foreground and background.
	}
	\label{fig:attention}
\end{figure}

Figure \ref{fig:attention} shows the learned attention masks from both streams in our model. We further show the gradients of our output w.r.t the input data. It can be seen how different streams focus on different parts of the image to come to an output. 

%
%
%

\clearpage

\subsection{Why Diverse Training Data Is Necessary}
\label{sup:dataset_mixes}
\begin{figure*}
	\centering
	\begin{overpic}[width=.19\linewidth, height=1.5cm, trim={0.0cm 0.05cm 0.0cm 0.0cm}, tics=0, clip]
		{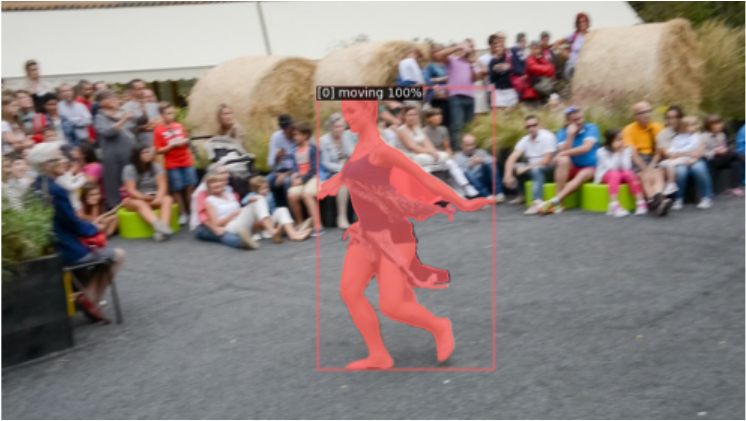}
		\put(40, 50){\footnotesize{GT}}
		\put(125, 50){\footnotesize{Optical Flow}}
		\put(247, 50){\footnotesize{Mix 1}}
		\put(347, 50){\footnotesize{Mix 2}}
		\put(445, 50){\footnotesize{Mix 3}}
	\end{overpic}\hspace*{-0.3em}
	\begin{overpic}[width=.19\linewidth, height=1.5cm, trim={0.0cm 0.05cm 0.0cm 0.6cm}, tics=0, clip]
		{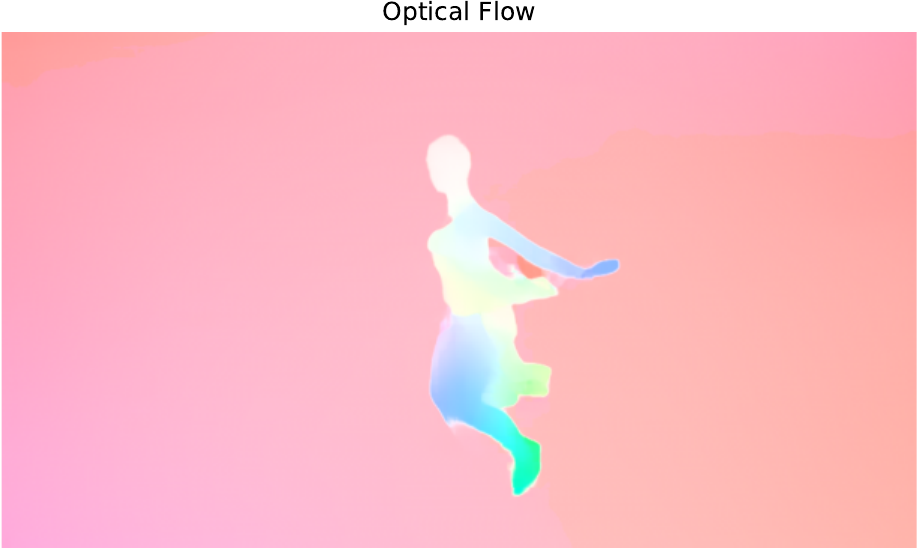}
	\end{overpic}
	\hspace{0.05cm}
	\begin{overpic}[width=.19\linewidth, height=1.5cm, trim={0.0cm 0.0cm 0.0cm 0.0cm}, tics=0, clip]
		{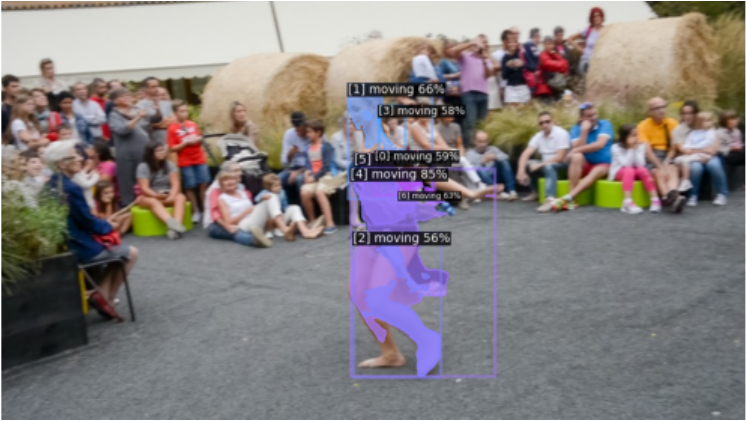}
	\end{overpic}\hspace*{-0.3em}
	\begin{overpic}[width=.19\linewidth, height=1.5cm, trim={0.0cm 0.0cm 0.0cm 0.0cm}, tics=0, clip]
		{figures/benchmarks/mix2/fuse_all_of/davis_10ep/pred/0030.pdf}
	\end{overpic}\hspace*{-0.3em}
	\begin{overpic}[width=.19\linewidth, height=1.5cm, trim={0.0cm 0.0cm 0.0cm 0.0cm}, tics=0, clip]
		{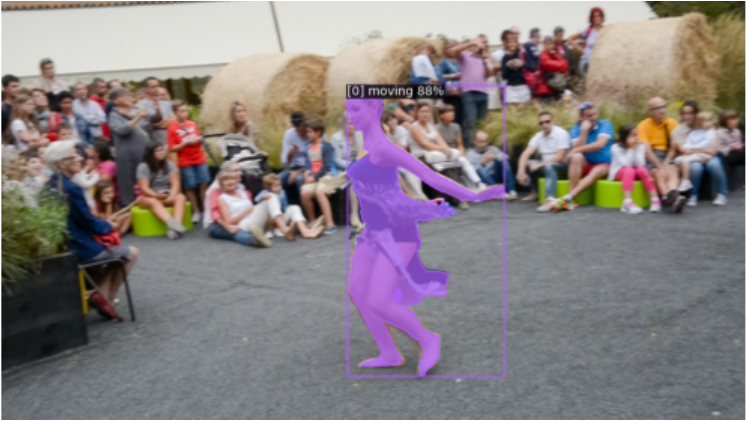}
	\end{overpic}
	\\[0.1em]
	
	\begin{overpic}[width=.19\linewidth, height=1.5cm, trim={0.0cm 0.05cm 0.0cm 0.0cm}, tics=0, clip]
		{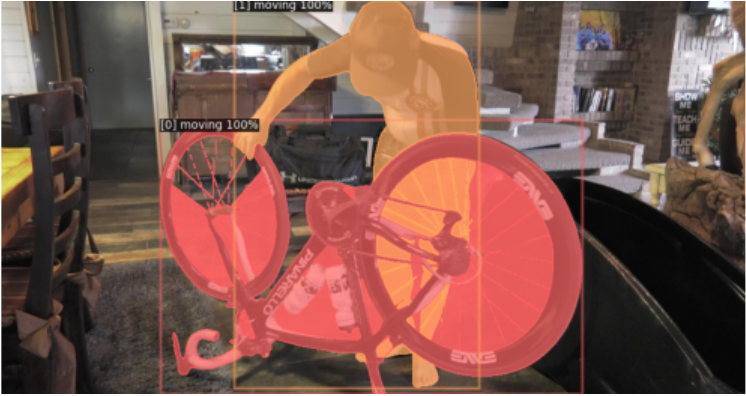}
	\end{overpic}\hspace*{-0.3em}
	\begin{overpic}[width=.19\linewidth, height=1.5cm, trim={0.0cm 0.05cm 0.0cm 0.54cm}, tics=0, clip]
		{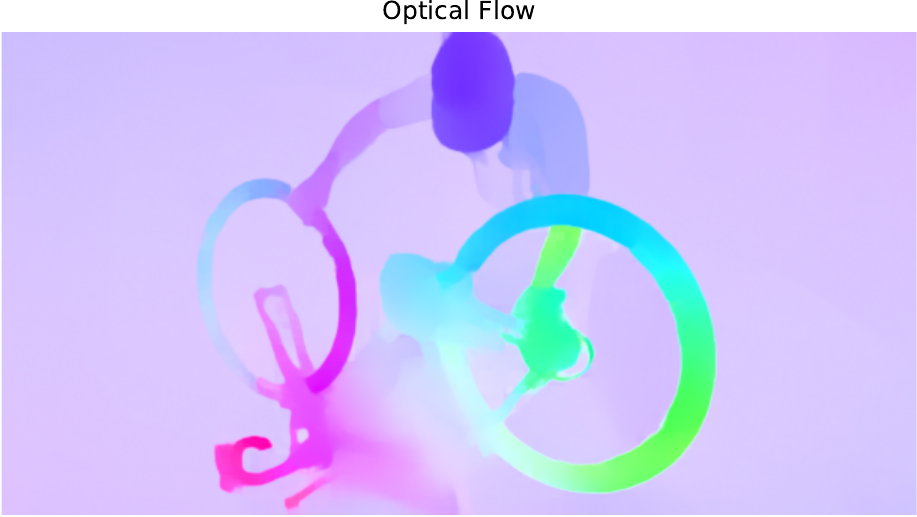}
	\end{overpic}
	\hspace{0.05cm}
	\begin{overpic}[width=.19\linewidth, height=1.5cm, trim={0.0cm 0.0cm 0.0cm 0.0cm}, tics=0, clip]
		{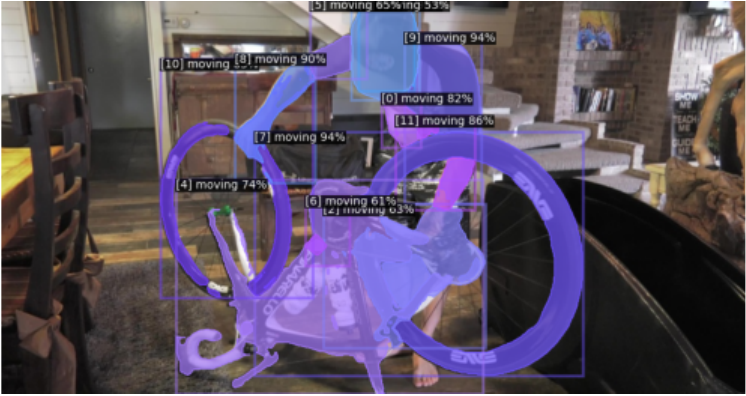}
	\end{overpic}\hspace*{-0.3em}
	\begin{overpic}[width=.19\linewidth, height=1.5cm, trim={0.0cm 0.0cm 0.0cm 0.0cm}, tics=0, clip]		
		{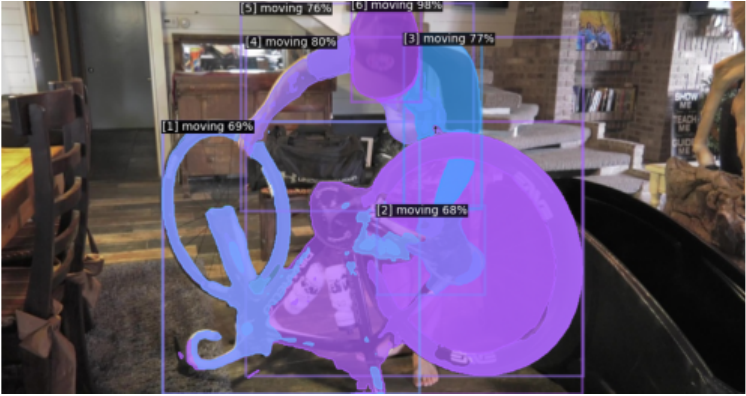}
	\end{overpic}\hspace*{-0.3em}
	\begin{overpic}[width=.19\linewidth, height=1.5cm, trim={0.0cm 0.0cm 0.0cm 0.0cm}, tics=0, clip]
		{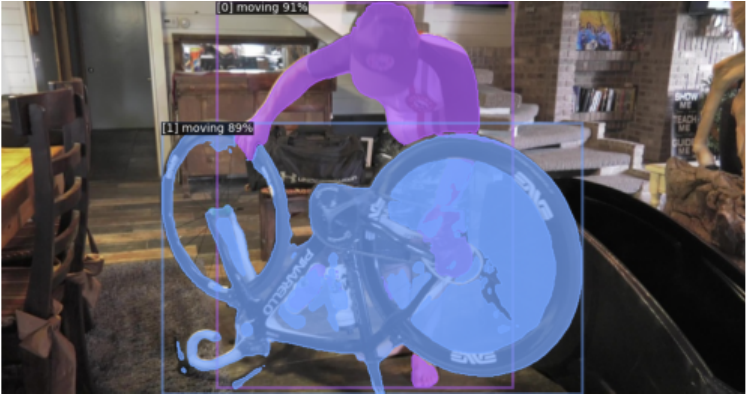}
	\end{overpic}
	\\[0.1em]	
	
	\begin{overpic}[width=.19\linewidth, height=1.5cm, clip, tics=0, clip, trim={0.0cm 0.0cm 0.0cm 0.0cm}]
		{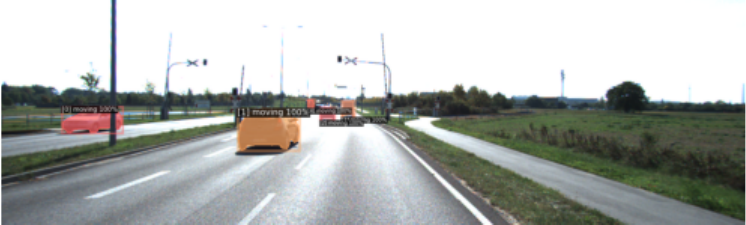}
	\end{overpic}\hspace*{-0.3em}
	\begin{overpic}[width=.19\linewidth, height=1.5cm, trim={0.0cm 0.0cm 0.0cm 0.6cm}, tics=0, clip]
		{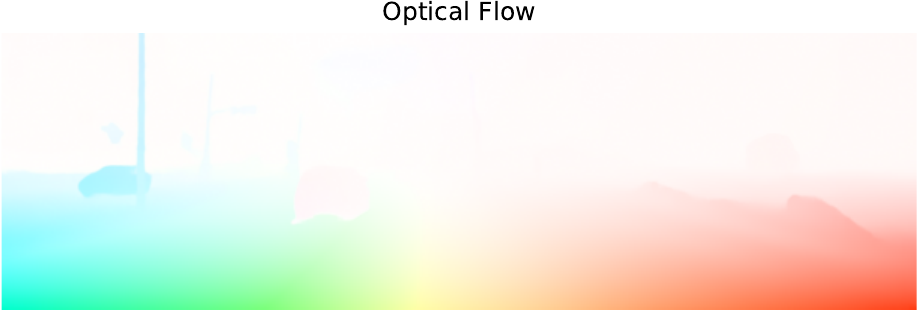}
	\end{overpic}
	\hspace{0.05cm}
	\begin{overpic}[width=.19\linewidth, height=1.5cm, trim={0.0cm 0.0cm 0.0cm 0.0cm}, tics=0, clip]
		{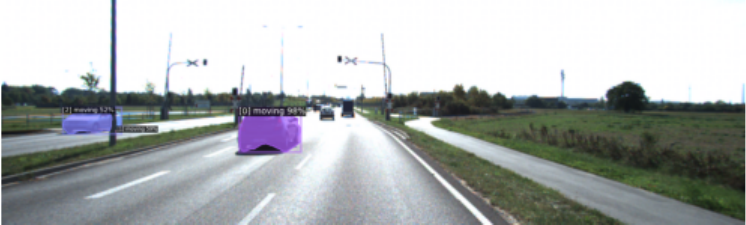}
	\end{overpic}\hspace*{-0.3em}
	\begin{overpic}[width=.19\linewidth, height=1.5cm, trim={0.0cm 0.0cm 0.0cm 0.0cm}, tics=0, clip]		
		{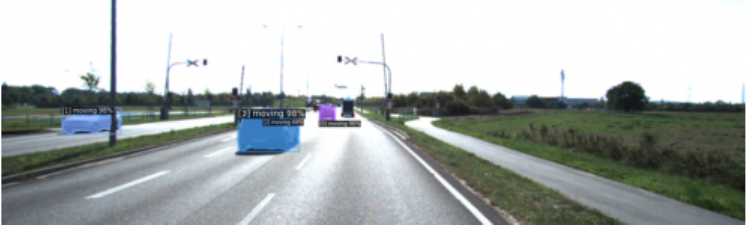}
	\end{overpic}\hspace*{-0.3em}
	\begin{overpic}[width=.19\linewidth, height=1.5cm, trim={0.0cm 0.0cm 0.0cm 0.0cm}, tics=0, clip]
		{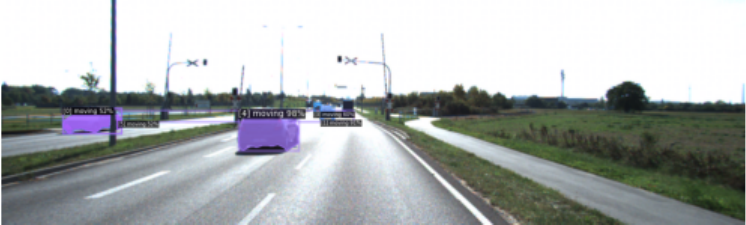}
	\end{overpic}
	\\[0.1em]	
	
	\begin{overpic}[width=.19\linewidth, height=1.5cm, clip, tics=0, clip, trim={0.0cm 0.0cm 0.0cm 0.0cm}]
		{figures/benchmarks/gt/kitti/0120.pdf}
	\end{overpic}\hspace*{-0.3em}
	\begin{overpic}[width=.19\linewidth, height=1.5cm, trim={0.0cm 0.0cm 0.0cm 0.6cm}, tics=0, clip]
		{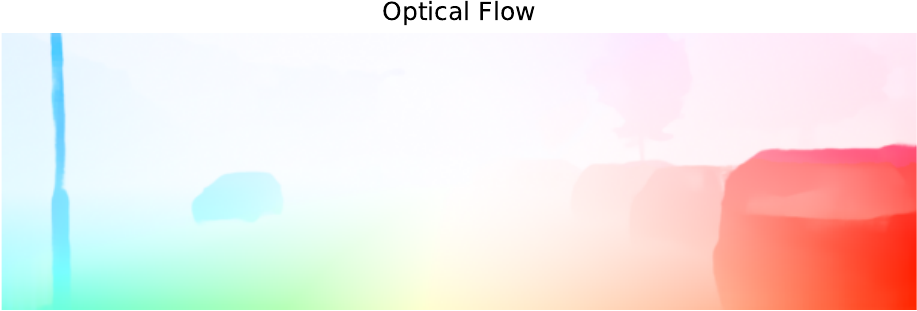}
	\end{overpic}
	\hspace{0.05cm}
	\begin{overpic}[width=.19\linewidth, height=1.5cm, trim={0.0cm 0.0cm 0.0cm 0.0cm}, tics=0, clip]
		{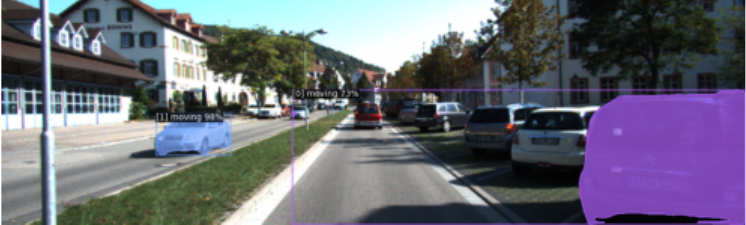}
	\end{overpic}\hspace*{-0.3em}
	\begin{overpic}[width=.19\linewidth, height=1.5cm, trim={0.0cm 0.0cm 0.0cm 0.0cm}, tics=0, clip]		
		{figures/benchmarks/mix2/fuse_all_of/kitti_10ep/pred/0120.pdf}
	\end{overpic}\hspace*{-0.3em}
	\begin{overpic}[width=.19\linewidth, height=1.5cm, trim={0.0cm 0.0cm 0.0cm 0.0cm}, tics=0, clip]
		{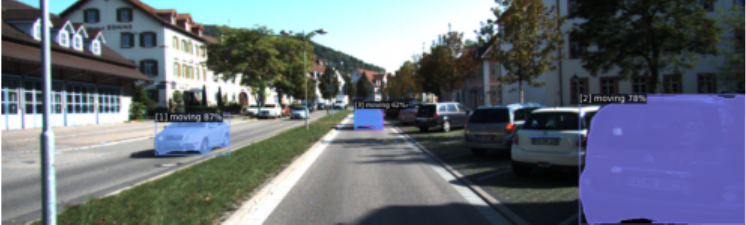}
	\end{overpic}
	\\[0.1em]
	
	\begin{overpic}[width=.19\linewidth, height=1.5cm, clip, tics=0, clip, trim={0.0cm 0.0cm 0.0cm 0.0cm}]
		{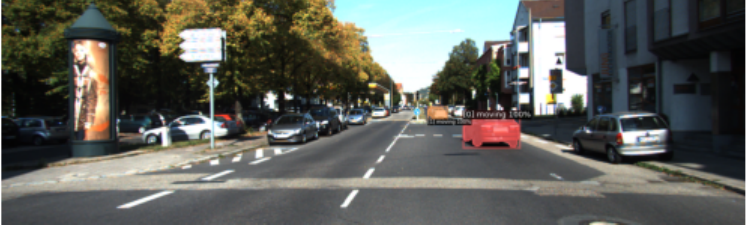}
	\end{overpic}\hspace*{-0.3em}
	\begin{overpic}[width=.19\linewidth, height=1.5cm, trim={0.0cm 0.0cm 0.0cm 0.6cm}, tics=0, clip]
		{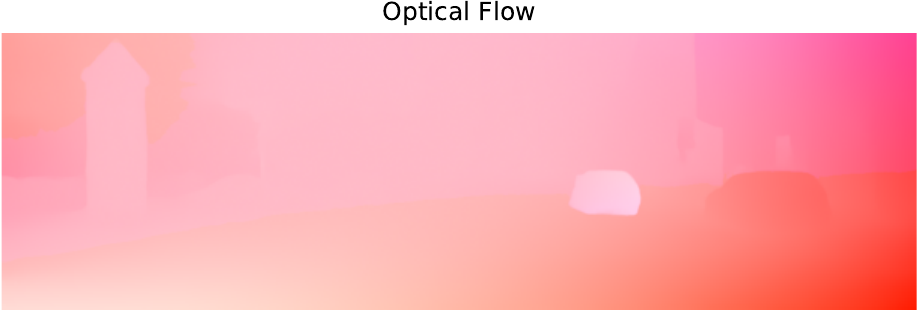}
	\end{overpic}
	\hspace{0.05cm}
	\begin{overpic}[width=.19\linewidth, height=1.5cm, trim={0.0cm 0.0cm 0.0cm 0.0cm}, tics=0, clip]
		{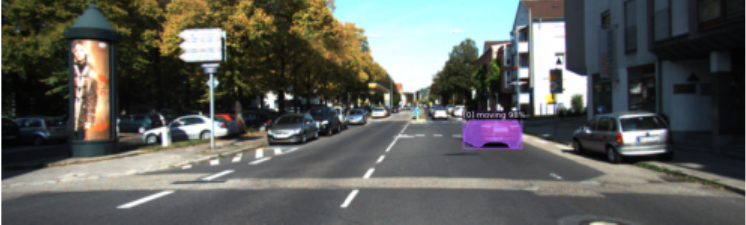}
	\end{overpic}\hspace*{-0.3em}
	\begin{overpic}[width=.19\linewidth, height=1.5cm, trim={0.0cm 0.0cm 0.0cm 0.0cm}, tics=0, clip]		
		{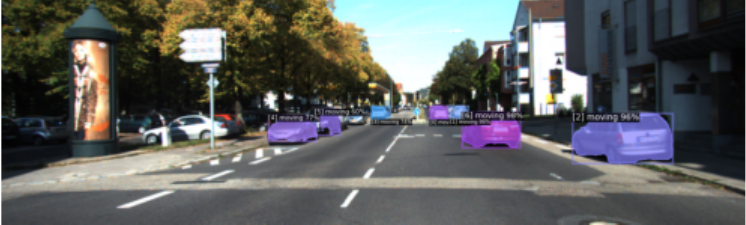}
	\end{overpic}\hspace*{-0.3em}
	\begin{overpic}[width=.19\linewidth, height=1.5cm, trim={0.0cm 0.0cm 0.0cm 0.0cm}, tics=0, clip]
		{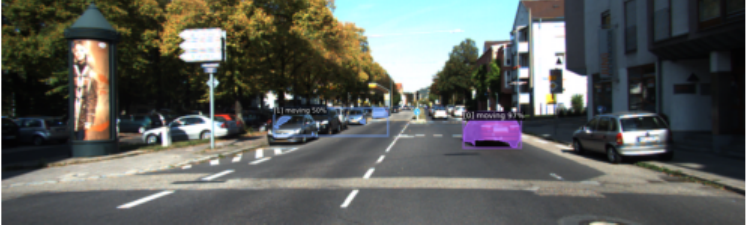}
	\end{overpic}
	\\[0.1em]

\caption{\textit{Effect of diverse training data}: Models have failure cases depending on their training data. We showcase predictions by our fusion model with optical flow on different training mixes. Datasets should have: i) many different semantic classes ii) diverse motion scenarios iii) non-rigid motions iv) group of objects moving in union v) objects that \textit{could} move but \textit{don't}. Leveraging more data will eliminate these failure cases. This shows, that we dont actually need many changes in the model architecture for motion segmentation. Using high quality motion data and a good training set resolves most issues. Finally, even ill-posed ambiguities by 2D motion representations can be compensated with image context, e.g. a car on a driving lane is more likely to drive than a parked car to the side.}
\label{fig:mix_data}
\end{figure*}

Figure \ref{fig:mix_data} shows an output progression depending on the training data mix. We observed that many common failure cases are very causal w.r.t the input training data: Models simply cannot learn non-rigid motion grouping, when not enough non-rigid motion patterns exist in the data. Simulteanously, we want driving data with common degenerate motion cases for autonomous driving. Including many diverse such cases in the training data can \textit{logically} resolve these issues. 
However, overfitting can be an issue: When not balancing the training dataset cautiously, performance degrades on some dataset (here Davis), while being very good on another (here Kitti). This observation is similar to experiments on other tasks \cite{ranftl2021vision} or multi-task training: When combining multiple datasets, the balancing/sampling strategy is yet another optimization problem. These issues are also partially visualized in Figure \ref{fig:training_data} and \ref{fig:moca}.
\newpage
We want to highlight that training on mix 2 seems to strike a very good balance when evaluating on Kitti and Davis. Note how \textit{no real} driving data is used, yet we achieve strong mAP on Kitti even when using ill-posed optical flow. Since Mix 3 includes much more driving data, we overfit on this type of data and performance on Davis degrades. Adding other datasets like YTVOS \cite{xu2018youtube} would resolve this problem. We leave the data balancing problem for future work.  

\clearpage
\subsection{Failure Cases}
\label{sup:failure}
\begin{figure*}
	\centering
	
	\begin{overpic}[width=.19\linewidth, height=2.0cm, clip, tics=0, clip, trim={0.0cm 0.0cm 0.0cm 0.0cm}]
		{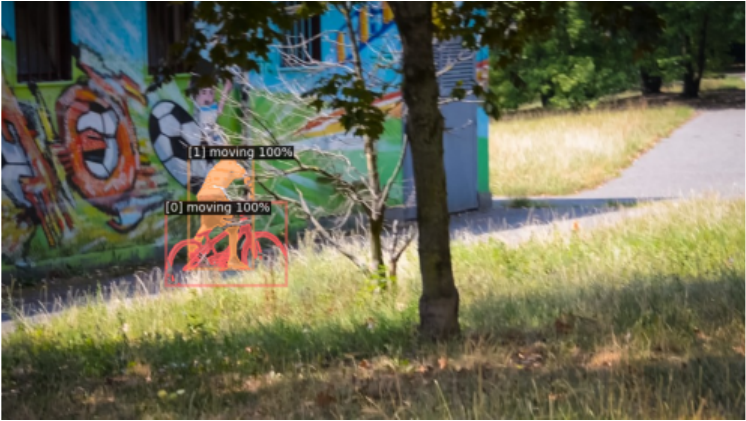}
		\put(45, 65){\footnotesize{GT}}
		\put(135, 65){\footnotesize{Prediction}}
		\put(250, 65){\footnotesize{GT}}
		\put(340, 65){\footnotesize{Prediction}}
	\end{overpic}\hspace*{-0.3em}
	\begin{overpic}[width=.19\linewidth, height=2.0cm, trim={0.0cm 0.0cm 0.0cm 0.0cm}, tics=0, clip]
		{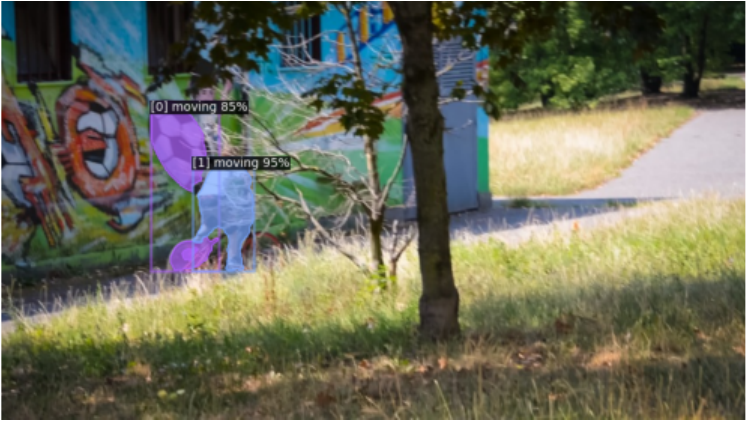}
	\end{overpic}
	\hspace{0.05cm}
	\begin{overpic}[width=.19\linewidth, height=2.0cm, trim={0.0cm 0.0cm 0.0cm 0.0cm}, tics=0, clip]
		{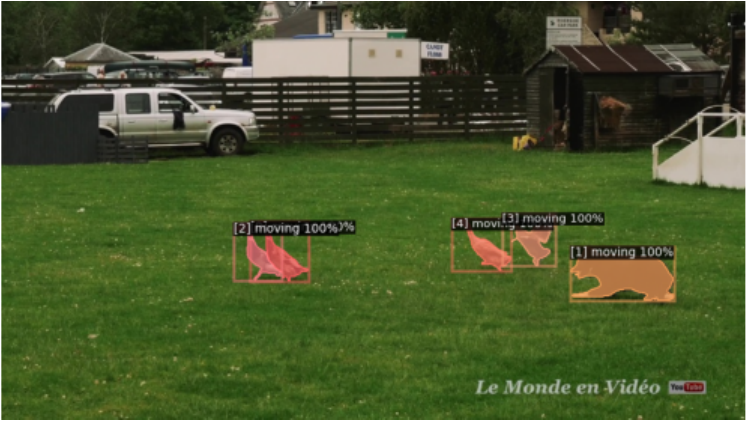}
	\end{overpic}\hspace*{-0.3em}
	\begin{overpic}[width=.19\linewidth, height=2.0cm, trim={0.0cm 0.0cm 0.0cm 0.0cm}, tics=0, clip]		
		{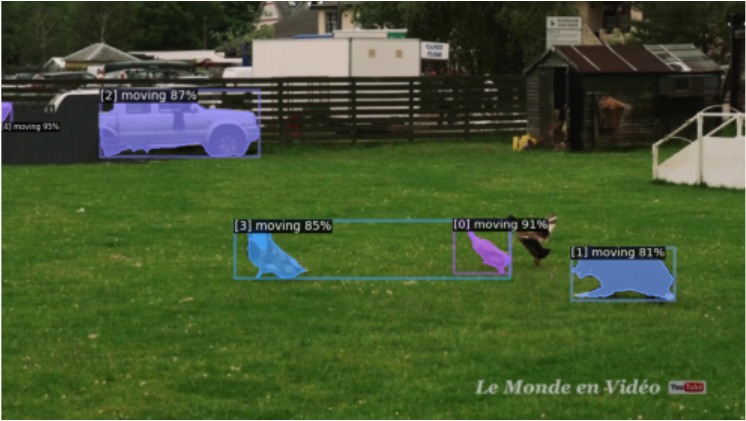}
	\end{overpic}\hspace*{-0.3em}
	\\[0.1em]
	
	\begin{overpic}[width=.19\linewidth, height=2.0cm, clip, tics=0, clip, trim={0.0cm 0.0cm 0.0cm 0.0cm}]
		{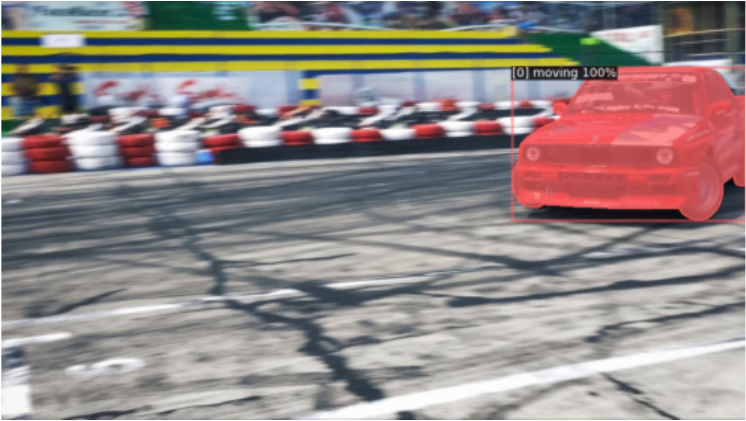}
	\end{overpic}\hspace*{-0.3em}
	\begin{overpic}[width=.19\linewidth, height=2.0cm, trim={0.0cm 0.0cm 0.0cm 0.0cm}, tics=0, clip]
		{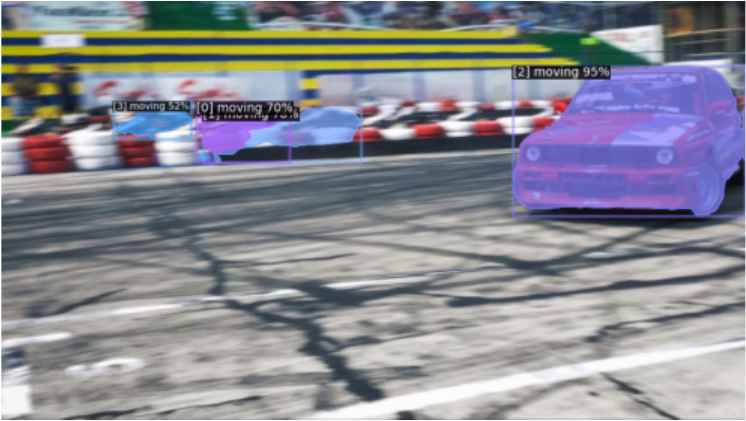}
	\end{overpic}
	\hspace{0.05cm}
	\begin{overpic}[width=.19\linewidth, height=2.0cm, trim={0.0cm 0.0cm 0.0cm 0.0cm}, tics=0, clip]
		{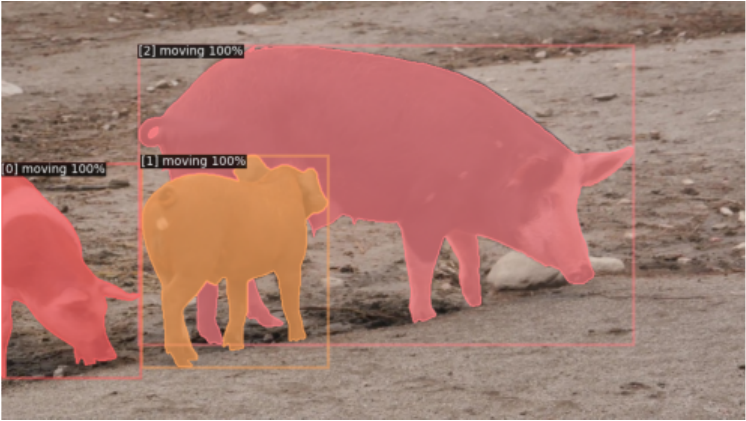}
	\end{overpic}\hspace*{-0.3em}
	\begin{overpic}[width=.19\linewidth, height=2.0cm, trim={0.0cm 0.0cm 0.0cm 0.0cm}, tics=0, clip]		
		{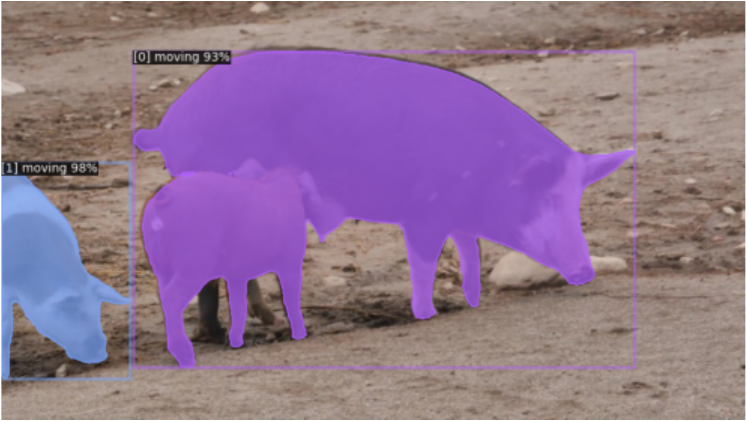}
	\end{overpic}\hspace*{-0.3em}
	\\[0.1em]	
	
	\begin{overpic}[width=.19\linewidth, height=2.0cm, clip, tics=0, clip, trim={0.0cm 0.0cm 0.0cm 0.0cm}]
		{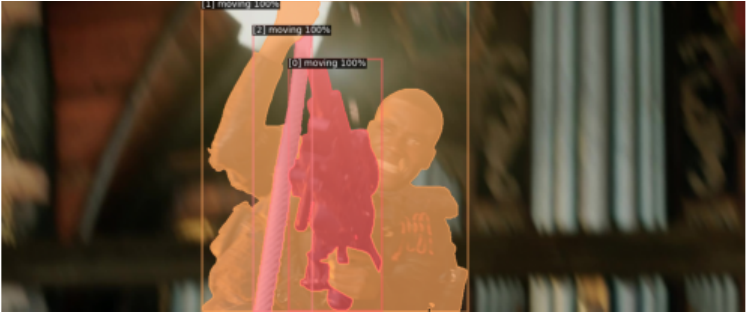}
	\end{overpic}\hspace*{-0.3em}
	\begin{overpic}[width=.19\linewidth, height=2.0cm, trim={0.0cm 0.0cm 0.0cm 0.0cm}, tics=0, clip]
		{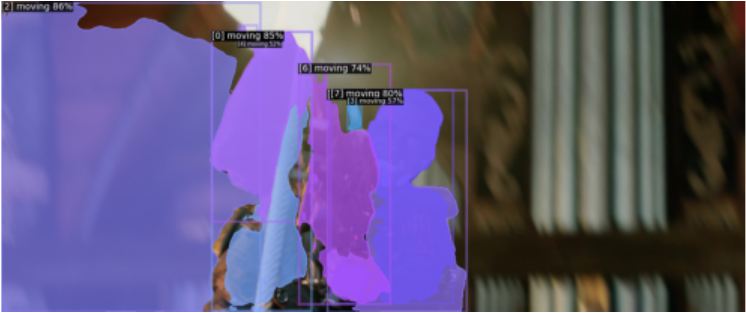}
	\end{overpic}
	\hspace{0.05cm}
	\begin{overpic}[width=.19\linewidth, height=2.0cm, trim={0.0cm 0.0cm 0.0cm 0.0cm}, tics=0, clip]
		{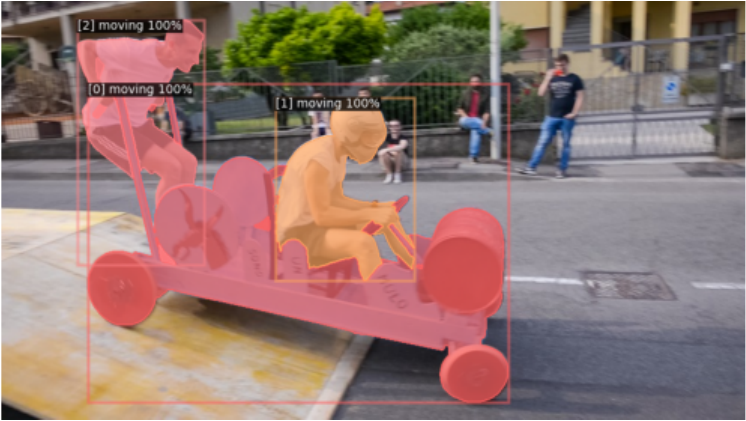}
	\end{overpic}\hspace*{-0.3em}
	\begin{overpic}[width=.19\linewidth, height=2.0cm, trim={0.0cm 0.0cm 0.0cm 0.0cm}, tics=0, clip]		
		{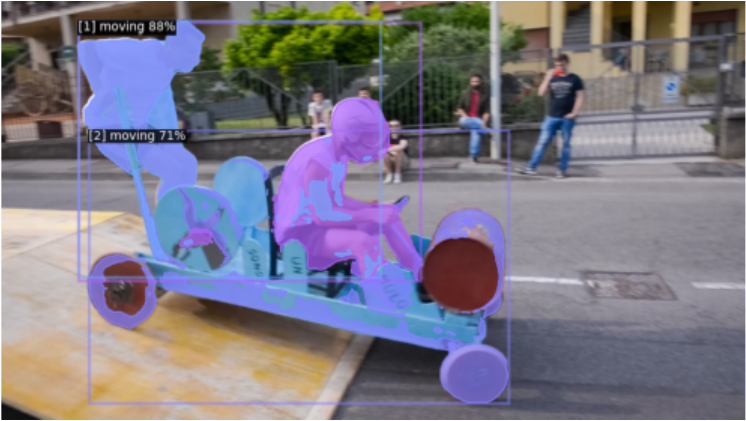}
	\end{overpic}\hspace*{-0.3em}
	\\[0.1em]
	
	\begin{overpic}[width=.19\linewidth, height=2.0cm, clip, tics=0, clip, trim={0.0cm 0.0cm 0.0cm 0.0cm}]
		{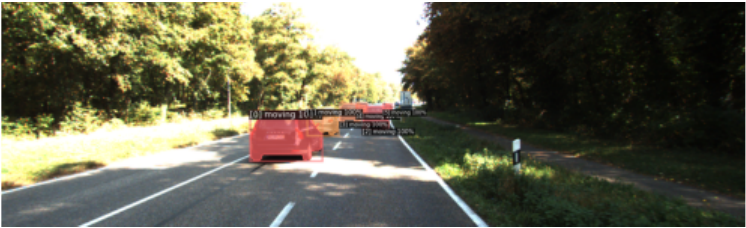}
	\end{overpic}\hspace*{-0.3em}
	\begin{overpic}[width=.19\linewidth, height=2.0cm, trim={0.0cm 0.0cm 0.0cm 0.0cm}, tics=0, clip]
		{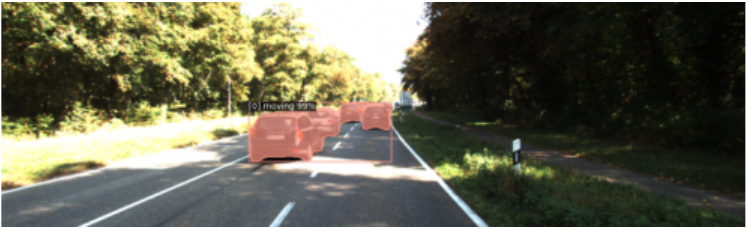}
	\end{overpic}
	\hspace{0.05cm}
	\begin{overpic}[width=.19\linewidth, height=2.0cm, trim={0.0cm 0.0cm 0.0cm 0.0cm}, tics=0, clip]
		{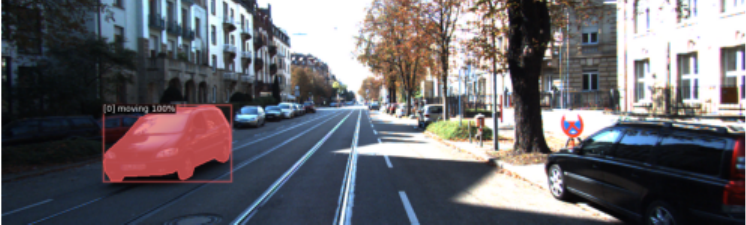}
	\end{overpic}\hspace*{-0.3em}
	\begin{overpic}[width=.19\linewidth, height=2.0cm, trim={0.0cm 0.0cm 0.0cm 0.0cm}, tics=0, clip]		
		{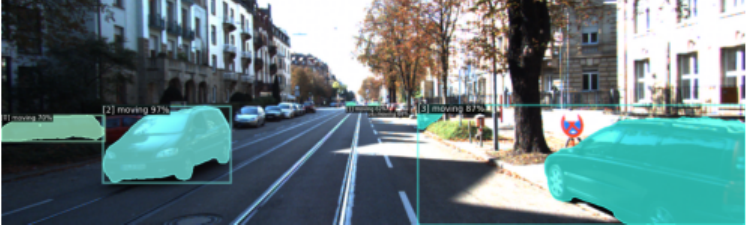}
	\end{overpic}\hspace*{-0.3em}
	\\[0.1em]

	\begin{overpic}[width=.19\linewidth, height=2.0cm, clip, tics=0, clip, trim={0.0cm 0.0cm 0.0cm 0.0cm}]
		{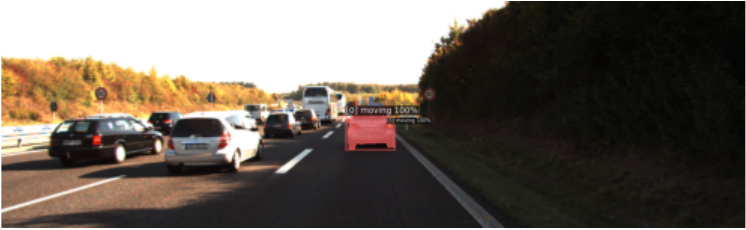}
	\end{overpic}\hspace*{-0.3em}
	\begin{overpic}[width=.19\linewidth, height=2.0cm, trim={0.0cm 0.0cm 0.0cm 0.0cm}, tics=0, clip]
		{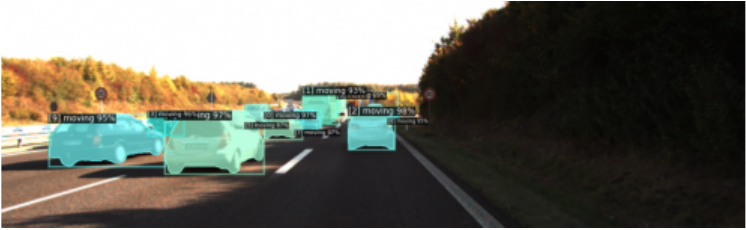}
	\end{overpic}
	\hspace{0.05cm}
	\begin{overpic}[width=.19\linewidth, height=2.0cm, trim={0.0cm 0.0cm 0.0cm 0.0cm}, tics=0, clip]
		{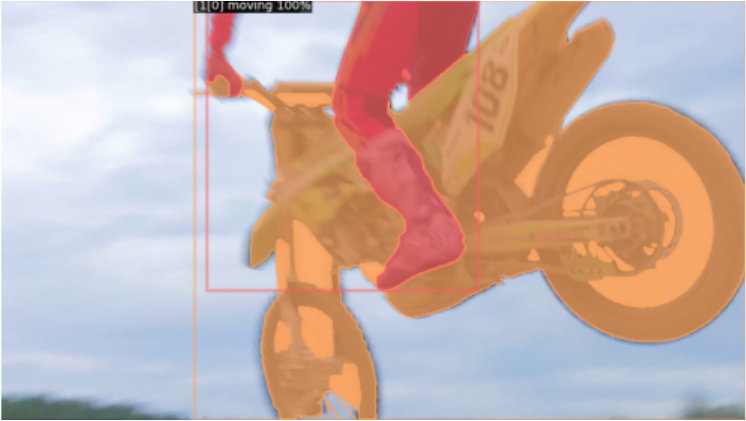}
	\end{overpic}\hspace*{-0.3em}
	\begin{overpic}[width=.19\linewidth, height=2.0cm, trim={0.0cm 0.0cm 0.0cm 0.0cm}, tics=0, clip]		
		{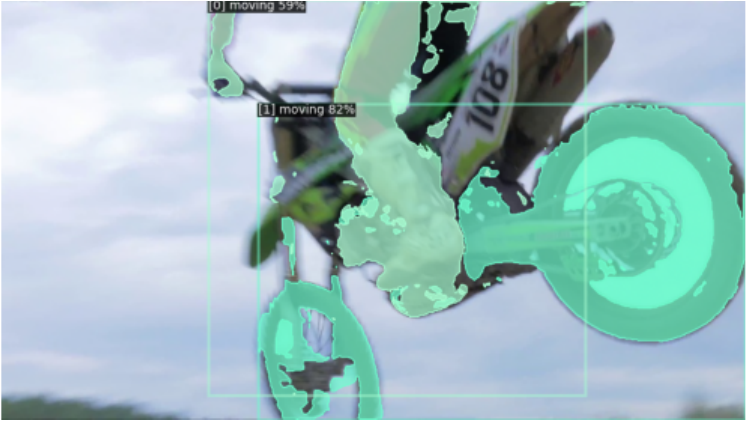}
	\end{overpic}\hspace*{-0.3em}
	\\[0.1em]
\caption{\textit{Failure cases}: i) Moving objects are missed even when the motion map defines them clearly. ii) Multiple objects are grouped together iii) Objects bleed into the background iv) False Positives due to noise in motion map v) Misalignment appearance and motion stream. Semantic classes that can move often overrule the motion stream. vi) Undersegmentation vii) Small objects}
\label{fig:training_data}
\end{figure*}

We show multiple failure cases explicitly in Figure \ref{fig:training_data}. Other failures can be partially observed in Figure \ref{fig:mix_data} and \ref{fig:moca}. 
We further encourage readers to view the additional videos, which contain much more information on multiple datasets and compare our trained models on multiple modalities with \cite{neoral2021monocular}.

\clearpage
\subsection{Why A Twin 2-stream Architecture Can be A Good Idea}
\label{sup:moca}
\begin{figure*}
	\centering
	\begin{overpic}[width=.19\linewidth, height=2.0cm, clip, tics=0, clip, trim={0.0cm 0.1cm 0.0cm 0.1cm}]
		{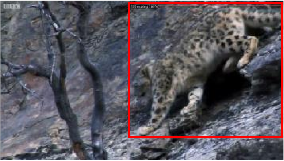}
		\put(40, 65){\footnotesize{GT}}
		\put(145, 65){\footnotesize{RGB}}
		\put(235, 65){\footnotesize{Optical Flow}}
		\put(325, 65){\footnotesize{RGB + Optical Flow}}
		\put(442, 65){\footnotesize{Raptor \cite{neoral2021monocular}}}
	\end{overpic}
	\hspace{0.05cm}
	\begin{overpic}[width=.19\linewidth, height=2.0cm, trim={0.0cm 2.0cm 0.0cm 0.1cm}, tics=0, clip]
		{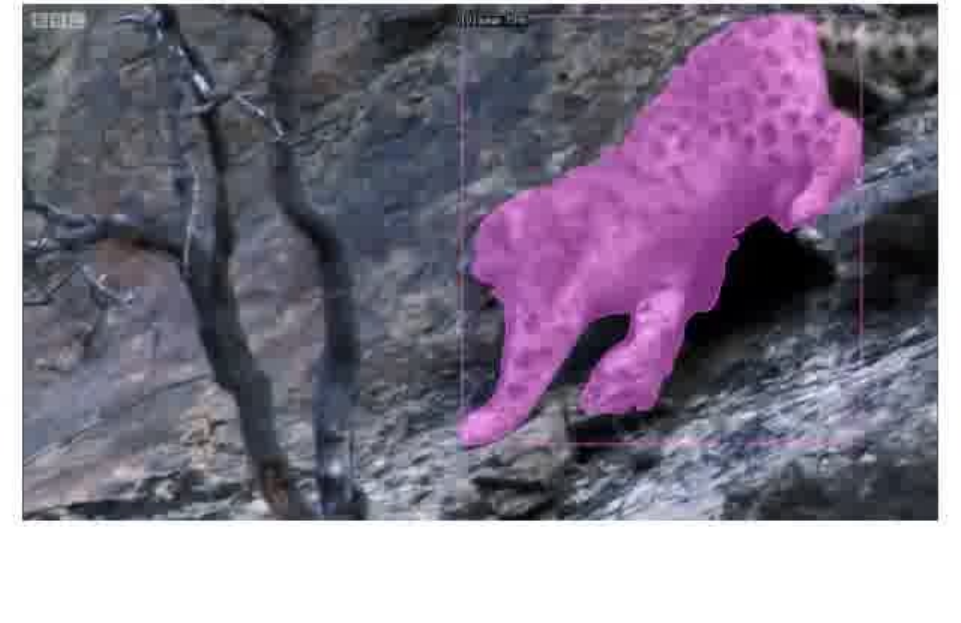}
	\end{overpic}\hspace*{-0.3em}
	\begin{overpic}[width=.19\linewidth, height=2.0cm, trim={0.0cm 2.0cm 0.0cm 0.1cm}, tics=0, clip]
		{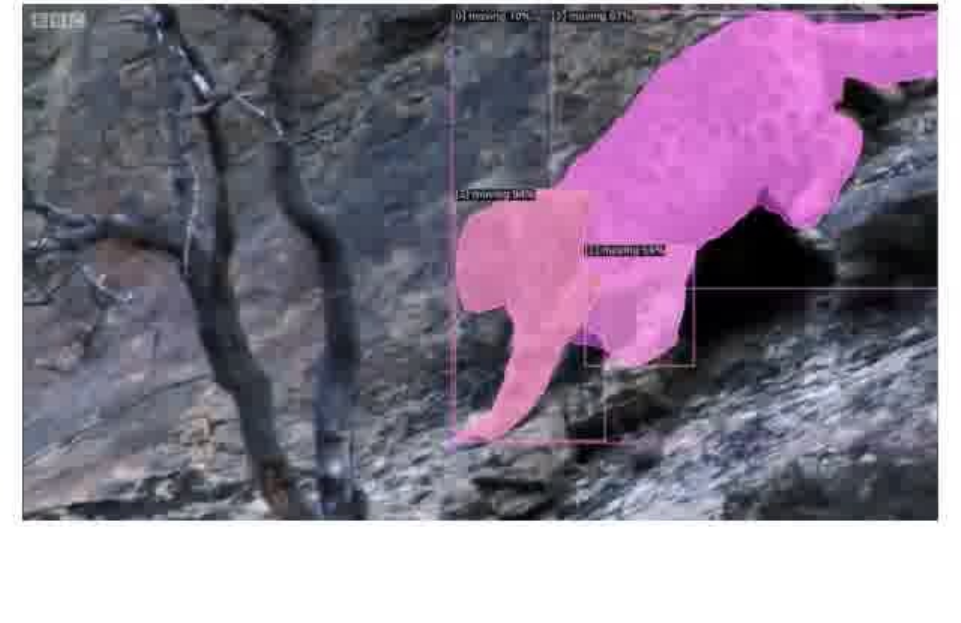}
	\end{overpic}\hspace*{-0.3em}
	\begin{overpic}[width=.19\linewidth, height=2.0cm, trim={0.0cm 2.0cm 0.0cm 0.1cm}, tics=0, clip]
		{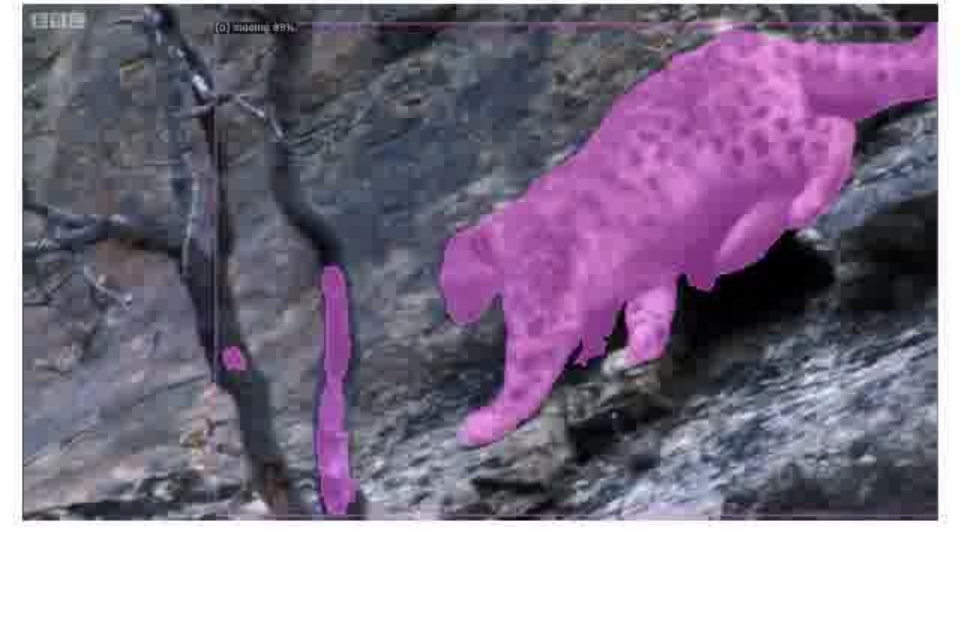}
	\end{overpic}\hspace*{-0.3em}
	\begin{overpic}[width=.19\linewidth, height=2.0cm, trim={0.0cm 2.0cm 0.0cm 0.1cm}, tics=0, clip]
		{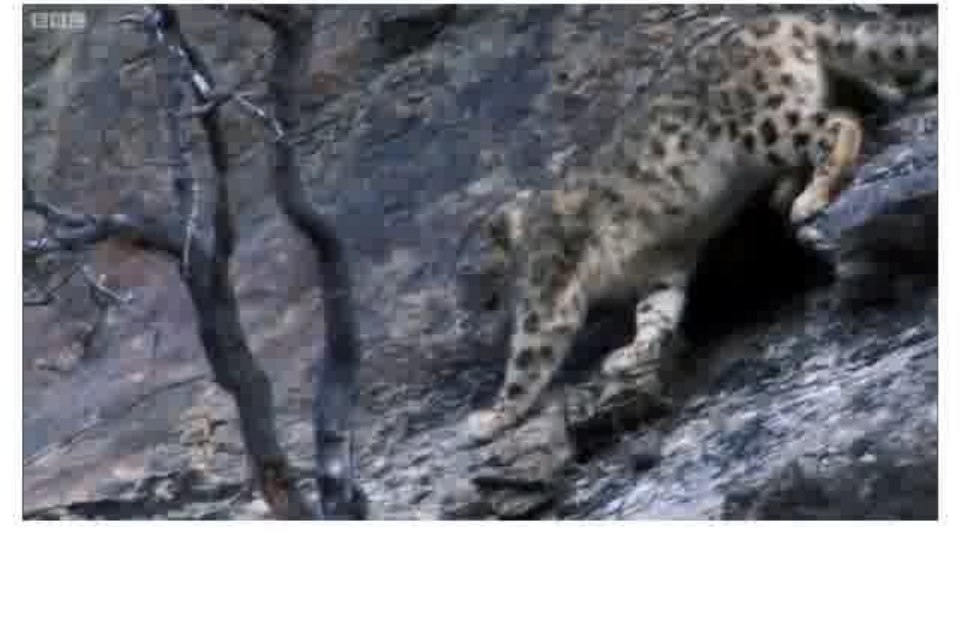}
	\end{overpic}
	\\[0.1em]
	
	\begin{overpic}[width=.19\linewidth, height=2.0cm, tics=0, clip, trim={0.0cm 0.1cm 0.0cm 0.1cm}]
		{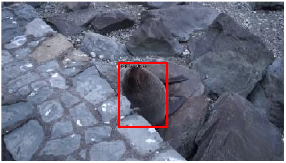}
	\end{overpic}
	\hspace{0.05cm}
	\begin{overpic}[width=.19\linewidth, height=2.0cm, trim={0.0cm 1.7cm 0.0cm 0.2cm}, tics=0, clip]
		{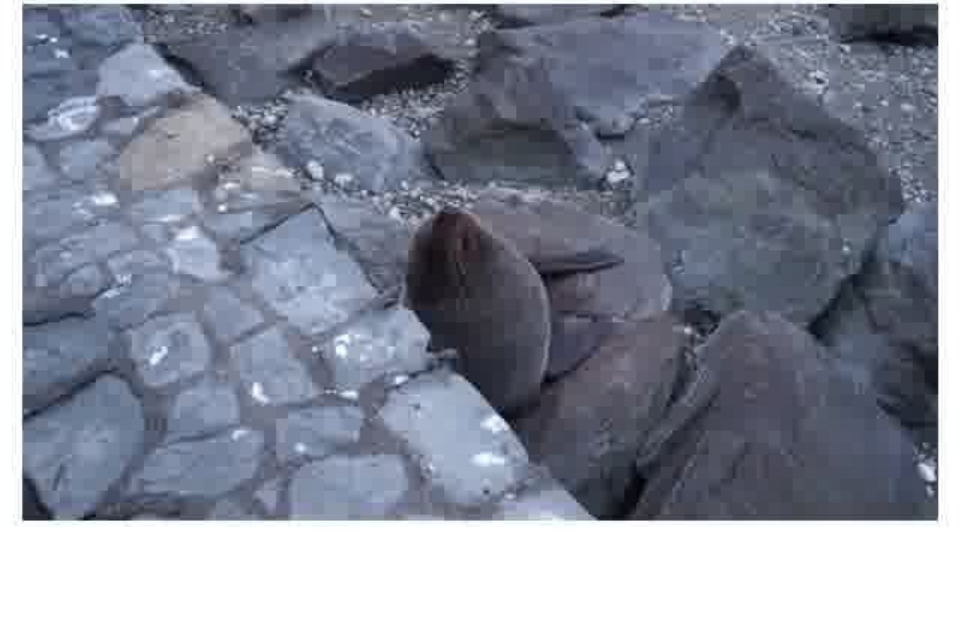}
	\end{overpic}\hspace*{-0.3em}
	\begin{overpic}[width=.19\linewidth, height=2.0cm, trim={0.0cm 1.7cm 0.0cm 0.2cm}, tics=0, clip]
		{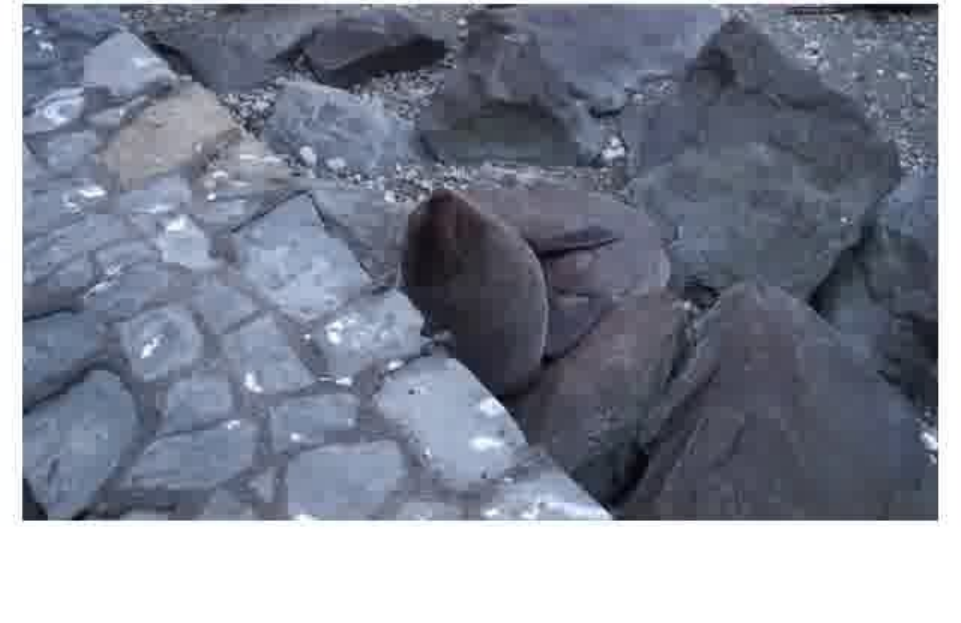}
	\end{overpic}\hspace*{-0.3em}
	\begin{overpic}[width=.19\linewidth, height=2.0cm, trim={0.0cm 1.7cm 0.0cm 0.2cm}, tics=0, clip]
		{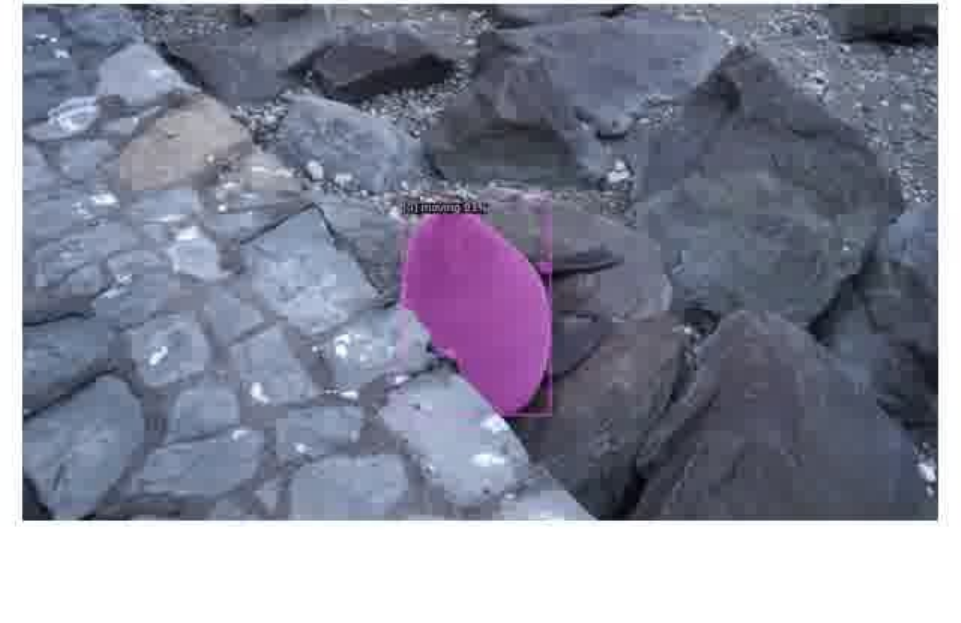}
	\end{overpic}\hspace*{-0.3em}
	\begin{overpic}[width=.19\linewidth, height=2.0cm, trim={0.0cm 1.7cm 0.0cm 0.2cm}, tics=0, clip]
		{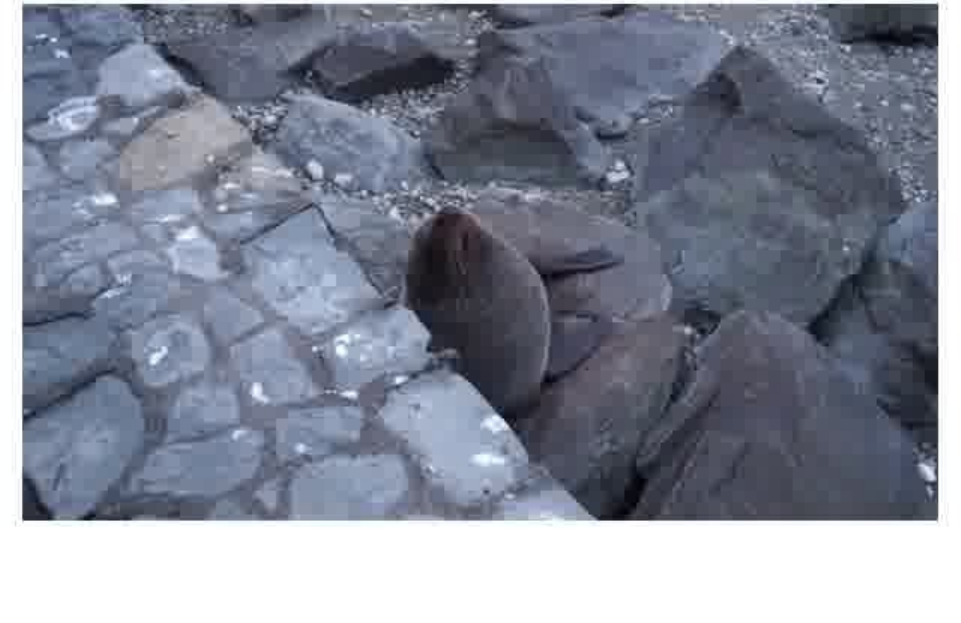}
	\end{overpic}
	\\[0.1em]
	
	\begin{overpic}[width=.19\linewidth, height=2.0cm, tics=0, clip, trim={0.0cm 0.1cm 0.0cm 0.1cm}]
		{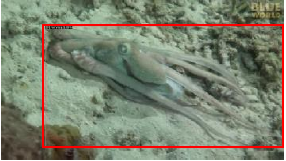}
	\end{overpic}
	\hspace{0.05cm}
	\begin{overpic}[width=.19\linewidth, height=2.0cm, trim={0.0cm 1.7cm 0.0cm 0.2cm}, tics=0, clip]
		{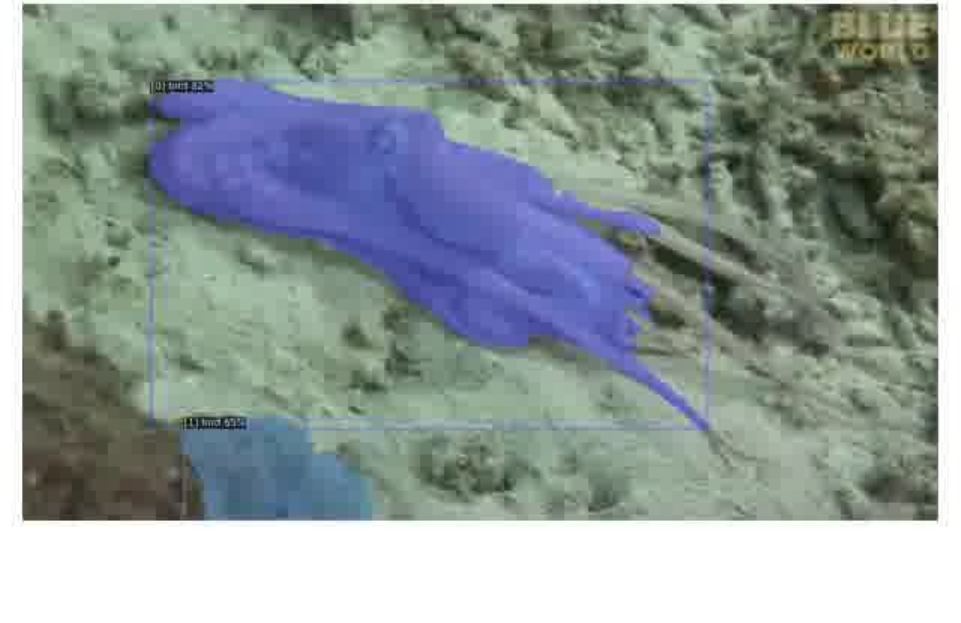}
	\end{overpic}\hspace*{-0.3em}
	\begin{overpic}[width=.19\linewidth, height=2.0cm, trim={0.0cm 1.7cm 0.0cm 0.2cm}, tics=0, clip]
		{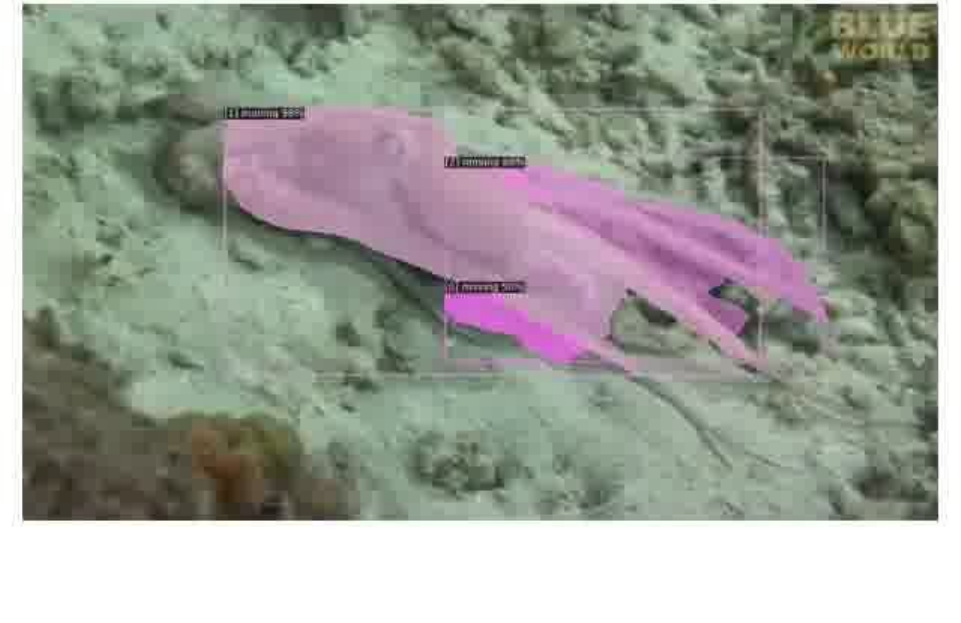}
	\end{overpic}\hspace*{-0.3em}
	\begin{overpic}[width=.19\linewidth, height=2.0cm, trim={0.0cm 1.7cm 0.0cm 0.2cm}, tics=0, clip]
		{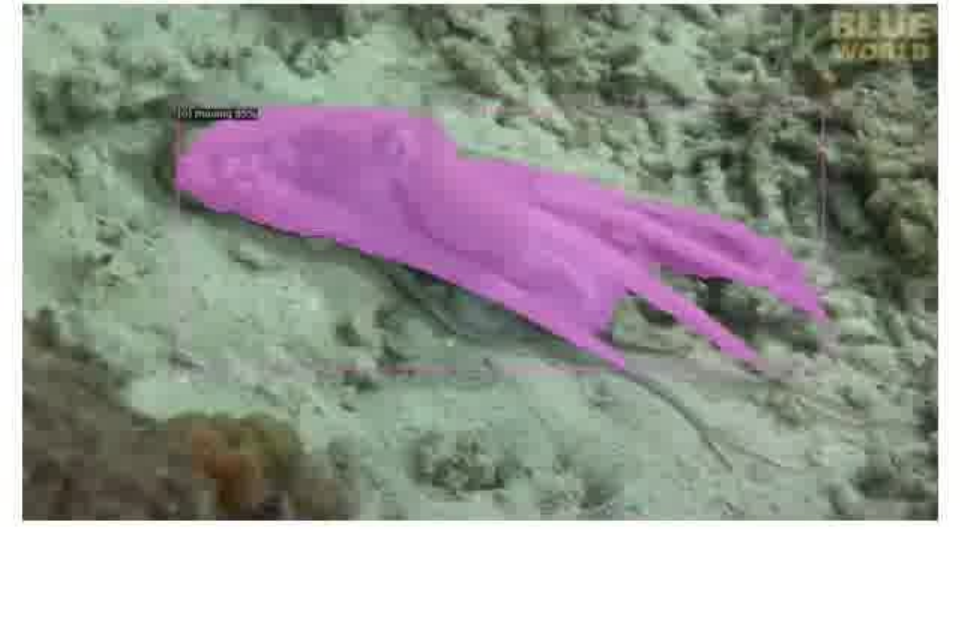}
	\end{overpic}\hspace*{-0.3em}
	\begin{overpic}[width=.19\linewidth, height=2.0cm, trim={0.0cm 1.7cm 0.0cm 0.2cm}, tics=0, clip]
		{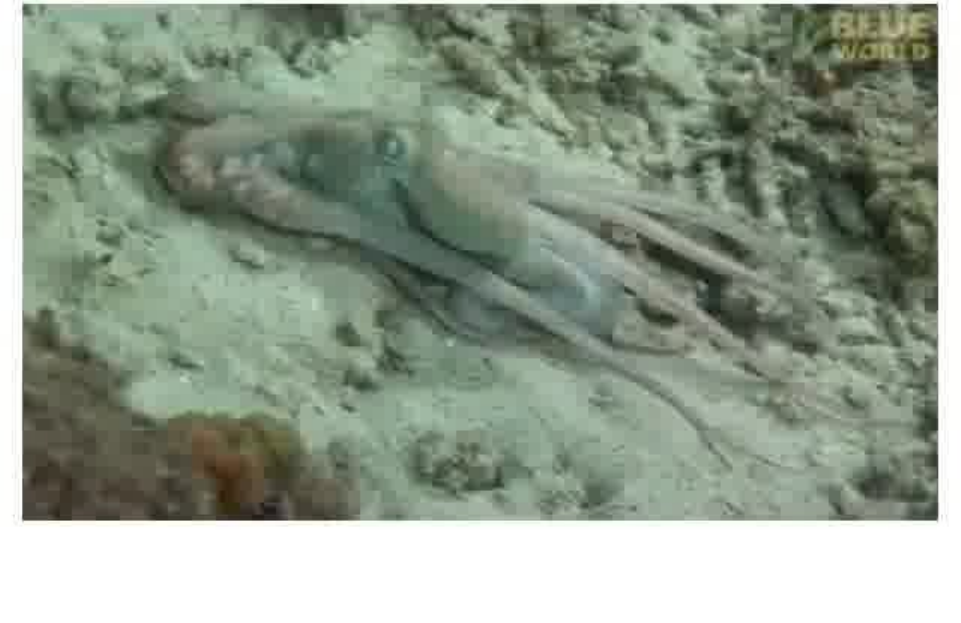}
	\end{overpic}
	\\[0.1em]
	
	\begin{overpic}[width=.19\linewidth, height=2.0cm, tics=0, clip, trim={0.0cm 0.1cm 0.0cm 0.1cm}]
		{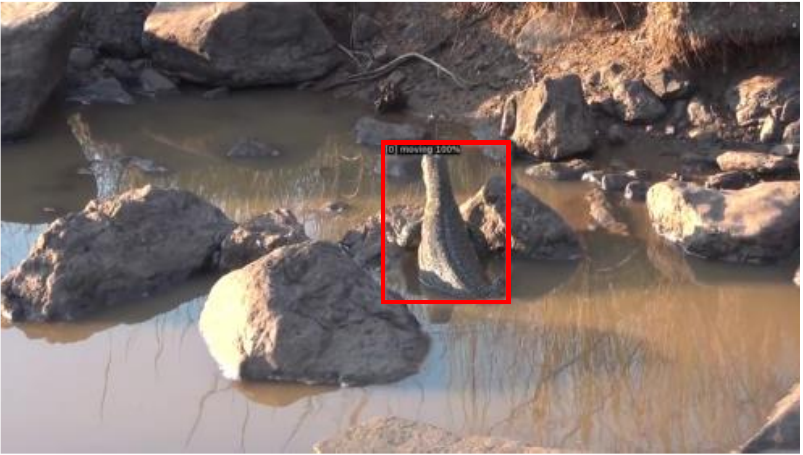}
	\end{overpic}
	\hspace{0.05cm}
	\begin{overpic}[width=.19\linewidth, height=2.0cm, trim={0.0cm 1.7cm 0.0cm 0.1cm}, tics=0, clip]
		{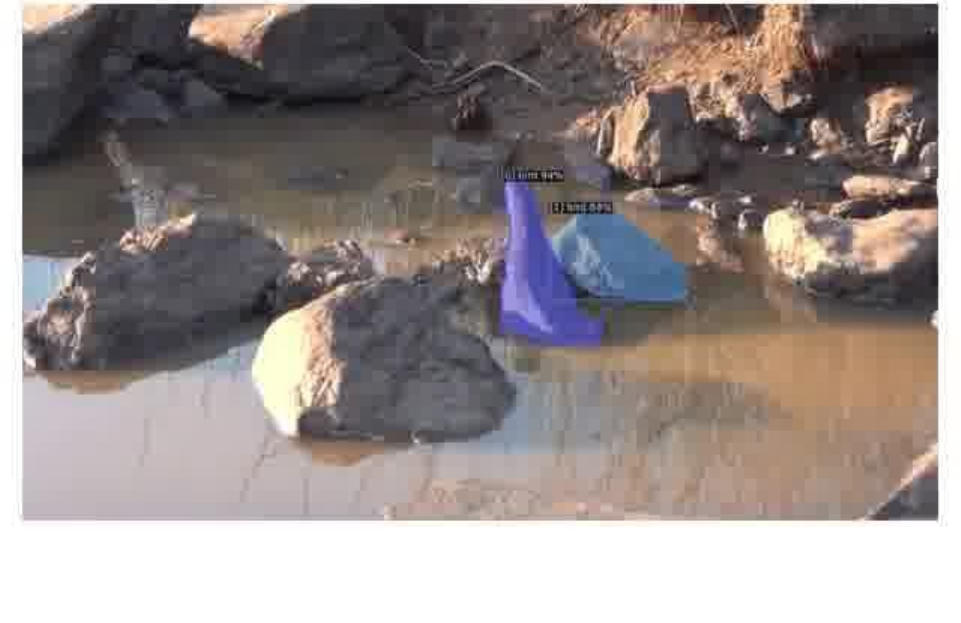}
	\end{overpic}\hspace*{-0.3em}
	\begin{overpic}[width=.19\linewidth, height=2.0cm, trim={0.0cm 1.7cm 0.0cm 1.75cm}, tics=0, clip]
		{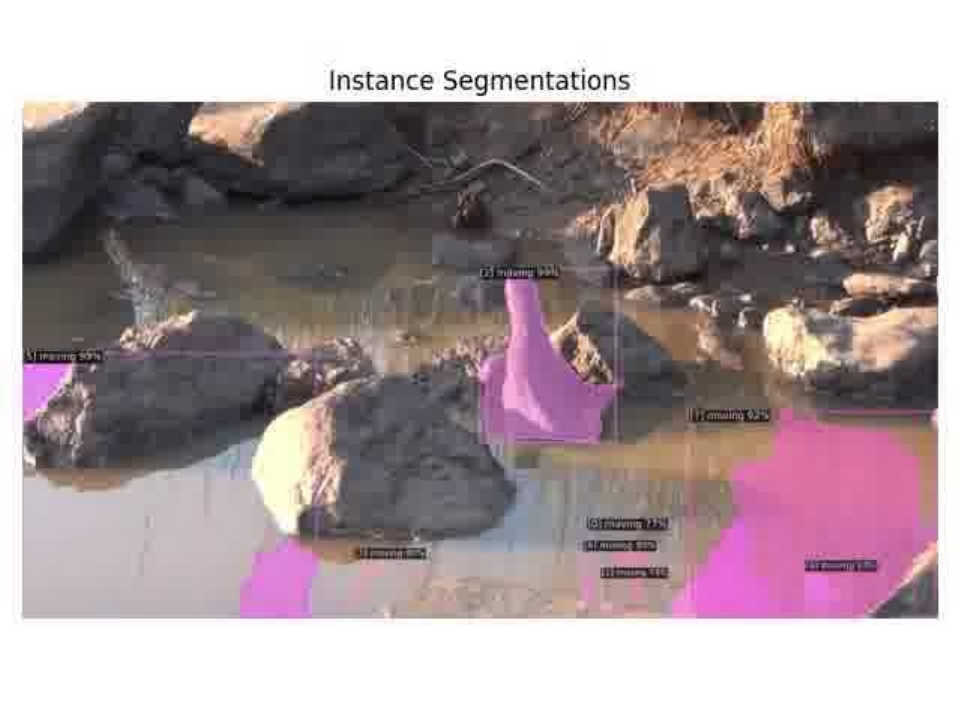}
	\end{overpic}\hspace*{-0.3em}
	\begin{overpic}[width=.19\linewidth, height=2.0cm, trim={0.0cm 1.7cm 0.0cm 0.1cm}, tics=0, clip]
		{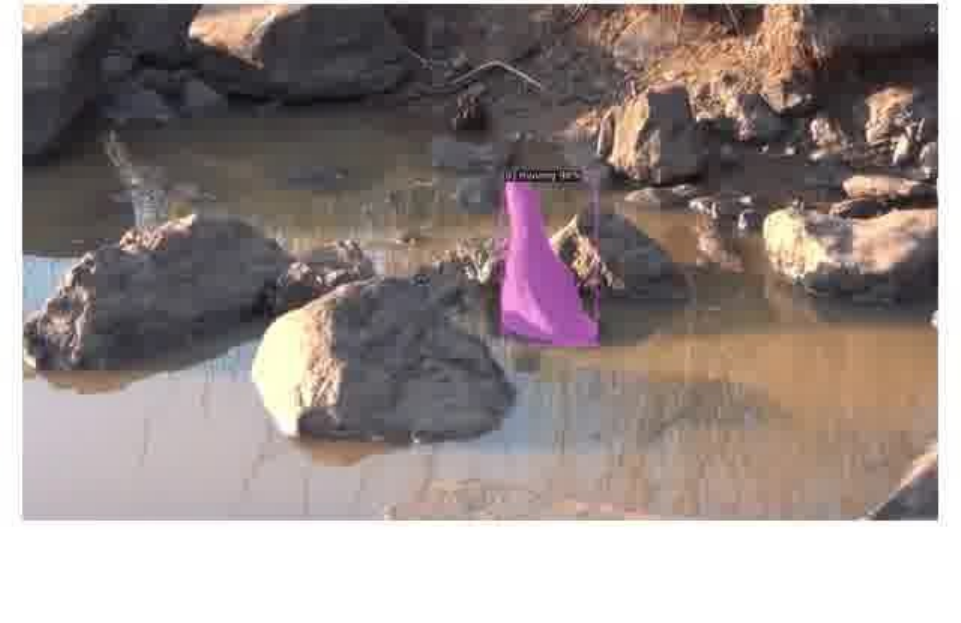}
	\end{overpic}\hspace*{-0.3em}
	\begin{overpic}[width=.19\linewidth, height=2.0cm, trim={0.0cm 1.7cm 0.0cm 0.1cm}, tics=0, clip]
		{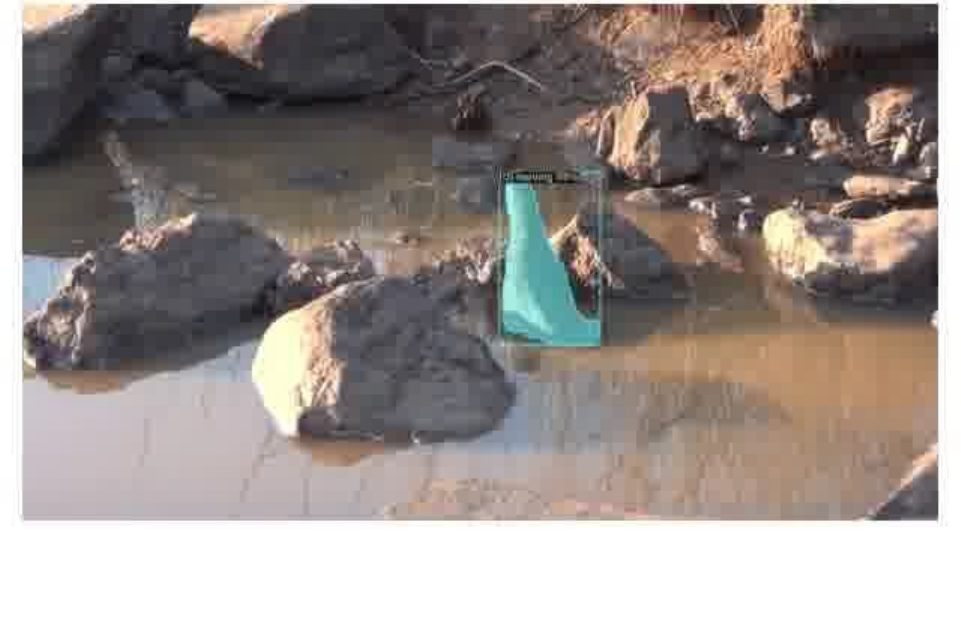}
	\end{overpic}
	\\[0.1em]
	
	\begin{overpic}[width=.19\linewidth, height=2.0cm, tics=0, clip, trim={0.0cm 0.1cm 0.0cm 0.1cm}]
		{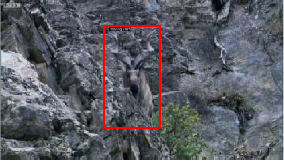}
	\end{overpic}
	\hspace{0.05cm}
	\begin{overpic}[width=.19\linewidth, height=2.0cm, trim={0.0cm 1.7cm 0.0cm 0.1cm}, tics=0, clip]
		{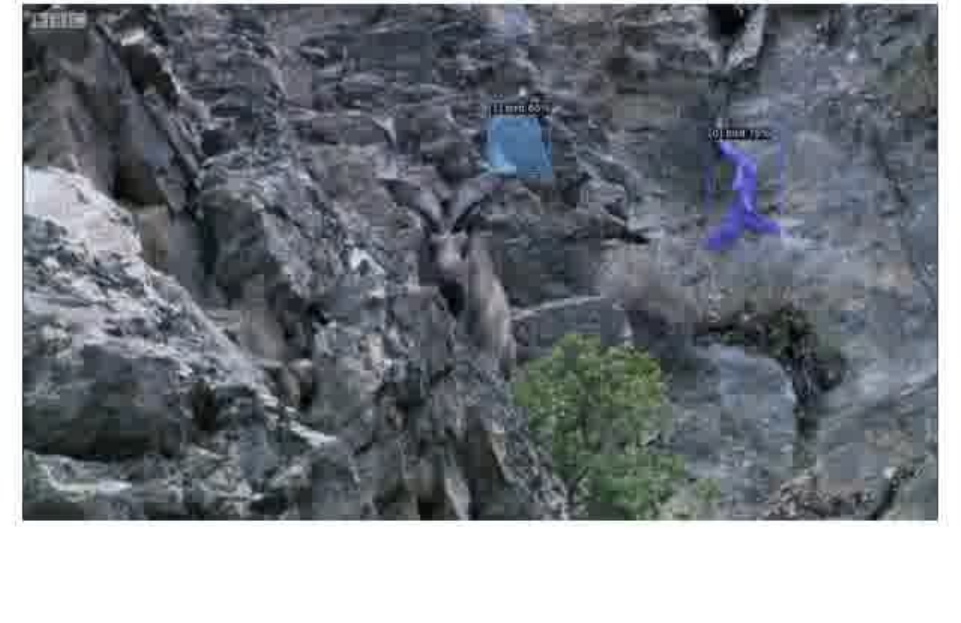}
	\end{overpic}\hspace*{-0.3em}
	\begin{overpic}[width=.19\linewidth, height=2.0cm, trim={0.0cm 1.7cm 0.0cm 0.1cm}, tics=0, clip]
		{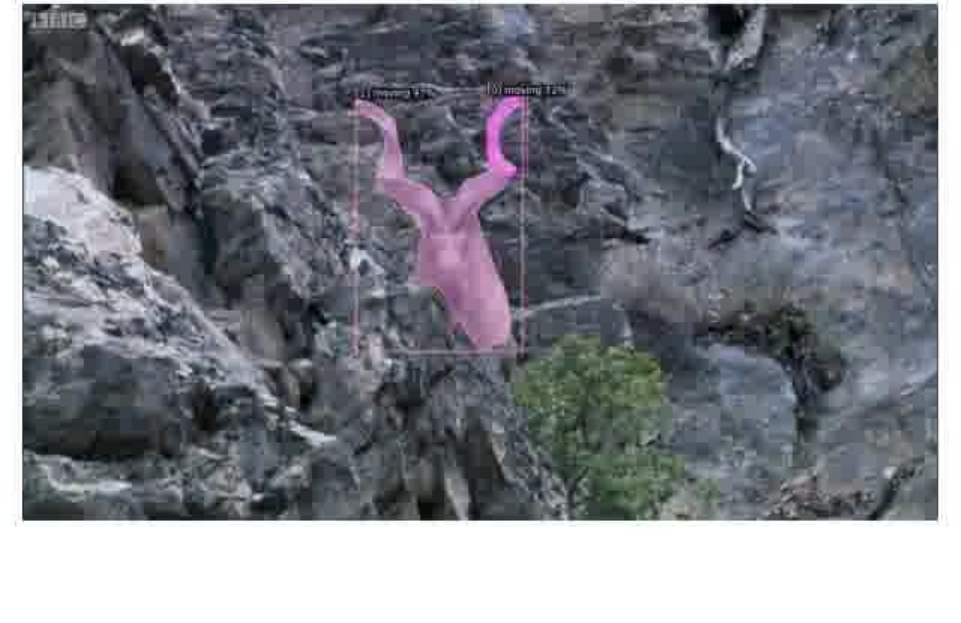}
	\end{overpic}\hspace*{-0.3em}
	\begin{overpic}[width=.19\linewidth, height=2.0cm, trim={0.0cm 1.7cm 0.0cm 0.1cm}, tics=0, clip]
		{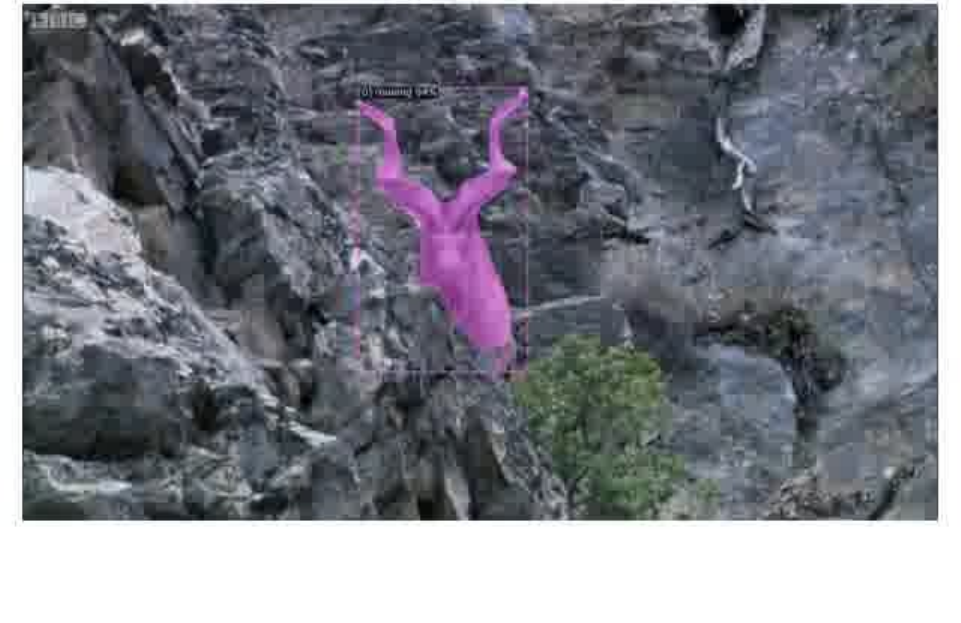}
	\end{overpic}\hspace*{-0.3em}
	\begin{overpic}[width=.19\linewidth, height=2.0cm, trim={0.0cm 1.7cm 0.0cm 0.1cm}, tics=0, clip]
		{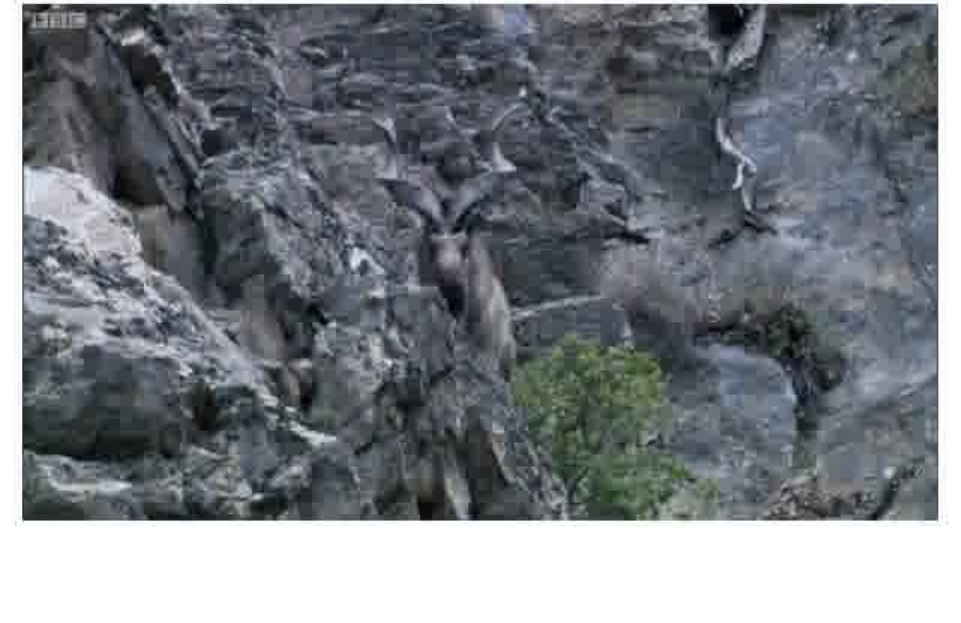}		
	\end{overpic}
	\\[0.1em]
	
	\begin{overpic}[width=.19\linewidth, height=2.0cm, tics=0, clip, trim={0.0cm 0.1cm 0.0cm 0.1cm}]
		{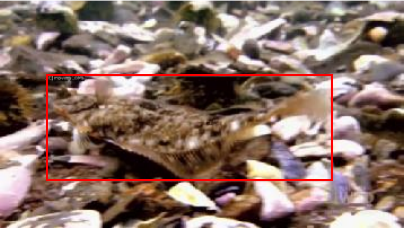}
	\end{overpic}
	\hspace{0.05cm}
	\begin{overpic}[width=.19\linewidth, height=2.0cm, trim={0.0cm 2.0cm 0.0cm 0.1cm}, tics=0, clip]
		{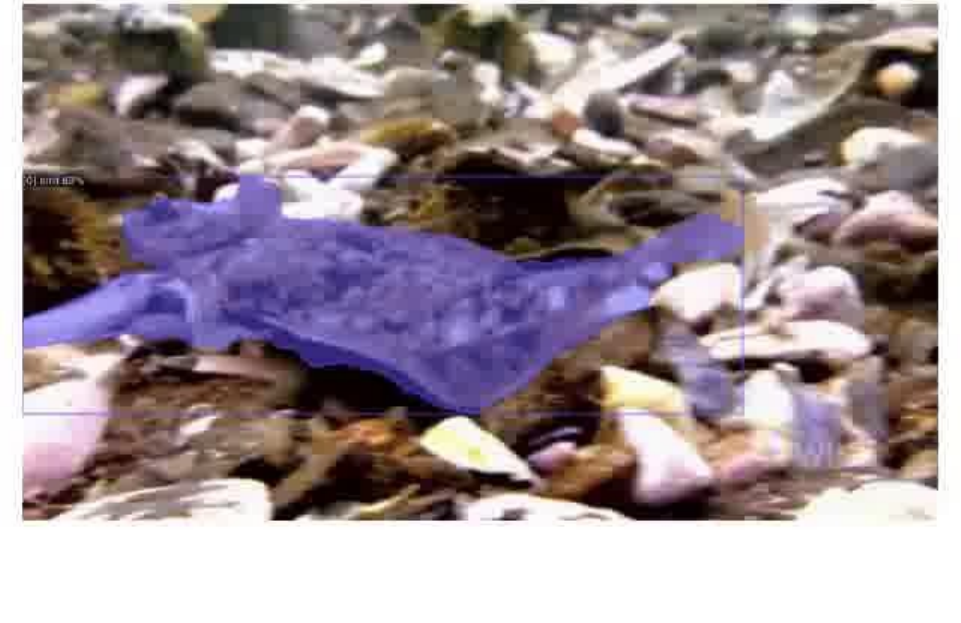}
	\end{overpic}\hspace*{-0.3em}
	\begin{overpic}[width=.19\linewidth, height=2.0cm, trim={0.0cm 2.0cm 0.0cm 0.1cm}, tics=0, clip]
		{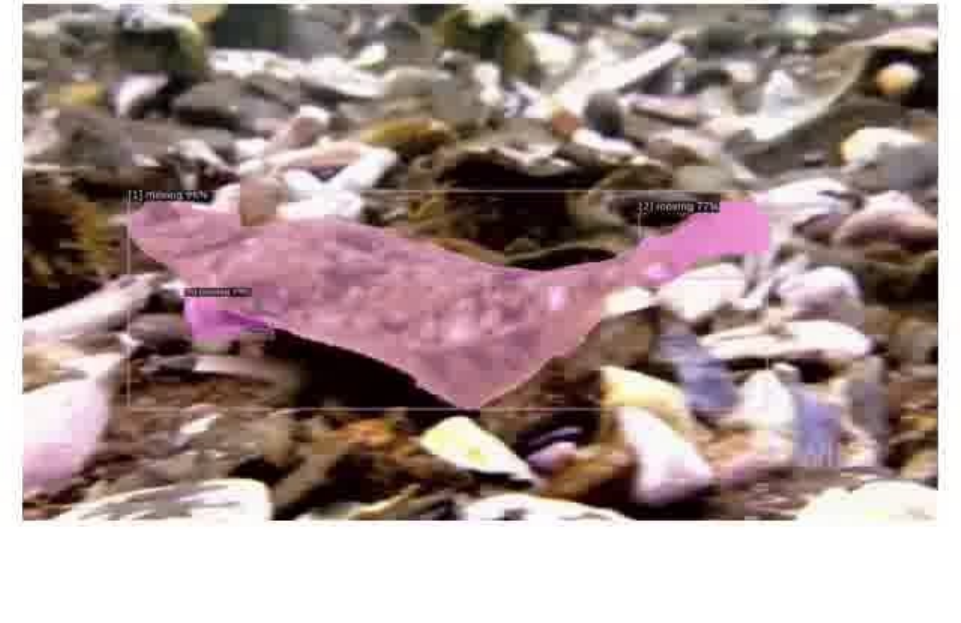}
	\end{overpic}\hspace*{-0.3em}
	\begin{overpic}[width=.19\linewidth, height=2.0cm, trim={0.0cm 2.0cm 0.0cm 0.1cm}, tics=0, clip]
		{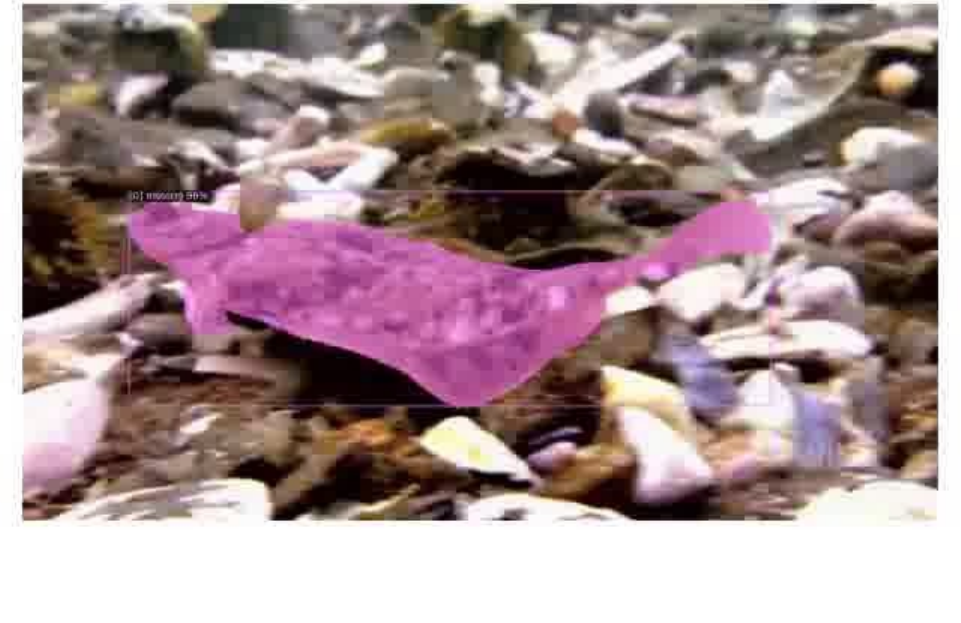}
	\end{overpic}\hspace*{-0.3em}
	\begin{overpic}[width=.19\linewidth, height=2.0cm, trim={0.0cm 2.0cm 0.0cm 0.1cm}, tics=0, clip]
		{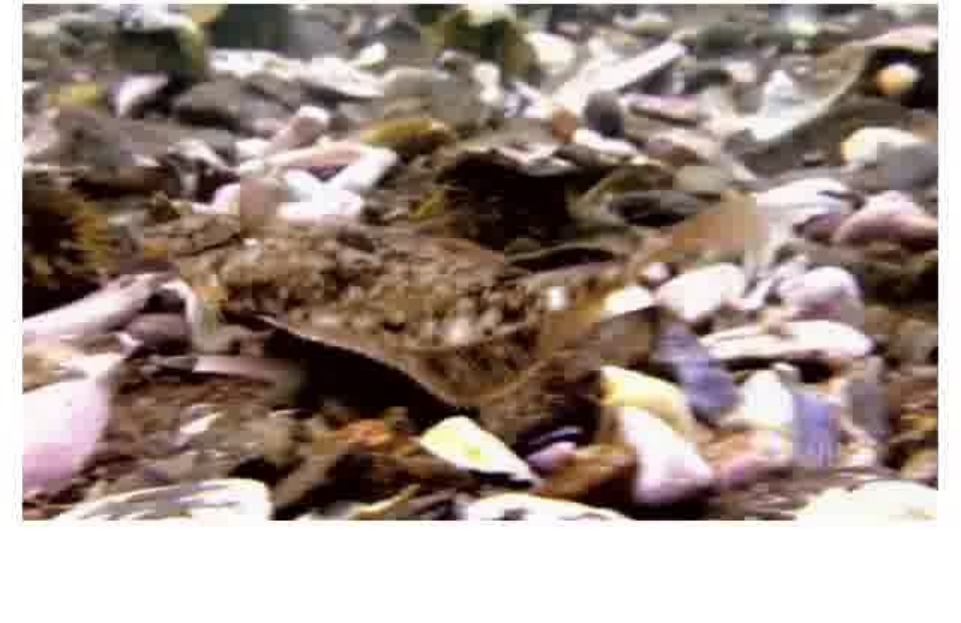}
	\end{overpic}
	\\[0.1em]
		
	\begin{overpic}[width=.19\linewidth, height=2.0cm, tics=0, clip, trim={0.0cm 0.1cm 0.0cm 0.1cm}]
		{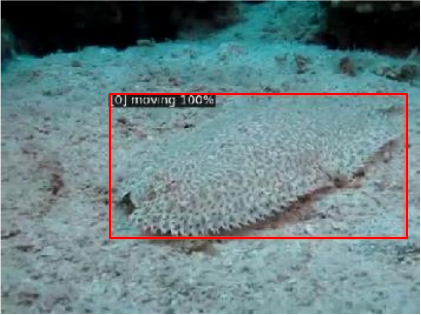}
	\end{overpic}
	\hspace{0.05cm}
	\begin{overpic}[width=.19\linewidth, height=2.0cm, trim={0.5cm 0.7cm 0.5cm 0.1cm}, tics=0, clip]
		{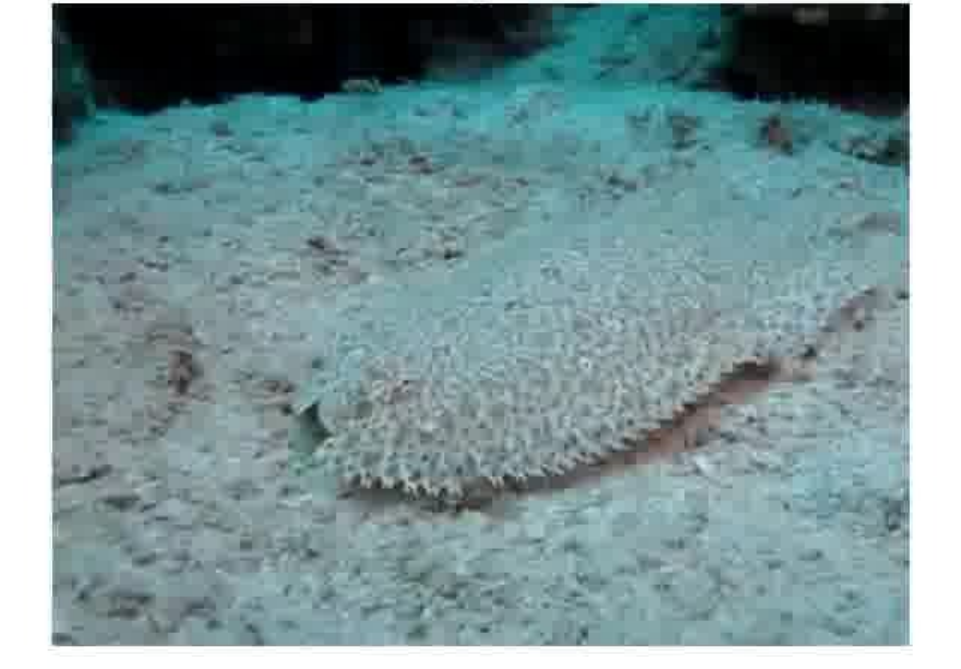}
	\end{overpic}\hspace*{-0.3em}
	\begin{overpic}[width=.19\linewidth, height=2.0cm, trim={0.5cm 0.7cm 0.5cm 0.1cm}, tics=0, clip]
		{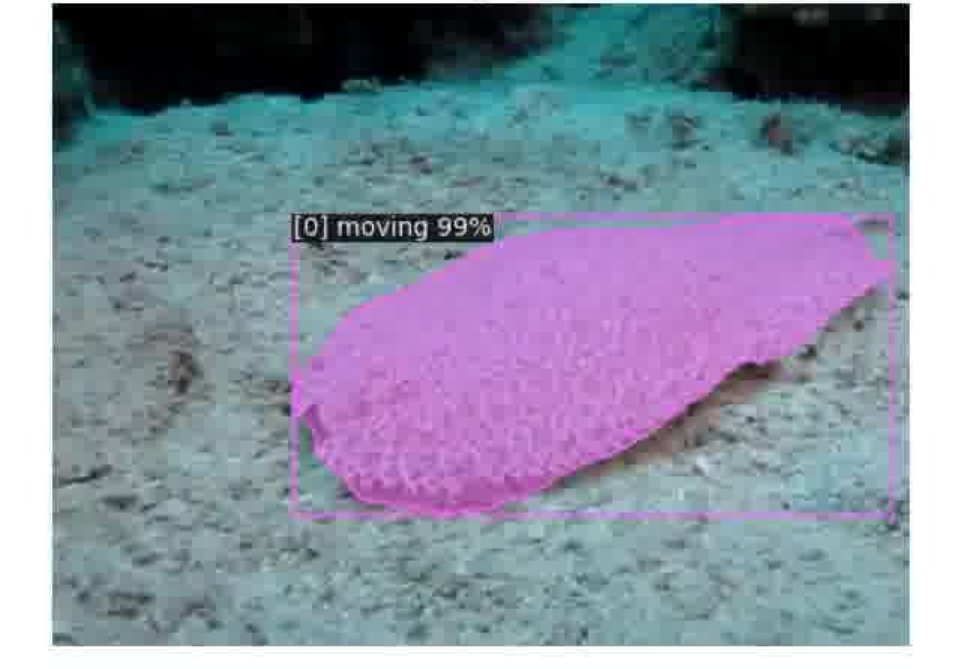}
	\end{overpic}\hspace*{-0.3em}
	\begin{overpic}[width=.19\linewidth, height=2.0cm, trim={0.5cm 0.7cm 0.5cm 0.1cm}, tics=0, clip]
		{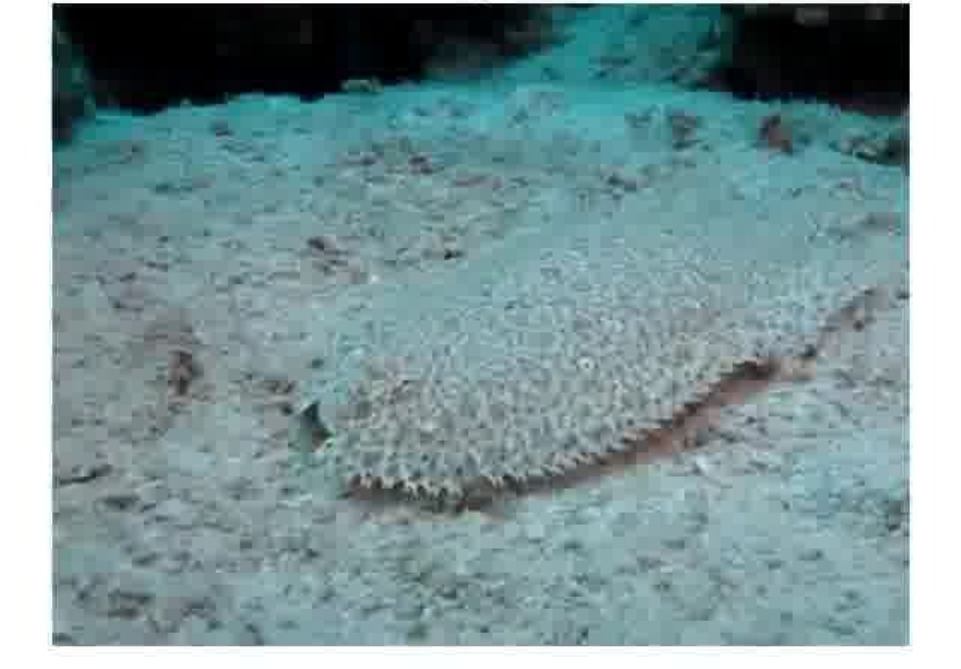}
	\end{overpic}\hspace*{-0.3em}
	\begin{overpic}[width=.19\linewidth, height=2.0cm, trim={0.5cm 0.7cm 0.5cm 0.1cm}, tics=0, clip]
		{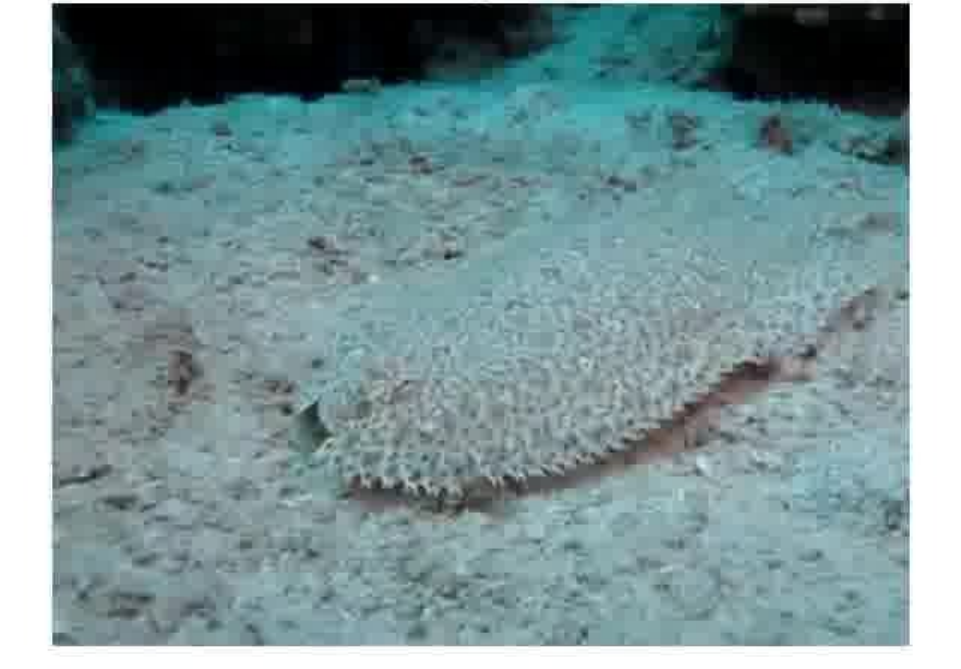}
	\end{overpic}
	\\[0.1em]
	
	\begin{overpic}[width=.19\linewidth, height=2cm, clip, tics=0, trim={0.0cm 0.1cm 0.0cm 0.1cm}]
		{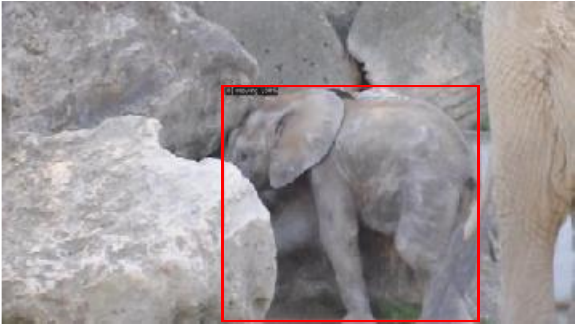}
	\end{overpic}
	\hspace{0.05cm}
	\begin{overpic}[width=.19\linewidth, height=2cm, trim={0.0cm 2.0cm 0.0cm 0.1cm}, tics=0, clip]
		{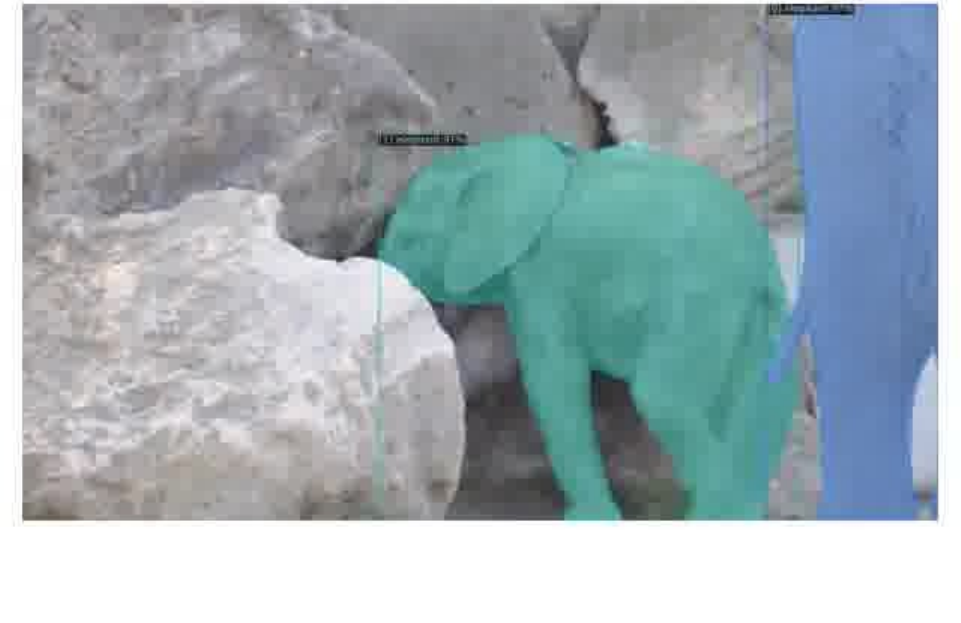}
	\end{overpic}\hspace*{-0.3em}
	\begin{overpic}[width=.19\linewidth, height=2cm, trim={0.0cm 2.0cm 0.0cm 0.1cm}, tics=0, clip]
		{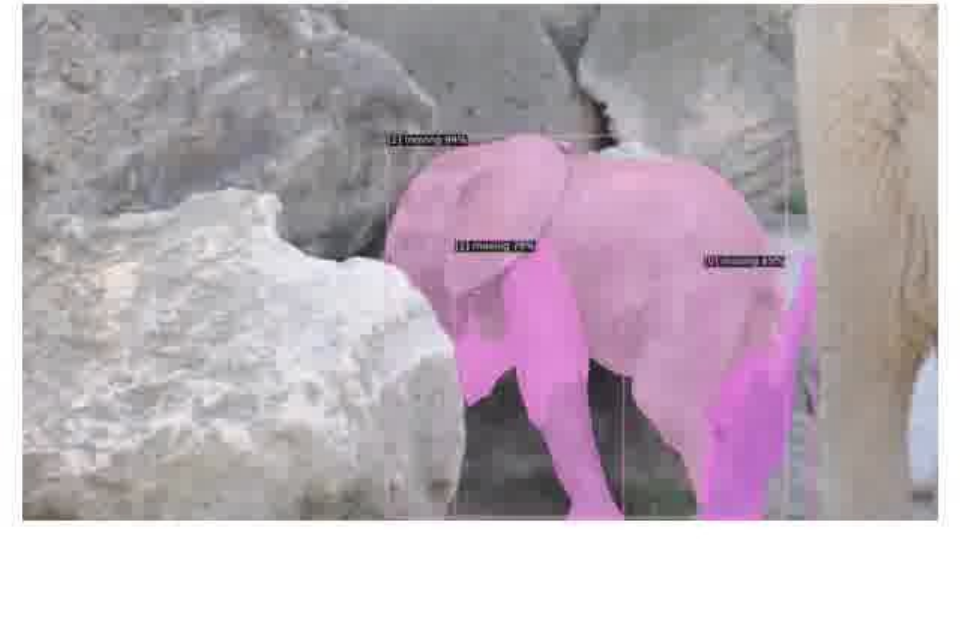}
	\end{overpic}\hspace*{-0.3em}
	\begin{overpic}[width=.19\linewidth, height=2cm, trim={0.0cm 2.0cm 0.0cm 0.1cm}, tics=0, clip]
		{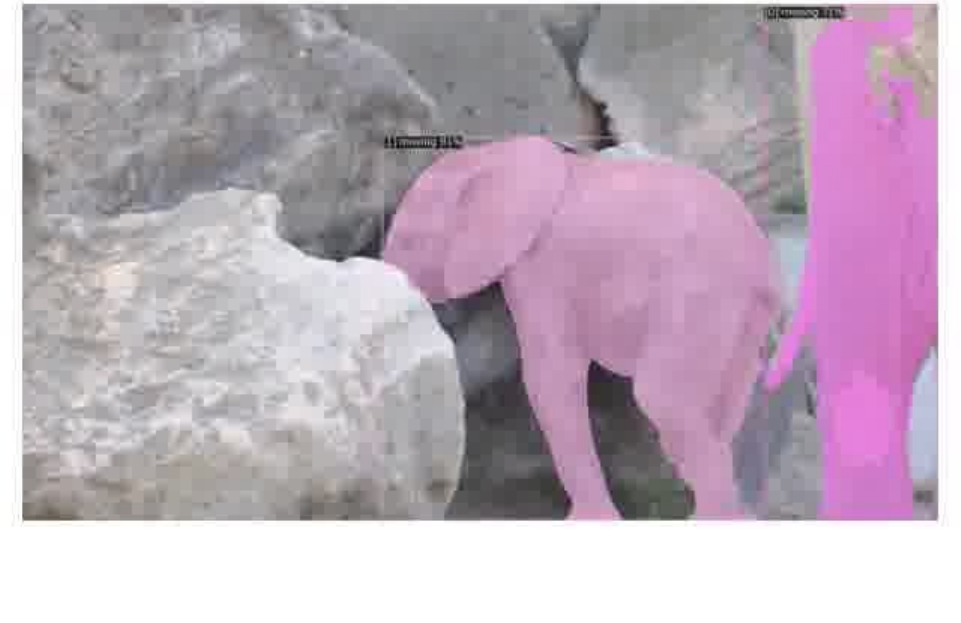}
	\end{overpic}\hspace*{-0.3em}
	\begin{overpic}[width=.19\linewidth, height=2cm, trim={0.0cm 2.0cm 0.0cm 0.1cm}, tics=0, clip]
		{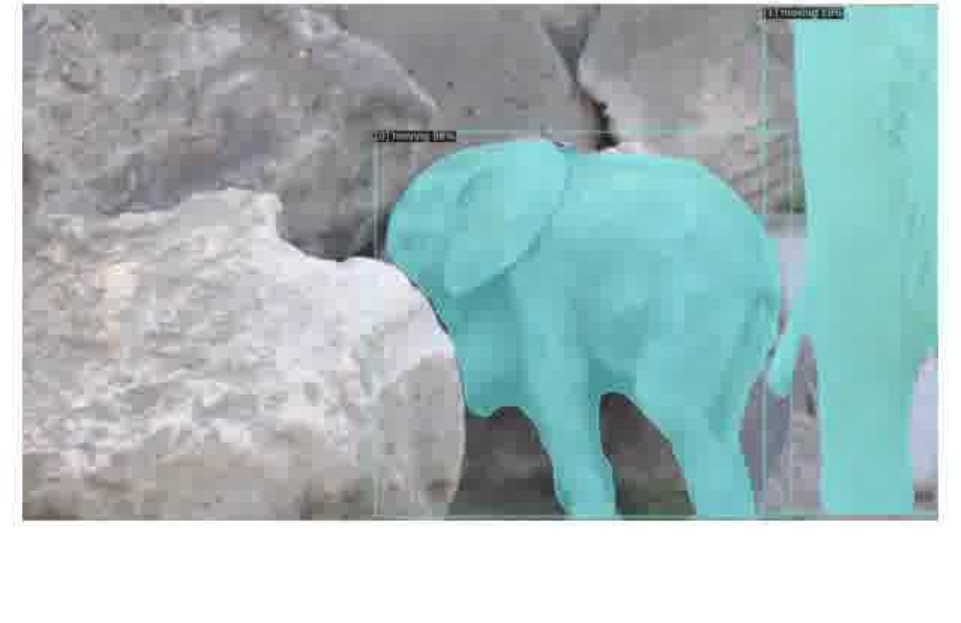}
	\end{overpic}
	
	\caption{\textit{Benefit of a twin two-stream architecture}: There exist scenes, where motion can drive the discovery and segmentation of objects, e.g. camouflaged animals \cite{lamdouar2020betrayed}. While a purely appearance based detector \cite{cheng2021mask2former} can fail, optical flow acts as a strong cue for detection, but cannot complete partial objects or leads to undersegmentations. Combining both motion and appearance cues allows to do part-based completion to semantically meaningful animals. Our model seems to generalize better than related approaches \cite{neoral2021monocular}, likely due to the bigger architecture, attention mechanism, better appearance stream and more diverse training data. We additionally show some failure cases, where the model streams are not aligned correctly.}
	\label{fig:moca}
\end{figure*}
In this paper, we chose an equal number of weights $ \Theta_{rgb} $ and $ \Theta_{motion} $. In light of the fact that usually video drives segmentation performance, this seems to be an overkill. In Figure \ref{fig:moca} we show a counter example where clearly the motion stream drives performance.

\end{document}